\documentclass[10pt,final,journal,twocolumn]{IEEEtran}
\usepackage{algorithm}
\usepackage{algpseudocode}
\usepackage{amsmath}
\usepackage{amssymb}
\usepackage{amsthm}
\usepackage{bm}
\usepackage{booktabs}
\usepackage{color}
\usepackage{cite}
\usepackage{epsfig}
\usepackage{diagbox}
\usepackage{epstopdf}
\usepackage{flafter}
\usepackage{float}
\usepackage{ifthen}
\usepackage{multirow}
\usepackage{amsfonts}
\usepackage{makecell} % Make references as [1-4], not [1,2,3,4]
\usepackage{pifont}
\usepackage{subfigure}
\usepackage{times}
\usepackage{threeparttable}
\usepackage{url} % Formatting web addresses % Conditional
\usepackage{subfigure,graphicx}
\usepackage{color,multirow}
\usepackage{makecell}
\usepackage{setspace}
\usepackage{cite} % Make references as [1-4], not [1,2,3,4]
\usepackage{url} % Formatting web addresses
\usepackage{ifthen} % Conditional
\usepackage{graphicx}
\usepackage{changepage}
\usepackage[colorlinks,linkcolor=red,anchorcolor=gray,citecolor=green,urlcolor=black]{hyperref}
\usepackage{graphics}
\usepackage{utfsym}
\usepackage{booktabs}

\newtheorem{lemma}{Lemma}

\newcommand{\tabincell}[2]{\begin{tabular}{@{}#1@{}}#2\end{tabular}}
\usepackage{color}
\definecolor{RED}{RGB}{146,57,49}
\definecolor{GREEN}{RGB}{120,148,64}
\definecolor{BLUE}{RGB}{49,133,155}
\definecolor{myGreen}{rgb}{0.945,0.972,0.980}
\definecolor{myblue}{rgb}{0.898,0.972,1}
\begin{document}
\title{Revisiting Nonlocal Self-Similarity from Continuous Representation
\thanks{Yisi Luo and Deyu Meng are with the School of Mathematics and Statistics, Xi'an Jiaotong University, Xi'an, P.R.China (e-mail: yisiluo1221@foxmail.com, dymeng@mail.xjtu.edu.cn).}
\thanks{Xile Zhao is with the School of Mathematical Sciences, University of Electronic Science and Technology of China, Chengdu, P.R.China (e-mail: xlzhao122003@163.com).}}
\author{Yisi Luo,
Xile Zhao, \IEEEmembership{Member, IEEE},
Deyu Meng, \IEEEmembership{Member, IEEE}}
\maketitle
\begin{abstract}
Nonlocal self-similarity (NSS) is an important prior that has been successfully applied in multi-dimensional data processing tasks, e.g., image and video recovery. However, existing NSS-based methods are solely suitable for meshgrid data such as images and videos, but are not suitable for emerging off-meshgrid data, e.g., point cloud and climate data. In this work, we revisit the NSS from the continuous representation perspective and propose a novel Continuous Representation-based NonLocal method (termed as CRNL), which has two innovative features as compared with classical nonlocal methods. First, based on the continuous representation, our CRNL unifies the measure of self-similarity for on-meshgrid and off-meshgrid data and thus is naturally suitable for both of them. Second, the nonlocal continuous groups can be more compactly and efficiently represented by the coupled low-rank function factorization, which simultaneously exploits the similarity within each group and across different groups, while classical nonlocal methods neglect the similarity across groups. This elaborately designed coupled mechanism allows our method to enjoy favorable performance over conventional NSS methods in terms of both effectiveness and efficiency. Extensive multi-dimensional data processing experiments on-meshgrid (e.g., image inpainting and image denoising) and off-meshgrid (e.g., climate data prediction and point cloud recovery) validate the versatility, effectiveness, and efficiency of our CRNL as compared with state-of-the-art methods.  
\end{abstract}
\IEEEpeerreviewmaketitle
\section{Introduction}
\iffalse
\begin{figure}[t]
	\scriptsize
	\setlength{\tabcolsep}{0.9pt}
	\begin{center}
		\begin{tabular}{c}
			\includegraphics[width=0.43\textwidth]{fig1.pdf}\\
		\end{tabular}
	\end{center}
	\vspace{-0.3cm}
	\caption{As compared with classical NSS-based methods, e.g., WNLRATV\cite{WNLRATV} and LTDL \cite{LTDL}, our continuous representation-based nonlocal method is more versatile for data processing on and off-meshgrid.\label{fig_1}	\vspace{-0.4cm}}
\end{figure}
\fi
\begin{table}[t]
	\caption{Comparisons of classical nonlocal methods and our new continuous representation-based nonlocal method in terms of applicabilities and encoded priors.\label{tab_1}}\vspace{-0.4cm}
	\begin{center}
		\tiny
		\setlength{\tabcolsep}{0.7pt}
		\begin{spacing}{1.8}
			\begin{tabular}{clcccccc}
				\hline
				\multicolumn{2}{c|}{\diagbox{Characteristic}{Method}}&BM3D\cite{BM3D}&NLR-CS\cite{TIP_14_CS}&WNNM\cite{IJCV_WNN}&WNLRATV\cite{WNLRATV}&NL-FCTN\cite{NLFCTN}&CRNL\\
				\hline
   				\multirow{2}{*}{Applicabilities}&Meshgrid data&\checkmark&\checkmark&\checkmark&\checkmark&\checkmark&\checkmark\\
   				~&Off-meshgrid data&&&&&&\checkmark\\
				\hline
				\multirow{2}{*}{Priors}&Similarity within a group&\checkmark&\checkmark&\checkmark&\checkmark&\checkmark&\checkmark\\
				~&Similarity across groups&&&&&&\checkmark\\
				\hline
			\end{tabular}
		\end{spacing}
	\end{center}
	\vspace{-0.9cm}
\end{table}    
\IEEEPARstart{W}{ith} the rapid development of imaging and sensing technologies, numerous types of multi-dimensional data are readily available. Among them, signals with meshgrid structures can be conventionally modeled as arrays with one or multiple dimensions, e.g., a gray image can be represented by a matrix and a color image can be represented by a third-order tensor. Digging their intrinsic structures via hand-crafted techniques is an effective approach for many signal processing tasks. For example, real-world data usually has internal low-dimensional structures, and thus low-rank representation is a popular technique for data analysis and processing, e.g., inpainting \cite{FTNN}, denoising \cite{LTDL}, and compressed sensing \cite{DeSCI}. Besides, many data have intense local smoothness, which can be finely characterized by smooth regularizations such as total variation \cite{TV_LRMR,E3DTV} and smooth factorization \cite{sp,spl_smooth}.\par 
One of the mostly employed prior structures of real-world data is the nonlocal self-similarity (NSS). The NSS refers to the fact that a signal often contains many repetitive local patterns, and thus a local pattern always has many similar patterns across the whole signal \cite{SCIS_NSS,PAMI_12}. The NSS of data has been widely used in different multi-dimensional processing tasks, e.g., image restoration \cite{SCIS_NSS,CVPR_18_NL,IJCV_WNN}, hyperspectral denoising \cite{TGRS_19_NL,CVPR_17}, compressed sensing \cite{NL_CS,PAMI_SCI}, and tensor completion \cite{TCYB_NLLR}. \par 
The conventional approach of these NSS-based methods is to search similar patches and then group these patches into a new matrix/tensor, followed by low-rank or/and smooth regularizations performed on the constructed matrix/tensor to effectively reconstruct the original signal\cite{WNLRATV,TGRS_Zha,SP_NL}. Such strategy utilizes the nonlocal information and is usually more effective than directly performing low-rank/smooth regularizations in the original domain \cite{AMM_derain,IP_nonlocal}, since the similar patch groups are expected to enjoy more evident low-rank/smooth structures. Currently, NSS has also been integrated into many deep learning frameworks \cite{TMM_NL,TMM_NL2,CVPR_CS}, showing certain advantages over traditional neural networks due to the beneficial nonlocal information.
\par 
Although these NSS-based methods have made great success in multiple areas, they are solely suitable to exploiting the NSS of meshgrid data, such as images. However, many emerging real-world data are not arranged as aligned arrays, but rather placed in unordered off-meshgrid positions in the space, e.g., point cloud and climate data. In order to exploit the underlying NSS of such off-meshgrid data, it is needed to develop a new fundamental nonlocal method that is suitable for both on-meshgrid and off-meshgrid data.\par
To meet this challenge, we revisit the NSS from the continuous representation perspective and propose a novel Continuous Representation-based NonLocal method (termed as CRNL), which can exploit the NSS of both on-meshgrid and off-meshgrid data under a unified framework. Concretely, we first use implicit neural representation (INR) to learn a continuous representation of observed discrete data. Then, we divide the continuous space into basic continuous cubes. We measure the similarity between continuous cubes, instead of discrete patches, using the learned continuous representation and stack similar cubes into a continuous group. To compactly and efficiently represent nonlocal continuous groups, we propose the coupled low-rank function factorization, where the shared factor functions exploit the similarity across different groups, and the unshared core tensors respect the individuality of each group. The compact coupled factorization allows our method to enjoy advantageous performance in terms of both effectiveness and computational efficiency. A general illustration of our CRNL is illustrated in Fig. \ref{fig_flow}.\par
Compared to classical NSS-based methods, the essential advantages of our CRNL are contained in two folds. First, based on the continuous representation, our CRNL can measure the self-similarity of both on-meshgrid and off-meshgrid data. It is thus expected to be more versatile than conventional NSS-based methods, which are solely suitable for meshgrid data. Second, the nonlocal continuous groups are compactly and efficiently represented by the coupled low-rank function factorization, which simultaneously exploits the similarity within each group and across different groups; see Lemma \ref{Pro_2}. As compared, classical NSS-based methods employ independent uncoupled representations for different groups, which neglects the similarity across groups. Such compact coupled factorization manner makes our method not only able to enjoy favorable computational efficiency with elaborate parameters sharing across different NSS groups, but also hopefully enhance the performance of multi-dimensional data recovery through essential correlation knowledge existed across different groups.\par 
In summary, this paper makes the following contributions: 
\begin{itemize}
\item We revisit the NSS from the continuous representation perspective and propose the CRNL, which can measure the self-similarity of both on-meshgrid and off-meshgrid data through the continuous representation. As compared with classical nonlocal methods, our CRNL is more versatile for real-world multi-dimensional data processing on and off-meshgrid. 
\item We propose the coupled low-rank function factorization to represent nonlocal continuous groups, which simultaneously exploits the similarity within each group and across different groups to more comprehensively modeling the underlying structures of multi-dimensional data. The elaborately designed coupled mechanism allows our method to enjoy advantageous computational efficiency as compared with classical nonlocal methods. 
\item To demonstrate the effectiveness of our method, we conduct experiments on multiple data recovery tasks including image inpainting, denoising (on-meshgrid data), and multivariate regression problems (off-meshgrid data, e.g., climate data and point clouds). Extensive results demonstrate the effectiveness of our method as compared with state-of-the-art methods.
\end{itemize}\par
The rest of this paper is organized as follows. In Sec. \ref{sec_rela}, we introduce related work. In Sec. \ref{sec_method}, we present the proposed method. In Sec. \ref{sec_exp}, we carry out experiments on different tasks and make some discussions on the proposed method. Sec. \ref{sec_conclusion} concludes this paper.
\section{Related Work}\label{sec_rela}
\subsection{Nonlocal Self-Similarity-Based Methods}
The NSS-based methods have been widely studied in the literature. The pioneer works, e.g., \cite{CVPR_05} and \cite{BM3D}, considered nonlocal means or nonlocal filters for image recovery. Later works mostly considered nonlocal sparse/low-rank regularizations for image recovery, e.g., by using the weighted nuclear norm \cite{IJCV_WNN} or groups sparsity \cite{TIP_Dong} performed on nonlocal patch groups. Recently, many nonlocal low-rank-based methods were proposed in different areas, such as compressive sensing \cite{TIP_14_CS,DeSCI}, hyperspectral image restoration \cite{TGRS_19_NL,WNLRATV}, video deraining \cite{AMM_derain}, magnetic resonance imaging \cite{IP_nonlocal}, and seismic data processing \cite{TGRS_Chen_21}. These wide applications have demonstrated the effectiveness of NSS-based methods. As compared, our method has two innovative features that are significantly different from existing NSS-based methods. First, existing NSS-based methods are solely suitable for meshgrid data processing\cite{SCIS_NSS,CVPR_18_NL,IJCV_WNN}, while our method learns a continuous representation of the observed discrete data, and hence is suitable for both on-meshgrid and off-meshgrid data. Second, existing NSS-based methods use classical low-rank regularization such as weighted nuclear norm \cite{IJCV_WNN} and low-rank tensor factorization \cite{NLLRF} to reconstruct nonlocal patch groups, while our method leverages the coupled low-rank function factorization defined in a continuous domain to represent continuous groups. The compact coupled factorization significantly reduces computational costs and simultaneously characterizes both the intra- and inter-group similarity, which helps to better recover the underlying data.\par 
Recently, Chen et al. \cite{TVCG_20} and Zhu et al. \cite{TIM_22} have proposed nonlocal methods for point cloud recovery, which mainly design feature descriptors to encode the information of points as patch matrix, and then perform nonlocal low-rank matrix recovery to recover the noisy matrix of point cloud. These methods cleverly transform the point cloud processing task into matrix recovery. However, these methods were specifically designed for point clouds, while our method is a unified method that can be applied to both on-meshgrid and off-meshgrid data representation, e.g., images and point clouds.
\begin{figure*}[t]
	\scriptsize
	\setlength{\tabcolsep}{0.9pt}
	\begin{center}
		\begin{tabular}{c}
			\includegraphics[width=0.86\textwidth]{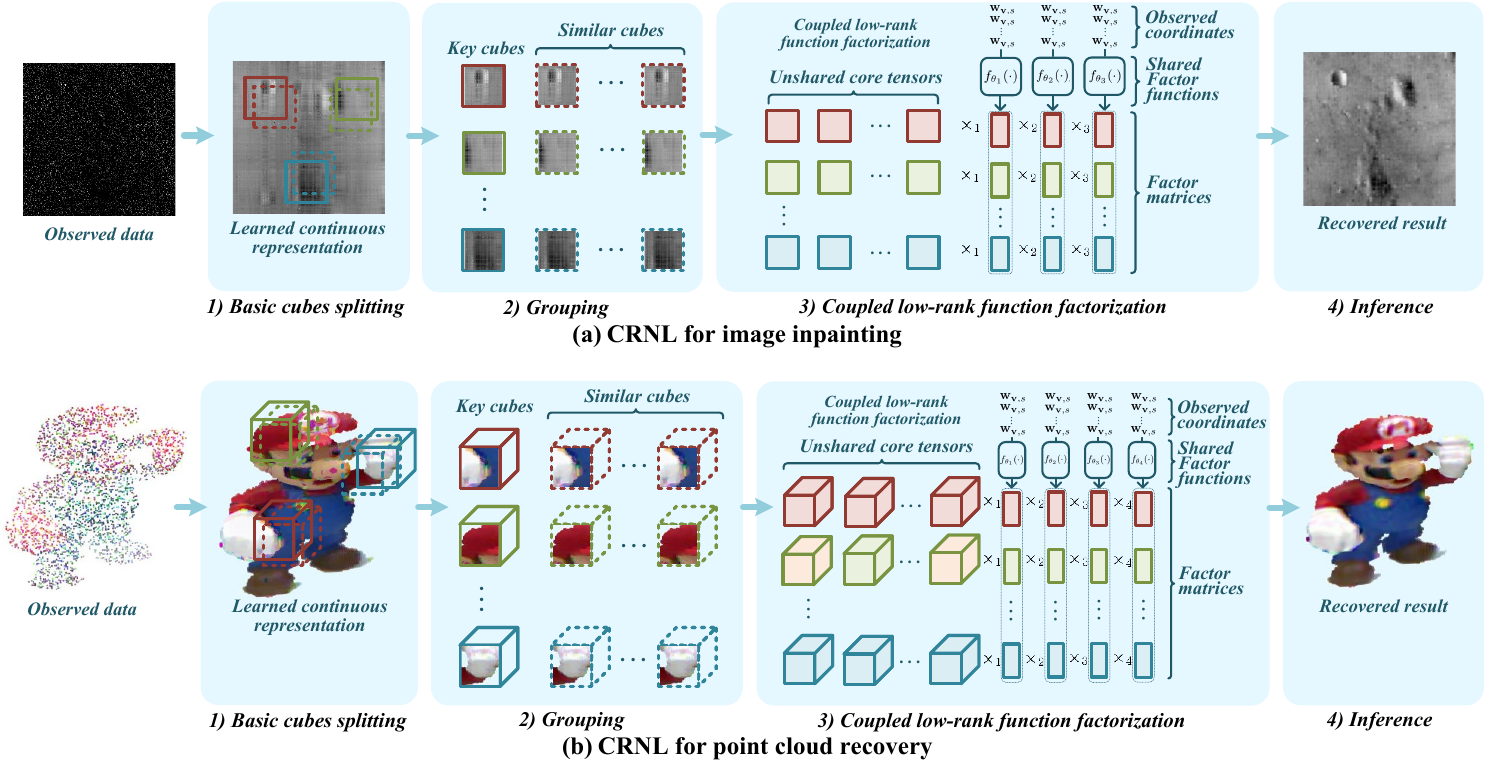}\\
		\end{tabular}
	\end{center}
	\vspace{-0.4cm}
	\caption{The overall flowchart of our continuous representation-based nonlocal method for data recovery. Here, we take the image inpainting and point cloud recovery as examlpes.\label{fig_flow}\vspace{-0.4cm}}
\end{figure*}
\subsection{Basis Function-Based Methods} 
In our CRNL, we propose the coupled low-rank function factorization to parameterize the nonlocal continuous groups for continuous representation. In the literature, some low-rank matrix/tensor methods also employ basis functions to parameterize the low-rank model, which is related to our method. The pioneer works \cite{sp,spl_smooth} considered using Gaussian basis or polynomial basis to parameterize the factor matrix in low-rank matrix factorization, which induces some implicit smoothness over the resultant matrix. The implicit smoothness was further introduced into higher-order low-Tucker-rank \cite{STD} and low-CP-rank tensor models \cite{TSP_CP} by using the Fourier basis. Oseledets \cite{TTF} introduced explicit representations of some multivariate functions (e.g., the polynomial and sine functions) in tensor-train factorization format. Hashemi and Trefethen \cite{SIAM_chebfun} studied Tucker factorization of functions and proposed algorithms for approximating functions in Tucker format, where the factor functions are represented by Chebyshev expansions. The main differences between our method and these works are three folds. First, our method additionally exploits the nonlocal information of data to more accurately capture the implicit low-rank structures. Second, our method can be applied to both on-meshgrid and off-meshgrid data processing, while previous basis function-based low-rank methods are hard to be directly applied to data processing tasks beyond meshgrid. Third, we employ the implicit deep neural network \cite{sine} as basis functions to parameterize the coupled low-rank function factorization, while previous methods use shallow basis functions, and thus our method is expected to hold stronger representation abilities.
\subsection{Continuous Representation-Based Methods} 
The continuous representation is a powerful tool for data representation on and off-meshgrid. The most popular method of continuous representation is the implicit neural representation, or INR \cite{sine,INR_NIPS_21}. The core idea of INR is to use a neural network that maps coordinates to the corresponding value for data representation. It is a powerful tool for representing various signals like images \cite{INR_SR_TIP}, shapes \cite{cvpr_3d}, and point clouds \cite{SAPCU}. INR has also shown promising performances for data recovery tasks. For example, Chen et al. \cite{LIIF} proposed the local implicit image function for image super-resolution. It was further enhanced by introducing details recovery technique \cite{INR_SR_TIP} and advanced deep network \cite{INR_SR_NIPS}. Zhang et al. \cite{TGRS_INR} proposed to use INR for hyperspectral image super-resolution. However, these methods are supervised methods that need supervised training process, while our CRNL is a hand-crafted unsupervised method, which learns the continuous representation directly from observed data. Another type of INR is the unsupervised one, e.g., Zhao et al. \cite{SAPCU} utilized the INR for self-supervised point cloud upsampling, and Kim et al. \cite{denoise_INR_1} proposed to use INR for zero-shot image recovery. As compared with these two unsupervised data recovery methods using INR, our CRNL additionally leverages the nonlocal low-rankness to capture internal low-dimensional structures of data, which is helpful to increase the effectiveness and generalization abilities of the continuous representation.
\section{The Proposed Method}\label{sec_method}
\subsection{Preliminaries}
Scalars, vectors, matrices, and tensors are denoted by $x$, $\bf x$, $\bf X$, and $\mathcal X$, respectively. The $i$-th element of $\bf x$ is denoted by ${\bf x}_{(i)}$, and it is similar for matrices and tensors, i.e., ${\bf X}_{(i_1,i_2)}$ and ${\mathcal X}_{(i_1,i_2\cdots,i_N)}$. When we use the index $:$, e.g., ${\bf X}_{(i,:)}$, it represents the $i$-th row of $\bf X$. The tensor Frobenius norm of a $N$-th-order tensor ${\mathcal X}\in{\mathbb R}^{n_1\times \cdots\times n_N}$ is defined as $\left\lVert{\mathcal X}\right\rVert_F:=\sqrt{\langle {\mathcal X},{\mathcal X}\rangle}=\sqrt{\sum_{i_1,\cdots,i_N}{\mathcal X}_{(i_1,\cdots,i_N)}^2}$. The tensor $\ell_1$-norm is defined as $\left\lVert {\mathcal X}\right\rVert_{\ell_1}:=\sum_{i_1,\cdots,i_N}|{\mathcal X}_{(i_1,\cdots,i_N)}|$. The unfolding operator of a tensor ${\mathcal X}\in{\mathbb R}^{n_1\times \cdots\times n_N}$ along the $d$-th mode ($d=1,2,\cdots,N$) is defined as ${\tt unfold}_d(\cdot):{\mathbb R}^{n_1\times \cdots\times n_N}\rightarrow{\mathbb R}^{n_d\times \prod_{j\neq d}n_j}$, which returns the unfolding matrix along the mode $d$, and the unfolding matrix is denoted by ${\bf X}^{(d)}:={\tt unfold}_d({\mathcal X})$. ${\tt fold}_d(\cdot)$ denotes the inverse operator of ${\tt unfold}_d(\cdot)$. The mode-$d$ ($d=1,2,\cdots,N$) tensor-matrix product is defined as ${\mathcal X}\times_d{\bf A}:={\tt fold}_d({\bf A}{\bf X}^{(d)})$, which returns a tensor.
\begin{lemma}(Tensor Tucker factorization \cite{SIAM_review})\label{th_Tucker}
	The Tucker rank of a $N$-th-order tensor ${\mathcal X}\in{\mathbb R}^{n_1\times \cdots\times n_N}$ is a vector defined as ${\rm rank}_T({\mathcal X}):=({\rm rank}({\bf X}^{(1)}), {\rm rank}({\bf X}^{(2)}),\cdots,{\rm rank}({\bf X}^{(N)}))$.
	\begin{itemize}
	\item[(i)] If ${\rm rank}_T({\mathcal X})=(r_1,r_2,\cdots,r_N)$, then there exist a core tensor ${\mathcal C}\in{\mathbb R}^{r_1\times r_2\times\cdots\times r_N}$ and $N$ factor matrices ${\bf U}_d\in{\mathbb R}^{n_d\times r_d}$ ($d=1,2,\cdots,N$) such that ${\mathcal X}={\mathcal C}\times_1{\bf U}_1\times_2{\bf U}_2\times_3\cdots\times_N{\bf U}_N$.
	\item[(ii)] Let ${\mathcal C}\in{\mathbb R}^{r_1\times \cdots\times r_N}$ be an arbitrary $N$-th-order tensor, ${\bf U}_d\in{\mathbb R}^{n_d\times r_d}$ ($d=1,2,\cdots,N$) be arbitrary matrices ($r_d\leq n_d$). Then $\big{(}{\rm rank}_T({\mathcal C}\times_1{\bf U}_1\times_2{\bf U}_2\times_3\cdots\times_N{\bf U}_N)\big{)}_{(d)}\leq r_d$ ($d=1,2,\cdots,N$).   
	\end{itemize}
\end{lemma}
\begin{table}[t]
\scriptsize
	\caption{Notations used in this section.\label{tab_notation}}\vspace{-0.4cm}
	\begin{center}
		\setlength{\tabcolsep}{2pt}
		\begin{spacing}{1.1}
			\begin{tabular}{l|l}
				\hline
Notation&Description\\
\hline
$N$&The number of dimensions of the observed data\\
$h(\cdot)$&The observed multivariate function\\
$D_h\subset{\mathbb R}^N$&The set of observed coordinates\\
${D_h}^C\subset{\mathbb R}^N$&The set of unobserved coordinates (testing set)\\
$f_{\theta}(\cdot)$&The continuous representation of the observed data\\
$D_{f_\theta}\subset{\mathbb R}^{N}$&The entire continuous domain\\
$[a_d,b_d]\subset{\mathbb R}$&The continuous domain in the $d$-th dimension ($d=1,\cdots,N$)\\
$n_d$&The number of basic units in the $d$-th dimension\\
$\delta_d\in{\mathbb R}$&The basic unit length in the $d$-th dimension\\
$p\delta_d$&The basic cube length in the $d$-th dimension ($p$ is an integer)\\
$D_t\subset{\mathbb R}^N$&The $t$-th overlapped basic continuous cube ($t=1,\cdots,T$)\\
$D_l^{\rm key}\subset{\mathbb R}^{N}$& The $l$-th non-overlapping key continuous cube ($l=1,\cdots,L$)\\
$D_{l_s}\subset{\mathbb R}^{N}$ & The $s$-th similar cube of the $l$-th key cube ($s=1,\cdots,S$)\\
${\hat D}_{h_l}\subset{\mathbb R}^{N+1}$&The set of observed points of the $l$-th continuous group\\ 
\hline
\end{tabular}
\end{spacing}
\end{center}
\vspace{-0.7cm}
\end{table}
\subsection{Overview of the Proposed CRNL}
Classical NSS-based methods are suitable for meshgrid data processing, e.g., image inpainting, but are not suitable for off-meshgrid data processing, e.g., point cloud processing. To address this limitation, we revisit the NSS from the continuous representation perspective and propose the CRNL for data processing on and off-meshgrid. Specifically, we view the observed data as a real function $h(\cdot):{\mathbb R}^{N}\rightarrow{\mathbb R}$, where $N$ denotes the number of dimensions of the observed data (e.g., for a color image, $N=3$ corresponds to height, width, and channels). Suppose that we are given some discrete observations of $h(\cdot)$, i.e., we have a coordinate set with limited number of coordinate points ${D_h}\subset{\mathbb R}^N$ and the corresponding observed function values $h({\bf v}),{\bf v}\in{D_h}$. The goal is often to infer the function values of $h(\cdot)$ at unobserved positions in ${D_h}^C$, where ${D_h}^C$ denotes the complementary set of ${D_h}$, i.e., the unobserved coordinates in the whole set. Typical examples include image inpainting and multivariate regression problems, where $D_h$ is located in meshgrid and non-meshgrid coordinate locations, respectively.\par 
To more clearly introduce our method, we first give a rough description in this subsection, and then illustrate the detailed algorithms step by step. In general, our method includes the following steps for multi-dimensional data processing:
\begin{itemize}
\item{\bf Basic cubes splitting} We first use INR to learn a continuous representation of the observed discrete data. Based on the continuous representation, we split the continuous space into basic continuous cubes. 
\item{\bf Grouping} We use the continuous representation to measure the similarity between basic continuous cubes. We stack similar continuous cubes into a continuous group. Each continuous group contain a certain number of observed points; see Fig. \ref{fig_nonlocal}.  
\item{\bf Coupled low-rank function factorization} We propose the coupled low-rank function factorization to compactly represent continuous groups, where the shared factor functions characterize the similarity across groups, and the unshared core tensors reveal the individuality of each group. The coupled function factorization allows our method to enjoy favorable computational efficiency for representing numerous number of continuous groups.
\end{itemize}\par
As compared with classical discrete representation-based NSS methods, the essential differences of our CRNL are in two folds. First, we use the INR to learn a continuous representation of data and split the continuous space into basic continuous cubes, which naturally allows us to handle data on and off-meshgrid. Second, different from classical NSS methods that adopt discrete and separate low-rank representation for each nonlocal patch group, our CRNL leverages the coupled low-rank function factorization defined in a continuous space to more compactly and accurately represent all nonlocal groups in a parameters sharing manner through globally extracting correlation knowledge among them. An overview of our proposed CRNL for multi-dimensional data recovery is illustrated in Fig. \ref{fig_flow}. 
\subsection{Detailed Algorithm of CRNL}
{\bf 1) Basic cubes splitting} 
Given the observed function $h(\cdot):{\mathbb R}^N\rightarrow {\mathbb R}$ with observed discrete coordinate set ${D_h}\subset{\mathbb R}^N$, we first use the INR \cite{sine} to learn a continuous representation of the observed data. Specifically, we use an explicit form of neural network $f_\theta(\cdot):{\mathbb R}^N\rightarrow {\mathbb R}$ to approximate the observed function $h(\cdot)$ over the observed discrete set ${D_h}$:
\begin{equation}\label{model_INR}
	\begin{split}
		&\min_{\theta}\sum_{{\bf v}\in {D_h}}(h({\bf v}) - f_\theta({\bf v}))^2+\psi[f_\theta(\cdot)],
	\end{split}
\end{equation}
where $\theta$ denote the learnable parameters of the neural network $f_\theta(\cdot)$ and $\psi[f_\theta(\cdot)]$ denotes a regularization term that enhances the performance (The regularization term is conditioned on tasks, specified in Sec. \ref{sec_reg}). We use the Adam optimizer\cite{Adam} to optimize the INR model (\ref{model_INR}). \par 
Next, we conduct basic cubes splitting. For easy reference, a notation table is provided in Table \ref{tab_notation}. The basic cubes splitting includes the following steps:
\begin{figure*}[t]
	\scriptsize
	\setlength{\tabcolsep}{0.9pt}
	\begin{center}
		\begin{tabular}{c}
			\includegraphics[width=0.73\textwidth]{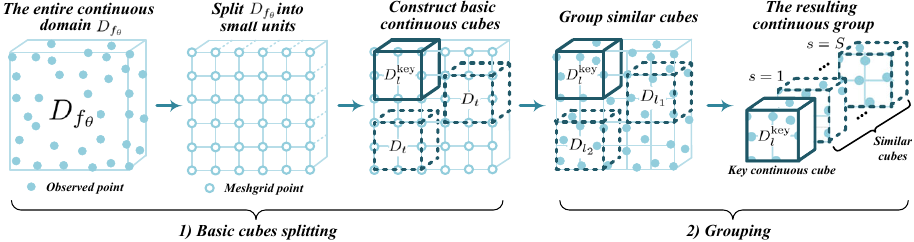}\\
		\end{tabular}
	\end{center}
	\vspace{-0.3cm}
	\caption{Illustrations of the proposed basic cubes splitting and grouping process in the three-dimensional case ($N=3$) with number of units $n_1=n_2=5$ and cube size $p=2$. Here, $D_t$ denotes the basic continuous cube, $D_l^{\rm key}$ denotes a key continuous cube, and $D_{l_1},D_{l_2}$ denote the two extracted similar continuous cubes of the $l$-th key cube $D_l^{\rm key}$.\label{fig_nonlocal}}
	\vspace{-0.4cm}
\end{figure*}
\begin{itemize}
\item First, the observed coordinate set $D_h\subset{\mathbb R}^N$ contains $N$ dimensions (e.g., $N=2$ for a gray image and $N=3$ for a color image). For the $d$-th dimension ($d=1,\cdots,N$), we denote the minimal value of the observed data coordinates as $a_d := \min_{{\bf v}\in D_h}{\bf v}_{(d)}$, where ${\bf v}_{(d)}$ denotes the $d$-th element of the vector ${\bf v}$. Similarly we denote the maximum value of the observed data coordinates in the $d$-th dimension as $b_d := \max_{{\bf v}\in D_h}{\bf v}_{(d)}$. The continuous domain underlying the observed data in the $d$-th dimension is $[a_d,b_d]\subset{\mathbb R}$ ($d=1,\cdots,N$). Based on the continuous domain of each dimension, the entire $N$-dimensional continuous domain $D_{f_\theta}\subset{\mathbb R}^{N}$ is a cubic space $D_{f_\theta}:=[a_1,b_1]\times [a_2,b_2]\times\cdots\times[a_N,b_N]\subset{\mathbb R}^N$.
\item Second, we split the continuous space $D_{f_\theta}$ into small basic continuous units. Specifically, we split the first dimension of $D_{f_\theta}$, i.e., $[a_1,b_1]$, into $n_1$ non-overlapping regions $[a_1,a_1+\delta_1),[a_1+\delta_1,a_1+2\delta_1),\cdots,[a_1+(n_1-1)\delta_1,a_1+n_1\delta_1)$, where $\delta_1$ denotes the interval in the first dimension such that $a_1+n_1\delta_1=b_1$. Similarly, we split the second dimension $[a_2,b_2]$ into $n_2$ non-overlapping regions, i.e., $[a_2,a_2+\delta_2),[a_2+\delta_2,a_2+2\delta_2),\cdots,[a_2+(n_2-1)\delta_2,a_2+n_2\delta_2)$, where $a_2+n_2\delta_2=b_2$. Such split leads to a finite disjoint coverage over $D_{f_\theta}$ with $n_1n_2$ units, where each unit has the volume of $\delta_1\times \delta_2\times (b_3-a_3)\times\cdots\times(b_N-a_N)$. 
\item Third, we construct basic continuous cubes. A continuous cube is composed of $p^2$ basic units, where $p$ determines the cube size. We first construct {\sl overlapped} basic continuous cubes with volume $p\delta_1\times p\delta_2 \times (b_3-a_3)\times\cdots\times(b_N-a_N)$ and stride 1 along the first and second dimensions. It results in a continuous cubes set $\{D_t\subset{\mathbb R}^N\}_{t=1}^{T}$ that contains $T=(n_1-p+1)(n_2-p+1)$ basic continuous cubes. Similarly, we construct {\sl non-overlapping} key continuous cubes with the same cube size but with stride $p$, which results in a key continuous cubes set $\{D_l^{\rm key}\subset{\mathbb R}^N\}_{l=1}^{L}$ that contains $L = (n_1/p)(n_2/p)$ key cubes\footnote{In the case where $n_1/n_2$ is not divisible by $p$, we use replication padding to expand data boundaries such that $n_1/n_2$ is divisible by $p$.}.
\end{itemize}\par
Fig. \ref{fig_nonlocal} illustrates a basic cubes splitting process with $n_1=n_2=5$ and $p=2$. \par
{\bf 2) Grouping} Next, we use the learned continuous representation $f_\theta(\cdot)$ to measure the similarity between basic continuous cubes. Specifically, for each non-overlapping key continuous cube $D_l^{\rm key}$, we aim to find its top-$S$ similar continuous cubes in the overlapped candidate cubes set $\{D_t\}_{t=1}^{T}$. The distance between two continuous cubes $D_l^{\rm key}$ and $D_t$ is calculated by the Euclidean distance between function values of $f_\theta(\cdot)$ at corresponding meshgrid positions in these two cubes, i.e., the sum of $(f_\theta({\bf v}^{\rm key})-f_\theta({\bf v}))^2$, where ${\bf v}^{\rm key}$ and ${\bf v}$ are one-to-one corresponding meshgrid coordinate points in $D_l^{\rm key}$ and $D_t$. For each key cube $D_l^{\rm key}$, we measure its similarity with each candidate cube in $\{D_t\}_{t=1}^T$ using this distance metric and stack the top-$S$ similar cubes into a continuous group $\{D_{l_s}\}_{s=1}^S$. Fig. \ref{fig_nonlocal} gives an intuitive illustration.\par  
A continuous group $\{D_{l_s}\}_{s=1}^S$ is then composed of $S$ similar cubes. The coordinate values in different cubes are not within the same area, which bring difficulty for the subsequent continuous group representation. Hence, for each continuous cubes group $\{D_{l_s}\}_{s=1}^S$, we gather all observed points in different cubes into the same area to construct a new observed points set. Specifically, for any $s\in\{1,2,\cdots,S\}$, we modify the values of the observed points in $D_{l_s}$ such that the new observed points fall into the key cube space $D_l^{\rm key}$. Formally, we denote the spatial distance (say $\Delta x$ and $\Delta y$) between the key cube $D_l^{\rm key}$ and the similar cube $D_{l_s}$ as
\begin{equation}
\begin{split}
&\Delta x = \min_{{\bf u}\in D_l^{\rm key}} {\bf u}_{(1)}-\min_{{\bf u}\in D_{l_s}} {\bf u}_{(1)},\\
&\Delta y = \min_{{\bf u}\in D_l^{\rm key}} {\bf u}_{(2)}-\min_{{\bf u}\in D_{l_s}} {\bf u}_{(2)}.
\end{split}
\end{equation}
Then, for any observed point ${\bf v}\in D_{l_s}\cap D_h$, where $D_{l_s}\cap D_h$ denotes the set of observed points in the continuous cube $D_{l_s}$, we modify its values to construct a new vector:
\begin{equation}
	\begin{split}
		\Big{(}{\bf v}_{(1)}+\Delta x, {\bf v}_{(2)}  +\Delta y, {\bf v}_{(3)},\cdots,{\bf v}_{(N)}\Big{)}\in D_l^{\rm key}. 
	\end{split}
\end{equation}
This new vector belongs to the key cube space $D_l^{\rm key}$. By further catenating this new vector with its cube index $s$, we obtain a new observed point
\begin{equation*}
	\begin{split}
		{\bf w}_{{\bf v},s}:=\Big{(}{\bf v}_{(1)}+\Delta x, {\bf v}_{(2)}  +\Delta y, {\bf v}_{(3)},\cdots,{\bf v}_{(N)},s\Big{)}\in{\mathbb R}^{N+1},
	\end{split}
\end{equation*}
which is conditioned on the original observed point $\bf v$ and the similar cube index $s$. We cluster all such points in the $l$-th continuous group $\{D_{l_s}\}_{s=1}^S$ to obtain a new observed dataset 
\begin{equation}
{\hat D}_{h_l} = \bigcup_{s\in\{1,2,\cdots, S\}}\bigcup_{{\bf v}\in{D}_{l_s}\cap{D_h}}\{{\bf w}_{{\bf v}, s}\}\subset{\mathbb R}^{N+1}.
\end{equation}
The new observed set ${\hat D}_{h_l}\subset{\mathbb R}^{N+1}$ contains all observed points in the $l$-th continuous group. The corresponding observed $(N+1)$-dimensional function $h_l(\cdot): \Omega\rightarrow {\mathbb R}$ ($l=1,2,\cdots,L$), where ${\hat D}_{h_l}\subset\Omega\subset{\mathbb R}^{N+1}$, map the observed points in ${\hat D}_{h_l}$ to the corresponding values, in which each multivariate function $h_l(\cdot)$ corresponds to a continuous group $\{D_{l_s}\}_{s=1}^S$. \par
{\bf 3) Coupled low-rank function factorization}
Inspired by many conventional strategies to represent NSS patch group by low-rank structures \cite{IJCV_WNN,NLLRF}, we also use the low-rank representation to represent each continuous group. Here, the observed points in ${\hat D}_{h_l}$ may be off-meshgrid, and thus standard discrete low-rank tensor representation is incapable of dealing with the off-meshgrid observations. Instead, our proposed coupled low-rank function factorization defined in a continuous domain can easily handle this issue.\par 
The core idea of low-rank function representation is to use a core tensor $\cal C$ and $N+1$ factor functions $f_{\theta_1}(\cdot),\cdots,f_{\theta_{N+1}}(\cdot)$ to represent the original ($N+1$)-dimensional function $h_l(\cdot)$ through the low-rank tensor function factorization \cite{LRTFR}. Theoretically, a multivariate function with low-rank structures can be exactly factorized into a core tensor and some factor functions. On the other hand, the product between a core tensor and some factor functions forms a multivariate function with low-rankness, as stated below\footnote{The proofs are placed in supplementary files.}.
\begin{lemma}\label{Pro_1}
	(Low-rank tensor function factorization) Let $h_l(\cdot):\Omega\rightarrow {\mathbb R}$ be a bounded multivariate function, where $\Omega=\Omega_1\times \Omega_2\times\cdots\times \Omega_{N+1}\subset{\mathbb R}^{N+1}$ is the definition domain. Define the ``function rank (F-rank)'' of $h_l(\cdot)$ as the maximum Tucker rank of tensors that can be sampled on the multivariate function $h_l(\cdot)$ with meshgrid partitions \cite{LRTFR}. Then:
\begin{itemize}
\item[(i)] (Existence of low-rank tensor function factorization) If ${\rm F\text{-}rank}[h_l]=(r_1,r_2,\cdots,r_{N+1})$, then there exist a core tensor ${\mathcal C}\in{\mathbb R}^{r_1\times r_2\times\cdots\times r_{N+1}}$ and $N+1$ bounded factor functions $f_1(\cdot):\Omega_1\rightarrow {\mathbb R}^{r_1}$, $f_2(\cdot):\Omega_2\rightarrow {\mathbb R}^{r_2}$,$\cdots$, $f_{N+1}(\cdot):\Omega_{N+1}\rightarrow {\mathbb R}^{r_{N+1}}$ such that for any ${\bf v}\in \Omega$, $h_l({\bf v})={\mathcal C}\times_1f_1({\bf v}_{(1)})\times_2f_2({\bf v}_{(2)})\times_3\cdots\times_{N+1}f_{N+1}({\bf v}_{(N+1)})$.\par
\item[(ii)] (Low-rankness guarantee of the tensor function factorization) Let ${\mathcal C}\in{\mathbb R}^{r_1\times r_2\times\cdots\times r_{N+1}}$ be an arbitrary tensor and $f_1(\cdot):\Omega_1\rightarrow {\mathbb R}^{r_1}$, $f_2(\cdot):\Omega_2\rightarrow {\mathbb R}^{r_2}$,$\cdots$, $f_{N+1}(\cdot):\Omega_{N+1}\rightarrow {\mathbb R}^{r_{N+1}}$ be $N+1$ arbitrary bounded factor functions. Then we have $({\rm F\text{-}rank}[h_l])_{
		(d)}\leq r_d$ ($d=1,2,\cdots,N+1$), where $h_l(\cdot):\Omega=\Omega_1\times \Omega_2\times\cdots\times \Omega_{N+1}\rightarrow{\mathbb R}$ is defined by $h_l({\bf v})={\mathcal C}\times_1 f_1({\bf v}_{(1)})\times_2 f_2({\bf v}_{(2)})\times_3 \cdots\times_{N+1} f_{N+1}({\bf v}_{(N+1)})$ for any ${\bf v}\in \Omega$.
		\end{itemize}
\end{lemma}
Lemma \ref{Pro_1} indicates that the low-rank function factorization can compactly represent a multivariate function in the low-rank format, and hence is suitable to represent the low-rank continuous group. Directly applying the low-rank function factorization to represent each group leads to large computational costs due to the numerous number of continuous groups. Specifically, it needs $L$ core tensors and $L\times(N+1)$ factor functions to represent $L$ continuous groups by using the low-rank function representation, which is quite computationally expensive. Meanwhile, similar to classical NSS-based methods \cite{NLring,WNLRATV,NLFCTN}, such uncoupled independent representations of nonlocal groups neglect the similarity across different groups, and hence is insufficient to accurately model the underlying structures of nonlocal groups.\par 
To address the above issues, we propose the coupled low-rank function factorization, which employs $N+1$ factor functions shared by all continuous groups. Specifically, we employ $L$ low-rank tensor functions $\{s_l(\cdot)\}_{l=1}^L$ parameterized by different core tensors ${\mathcal C}_l\in{\mathbb R}^{r_1\times r_2\times r_3\times\cdots\times r_{N+1}}$ ($l=1,2,\cdots,L$) but share the same $N+1$ factor functions $\{f_{\theta_d}(\cdot)\}_{d=1}^{N+1}$ to represent the continuous groups in the low-rank format. The coupled low-rank function factorization solely uses $L$ core tensors and $N+1$ factor functions to represent $L$ continuous groups, which inclines to largely reduce computational costs. Meanwhile, the coupled mechanism simultaneously exploits the similarity within each group and across different groups, which is expected to more faithfully take advantage of the intrinsic correlation knowledge across different local areas of multi-dimensional data (see Sec. \ref{simi} for detailed analysis).\par 
Formally, we propose the following coupled low-rank function factorization-based model to compactly represent continuous groups $\{h_{l}(\cdot)\}_{l=1}^L$:
\begin{equation}\label{model_NL}
	\begin{split}
		&\min_{\{{\mathcal C}_l\}_{l=1}^L, \{\theta_d\}_{d=1}^{N+1}}\sum_{l=1}^L\Big{(}\sum_{{\bf v}\in {\hat D}_{h_l}}(h_l({\bf v}) - s_l({\bf v}))^2+\psi[s_l(\cdot)]\;\Big{)},\;\\
		&s_l({\bf v}):={\mathcal C}_l\times_1 f_{\theta_1}({\bf v}_{(1)})\times_2\cdots\times_{N+1} f_{\theta_{N+1}}({\bf v}_{(N+1)}),
	\end{split}
\end{equation}
where $h_l(\cdot)$ is the observed function of the $l$-th continuous group, $s_l(\cdot):{\mathbb R}^{N+1}\rightarrow {\mathbb R}$ ($l=1,2,\cdots,L$) denotes the desired recovered low-rank function of the $l$-th continuous group, and $\psi[s_l(\cdot)]$ is a regularization term conditioned on tasks, specified in Sec. \ref{sec_reg}. Here, the $N+1$ factor functions $\{f_{\theta_d}(\cdot): {\mathbb R}\rightarrow {\mathbb R}^{r_d}\}_{d=1}^{N+1}$, which are shared by all continuous groups, are parameterized by $N+1$ fully-connected neural networks, where $\{\theta_d\}_{d=1}^{N+1}$ denote the learnable weights. We remark that the sine function is used as the nonlinear activation function of the fully connected neural networks in both INR model \eqref{model_INR} and our CRNL model \eqref{model_NL}. The sine function is shown to be more effective than other commonly used nonlinear functions for continuous representation \cite{sine}. We directly use the Adam \cite{Adam} algorithm to minimize the object (\ref{model_NL}) by optimizing core tensors $\{{\mathcal C}_l\}_{l=1}^L$ and weights of factor functions $\{\theta_d\}_{d=1}^{N+1}$.\par
{\bf 4) Inference}
By optimizing (\ref{model_NL}), we obtain the coupled low-rank function representations $\{s_l(\cdot)\}_{l=1}^L$ for the $L$ continuous groups. To infer the function values of $h(\cdot)$ in the unobserved set ${D_h}^C$, we feed the unobserved coordinate points in each key cube, i.e., the points in $D_l^{\rm key}\cap{D_h}^C$, into their corresponding low-rank function $s_l(\cdot)$, and then readily obtain the output function values. Then we rearrange these output results to their original positions to obtain the final results. %For both image recovery and multivariate regression problems, the key cubes $\{D_l^{\rm key}\}_{l=1}^L$ have no overlapping areas. Hence, we do not need to perform an aggregation process by averaging overlapped elements like in classical nonlocal methods \cite{NLring,NLFCTN}.
\subsection{Similarity within Each Group and across Different Groups}\label{simi}
As mentioned before, benefiting from the shared factor functions of the coupled function factorization, our method could characterize the similarity across different continuous groups, while the unshared core tensors respect the {individuality} of each continuous group. Hence, our method tends to capture both intra- and inter-group similarity. To theoretically verify this, we show the following result from the Lipschitz smooth perspective.
\begin{lemma}\label{Pro_2}
Let $\{{\mathcal C}_l\}_{l=1}^L$ be some bounded core tensors, and $f_{\theta_1}(\cdot)$, $f_{\theta_2}(\cdot)$,$\cdots$, $f_{\theta_{N+1}}(\cdot)$ be $N+1$ fully-connected neural networks with sine activation function $\sin(\omega \;\cdot)$ and depth $M$. Suppose that $\eta\in{\mathbb R}$ is an upper bound of the $\ell_1$-norm of core tensors and weight matrices of neural networks. Define the following coupled low-rank functions for $L$ continuous groups:
	\begin{equation}
		\begin{split}
		s_l({\bf v}):={\mathcal C}_l\times_1 &f_{\theta_1}({\bf v}_{(1)})\times_2\cdots\times_{N+1} f_{\theta_{N+1}}({\bf v}_{(N+1)}):\\&\;\;\;\;\;\;\;\;\;\;\;\;\;\;\;\;\;{\mathbb R}^{N+1}\rightarrow {\mathbb R},\;l=1,2,\cdots,L.
		\end{split}
	\end{equation}
	Then the following inequality is true for any two groups $l_1,l_2$ $\in\{1,2,\cdots,L\}$, any dimension $d\in\{1,2,\cdots,N+1\}$, and any two coordinates $v_d',v_d''\in{\mathbb R}:$
	\begin{equation}\label{ineq}
		\begin{split}
		|s_{l_1}(v_1,v_2,\cdots,&v_d',\cdots,v_{N+1})-s_{l_2}(v_1,v_2,\cdots,\\&v_d'',\cdots,v_{N+1})|\leq\delta_1 |v_d'-v_d''| + \delta_2,
		\end{split}
	\end{equation} 
	where 
	\begin{equation}
		\begin{split}
	&\delta_1=\eta^{MN+M+1}\omega^{(M-1)(N+1)}\xi^N,\\ &\delta_2=2\eta^{MN+M+1}\omega^{(M-1)(N+1)}\xi^{N+1},\\
	&\xi = \max\{|v_1|,\cdots,|v_d'|,|v_d''|,\cdots,|v_{N+1}|\}. 
	\end{split}
		\end{equation}
When $l_1=l_2$, the upper bound reduces to ${\delta_1 |v_d'-v_d''|}$.
\end{lemma}
From the inequality (\ref{ineq}), we can see that any two coupled low-rank functions $s_{l_1}(\cdot),s_{l_2}(\cdot)$ with shared factor functions mutually enjoy a Lipschitz smooth-type correlation, i.e., the difference of function values is bounded by the difference of input coordinates plus a constant. This inequality reveals that our CRNL characterizes the {similarity} across different continuous groups. When considering the internal correlation of the same low-rank function, i.e., when $l_1=l_2$, the constant can be further omitted, i.e., the internal similarity inside a group is tighter than the similarity across different groups. This reveals that our CRNL also respects the {individuality} of each continuous group, which mainly comes from the unshared core tensors. The encoded similarity intra- and inter-groups enables our method to more accurately characterize the structure relationships among different nonlocal groups, and hence are expected to enhance the effectiveness for multi-dimensional data processing.
\subsection{The Design of Regularization Term}\label{sec_reg}
Next, we introduce the regularization terms in the optimization models (\ref{model_INR}) and (\ref{model_NL}). In the optimization model (\ref{model_INR}), we consider the simple energy regularization to constrain the neural network weights $\theta$ for multivariate regression problems, i.e., $\psi[f_\theta(\cdot)] = \lVert\theta\rVert_F^2$. The energy regularization can be conveniently controlled by the weight decay parameter of the Adam optimizer in PyTorch. For image recovery problems (inpainting and denoising), we consider the same energy regularization plus a total variation (TV) regularization conditioned on the recovered image\footnote{Here, the recovered image $\cal X$ can be readily obtained by inferring the function values of $f_\theta(\cdot)$ on the image meshgrid positions.} ${\cal X}\in{\mathbb R}^{n_1\times n_2\times n_3}$, i.e., $\psi[f_\theta(\cdot)] = \lVert\theta\rVert_F^2+\gamma\lVert{\cal X}\rVert_{\rm TV}$, where $\lVert{\cal X}\rVert_{\rm TV}:=(\sum_{i=1}^{n_1-1}|{\cal X}_{(i,:,:)}-{\cal X}_{(i+1,:,:)}|+\sum_{j=1}^{n_2-1}|{\cal X}_{(:,j,:)}-{\cal X}_{(:,j+1,:)}|)$ and $\gamma$ is a trade-off parameter.\par  
In the CRNL optimization model (\ref{model_NL}), we consider the same energy regularization for multivariate regression problems to constrain the core tensor ${\cal C}_l$ and weights of factor functions, i.e., $\psi[s_l(\cdot)] = \lVert{\cal C}_l\rVert_F^2+\sum_{d=1}^{N+1}\lVert\theta_d\rVert_F^2$. For image recovery problems (inpainting and denoising), we consider the energy regularization plus a TV regularization conditioned on the recovered image\footnote{Similarly, the recovered image of CRNL can be readily obtained by inferring the function values of $\{s_l(\cdot)\}_{l=1}^L$ on meshgrid points and then rearranging the output image patches to their original positions.} ${\cal X}\in{\mathbb R}^{n_1\times n_2\times n_3}$, i.e., $\psi[s_l(\cdot)] = \lVert{\cal C}_l\rVert_F^2+\sum_{d=1}^{N+1}\lVert\theta_d\rVert_F^2+\gamma\lVert{\cal X}\rVert_{\rm TV}$, where $\gamma$ is a trade-off parameter. The TV preserves the spatial local smoothness of the recovered image to enhance the recovery performance.
\begin{figure*}[t]
	\scriptsize
	\setlength{\tabcolsep}{0.9pt}
	\begin{center}
		\begin{tabular}{ccccccccc}
\includegraphics[width=0.106\textwidth]{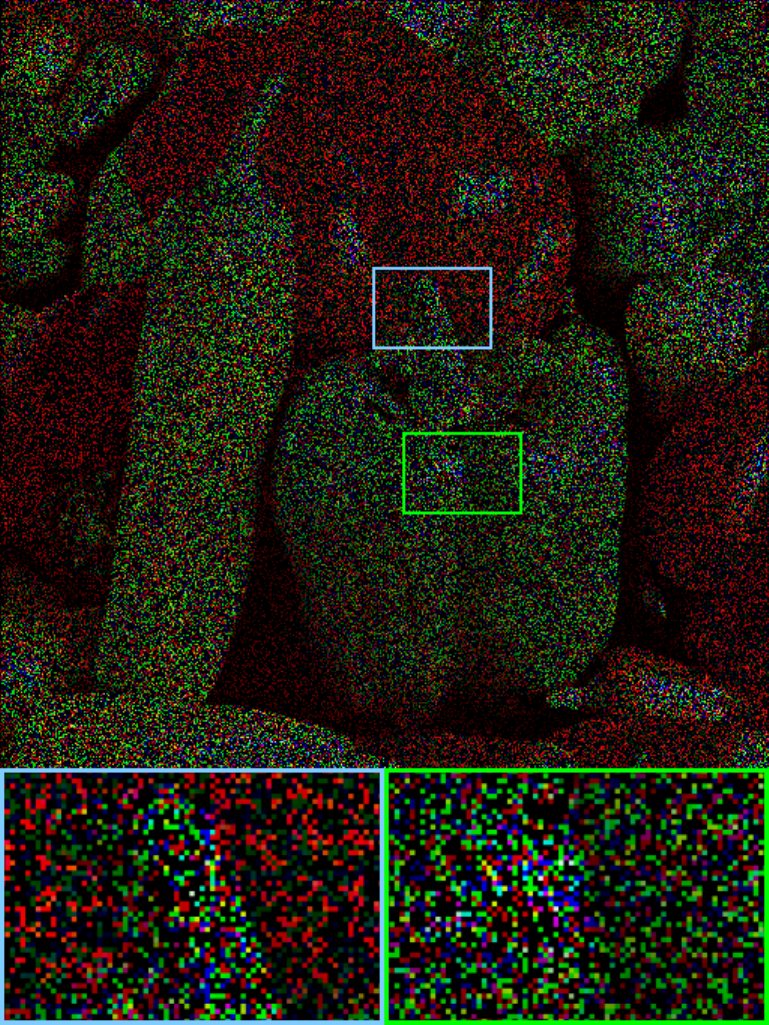}&
\includegraphics[width=0.106\textwidth]{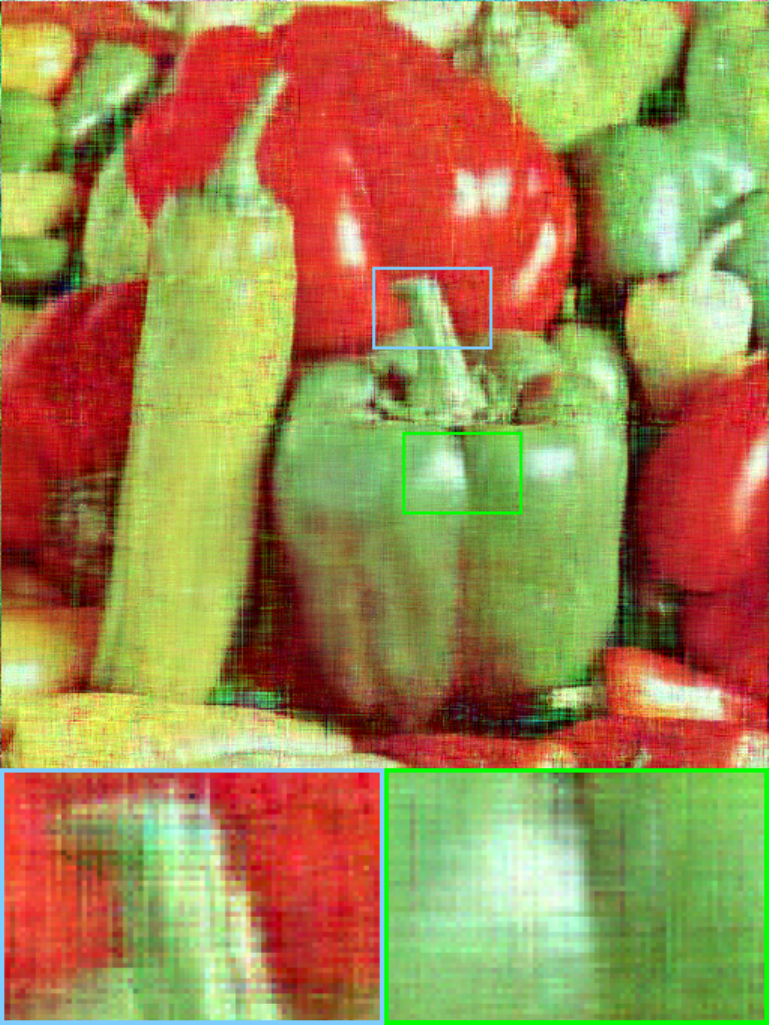}&
\includegraphics[width=0.106\textwidth]{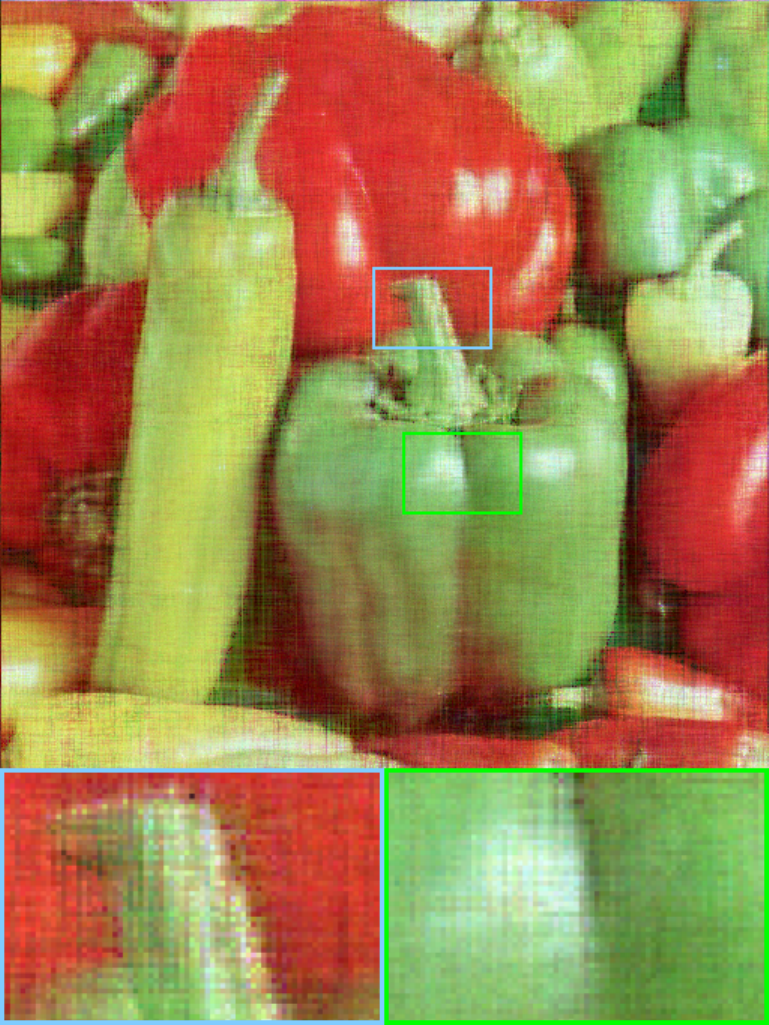}&
\includegraphics[width=0.106\textwidth]{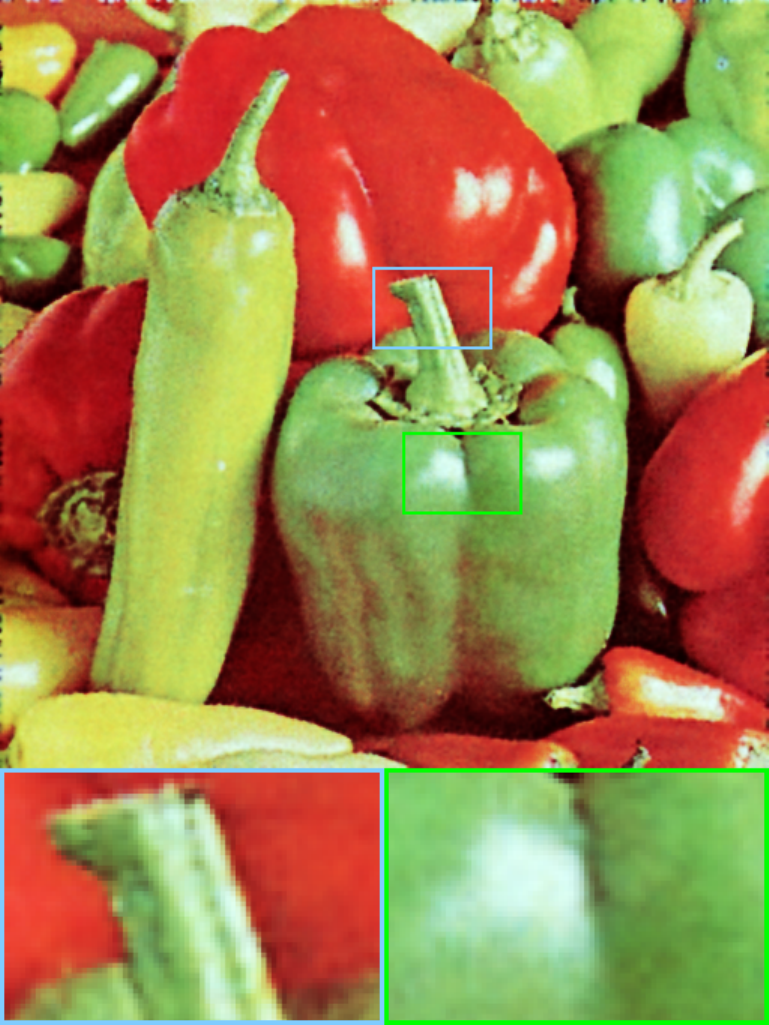}&
\includegraphics[width=0.106\textwidth]{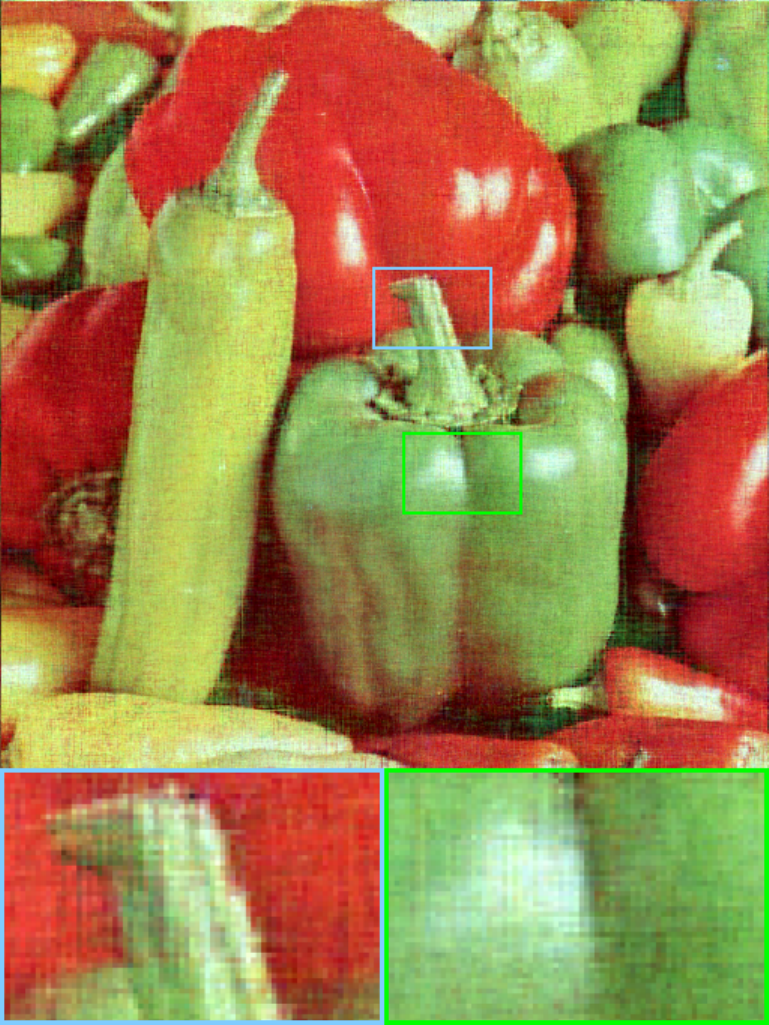}&
\includegraphics[width=0.106\textwidth]{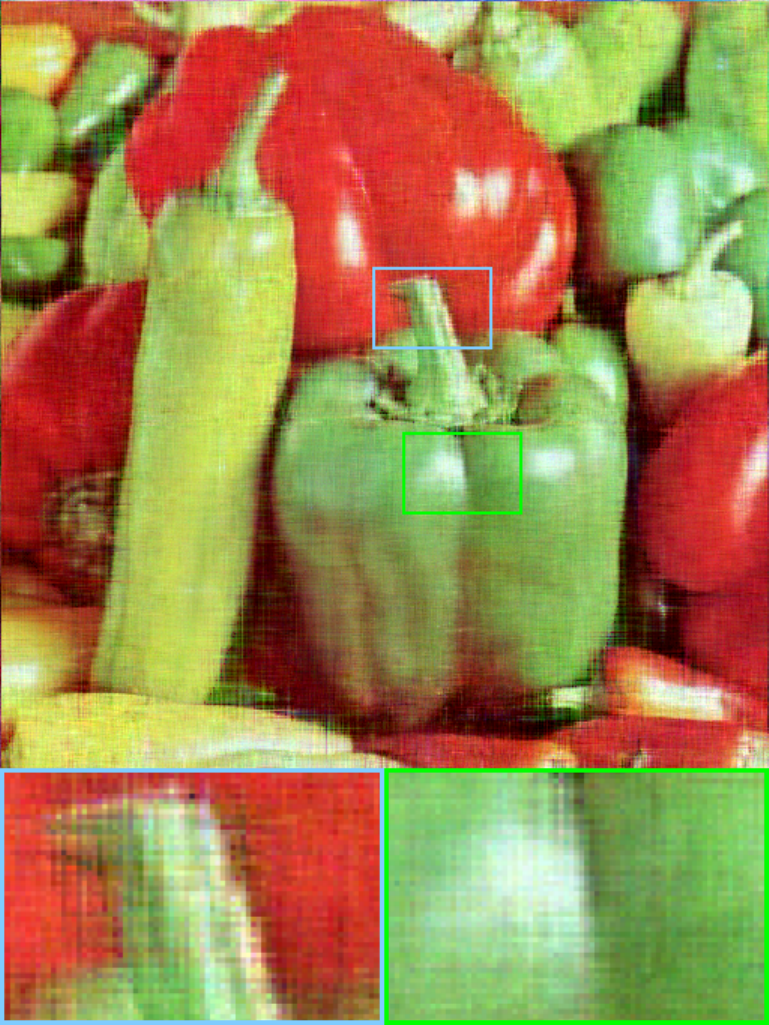}&
\includegraphics[width=0.106\textwidth]{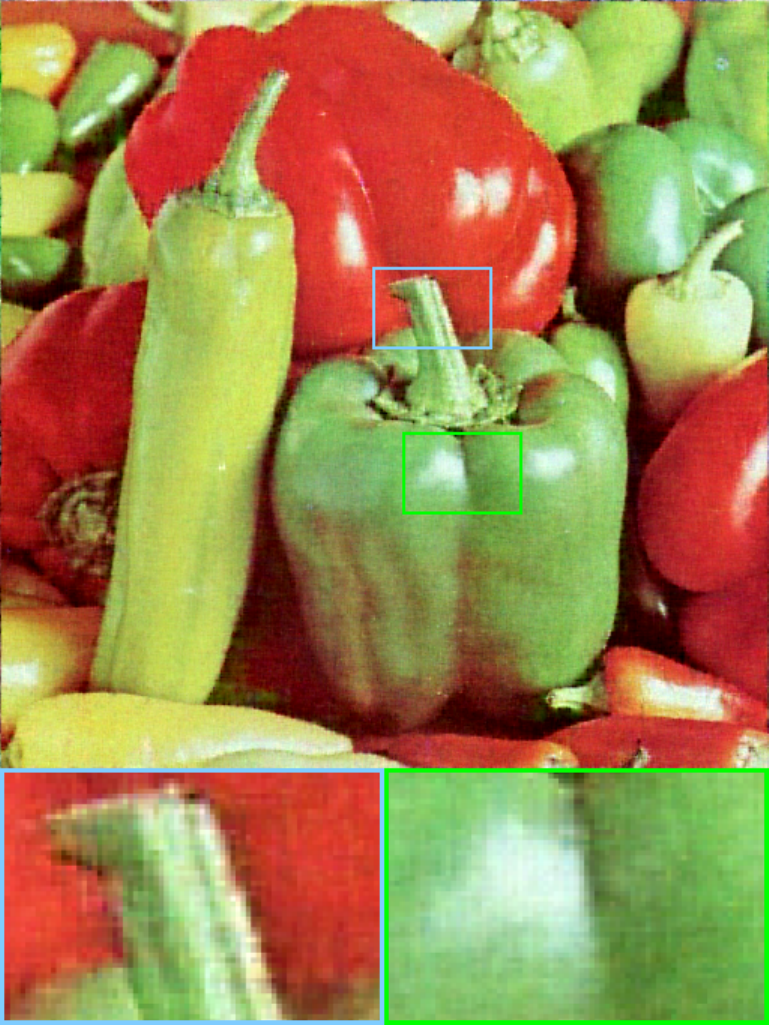}&
\includegraphics[width=0.106\textwidth]{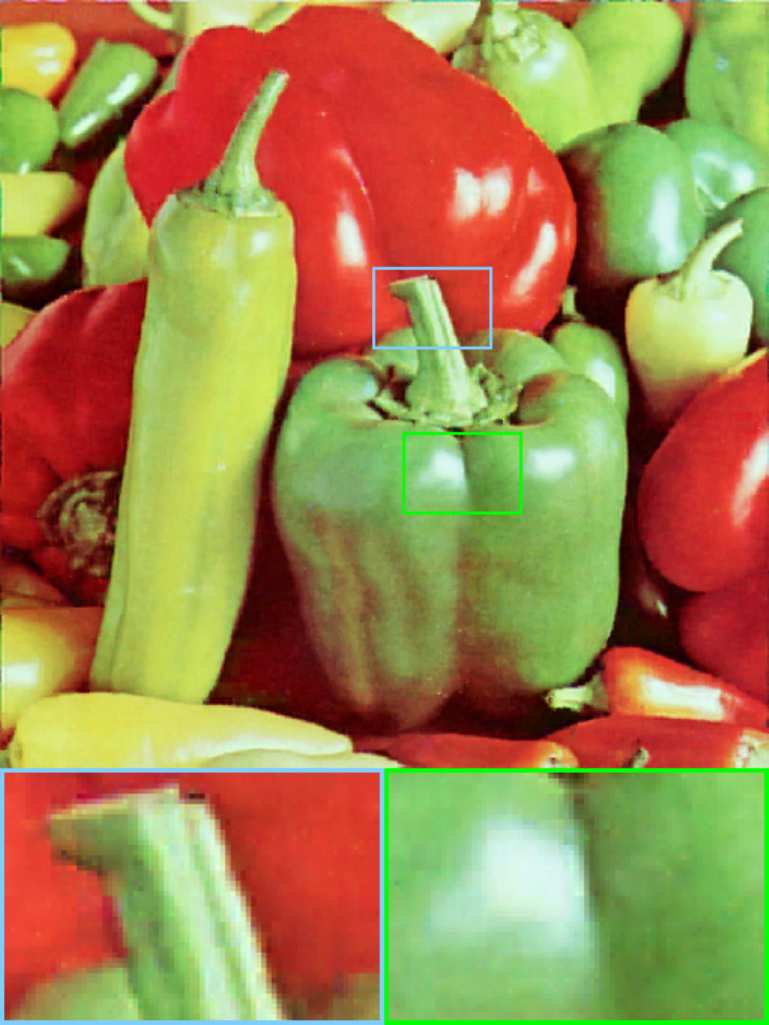}&
\includegraphics[width=0.106\textwidth]{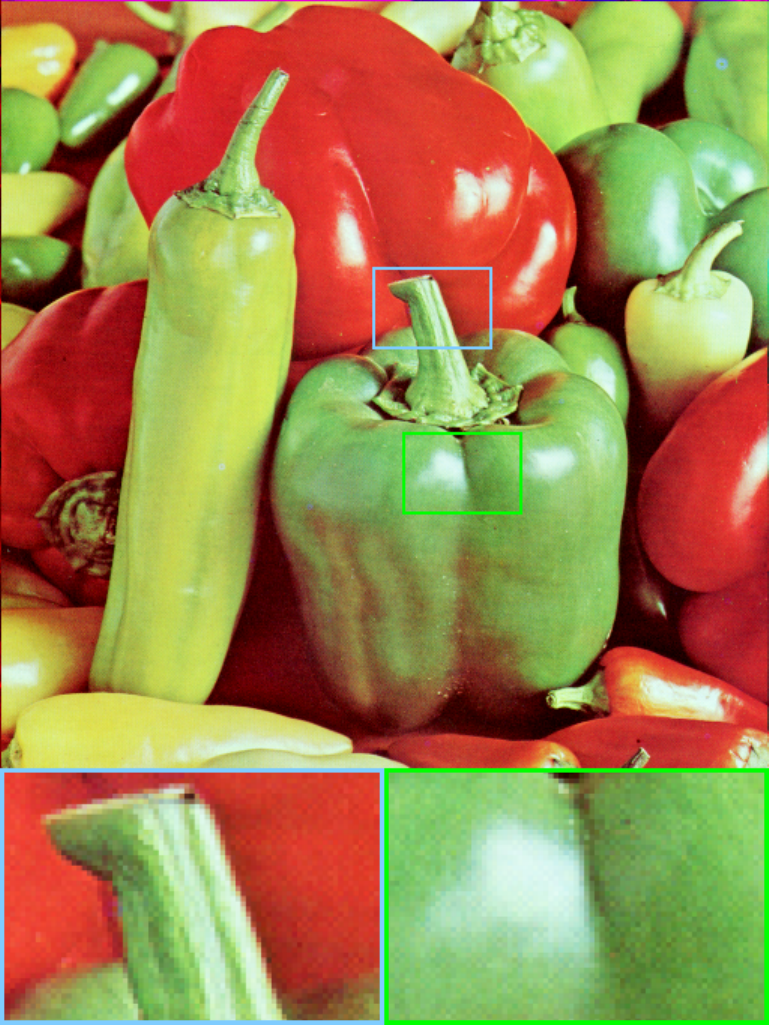}\\
PSNR 7.52 dB &
PSNR 22.61 dB &
PSNR 22.84 dB &
PSNR 27.80 dB &
PSNR 24.53 dB &
PSNR 23.94 dB &
PSNR 26.77 dB &
PSNR 28.39 dB &
PSNR Inf\\
\includegraphics[width=0.106\textwidth]{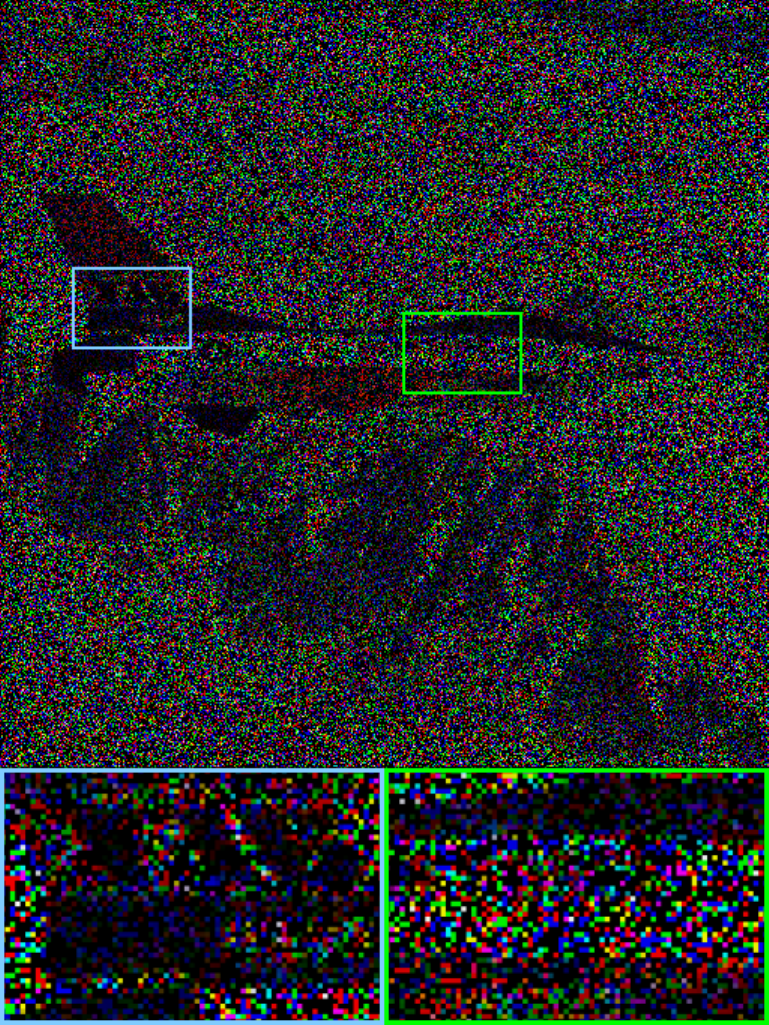}&
\includegraphics[width=0.106\textwidth]{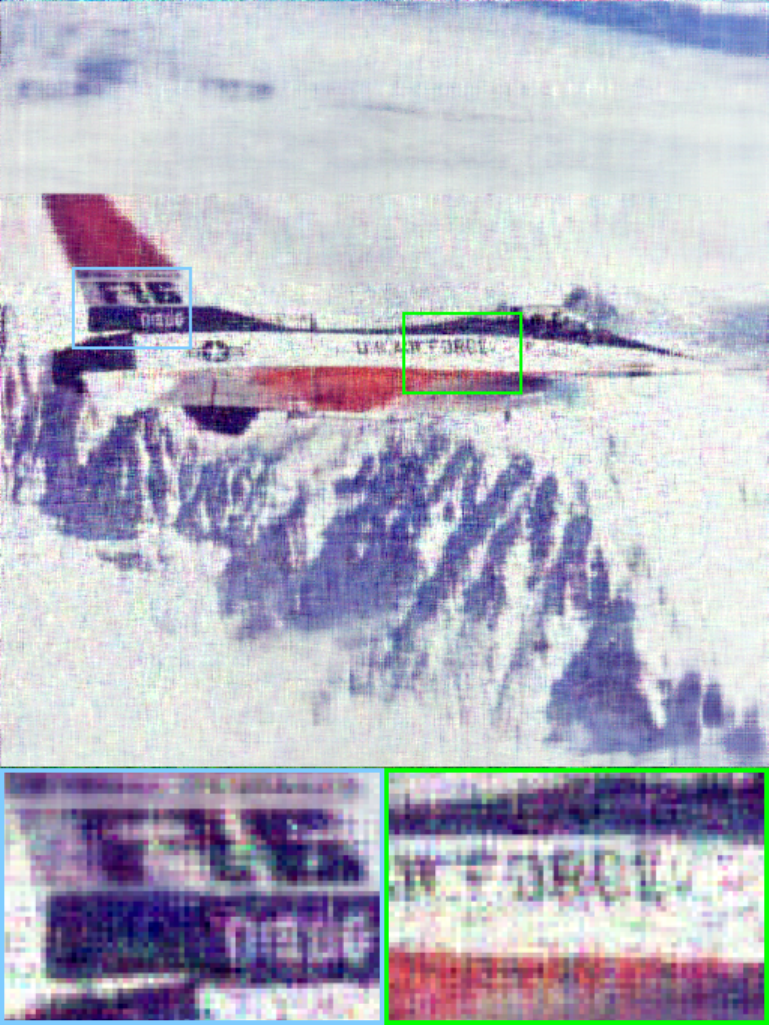}&
\includegraphics[width=0.106\textwidth]{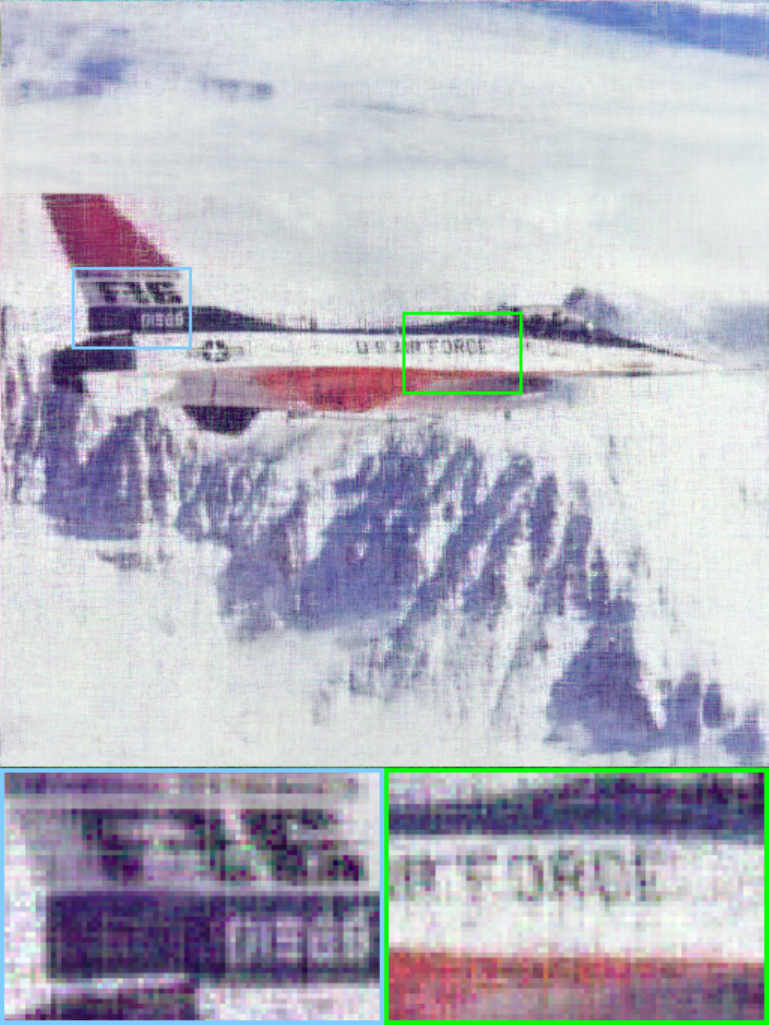}&
\includegraphics[width=0.106\textwidth]{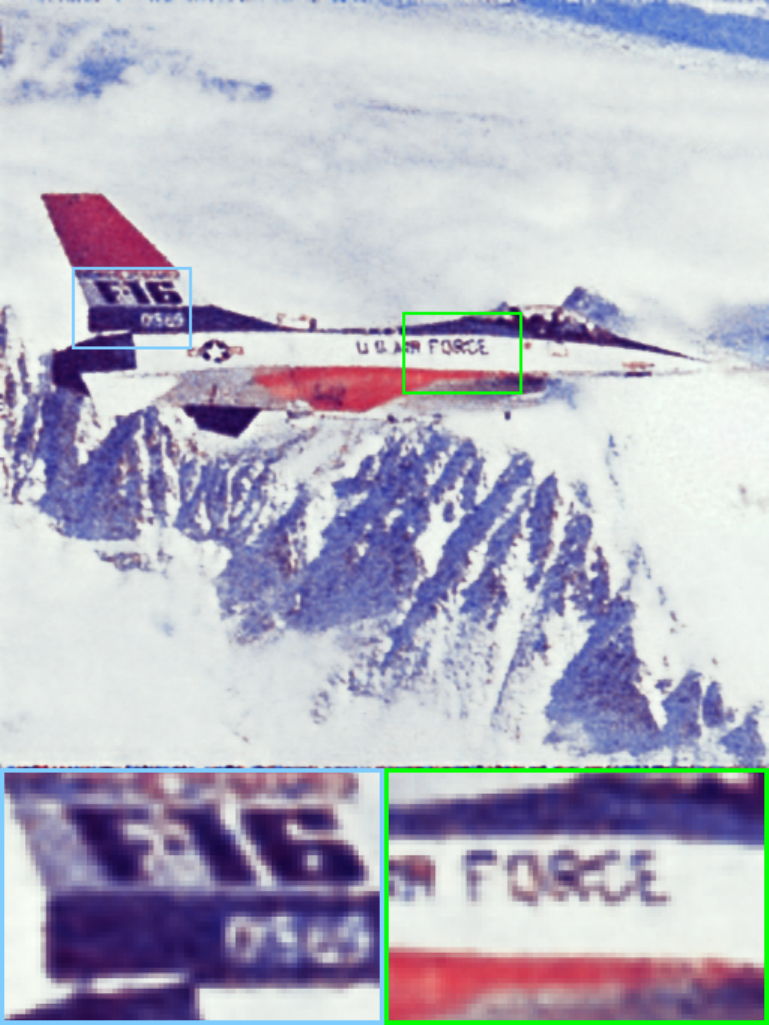}&
\includegraphics[width=0.106\textwidth]{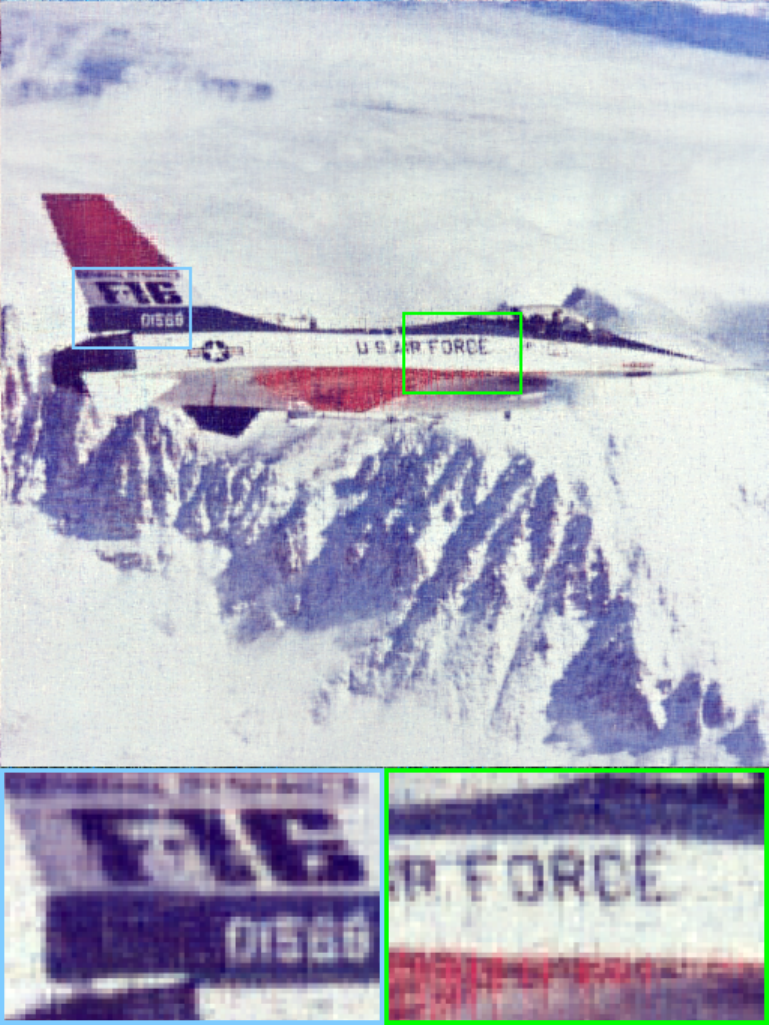}&
\includegraphics[width=0.106\textwidth]{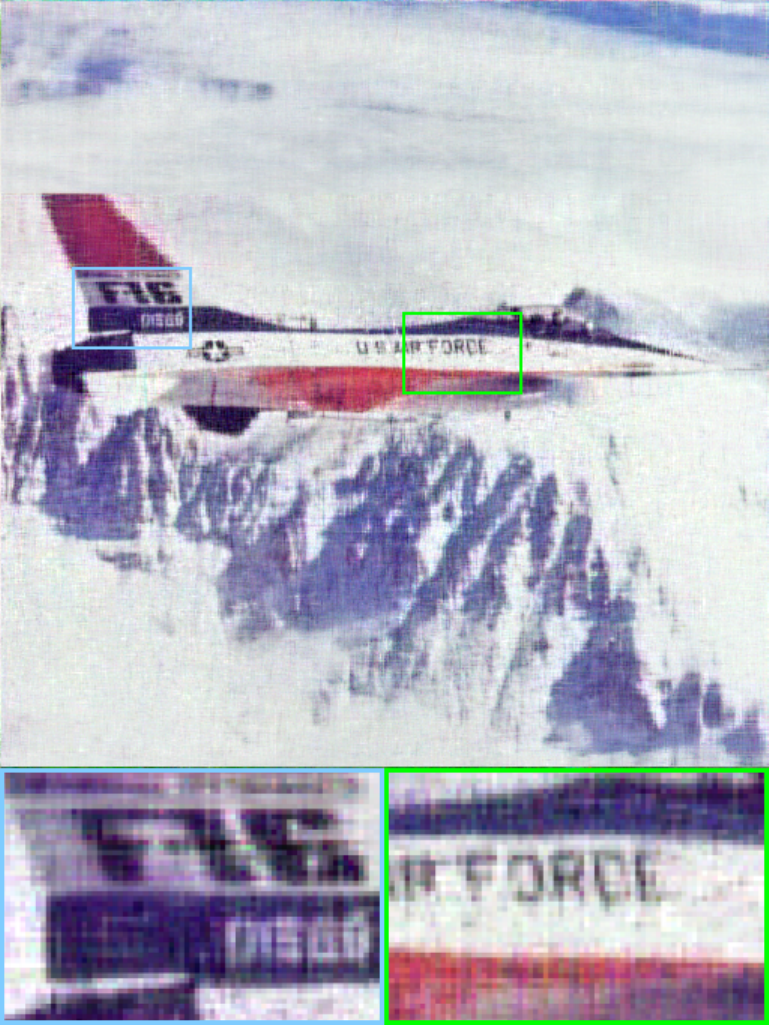}&
\includegraphics[width=0.106\textwidth]{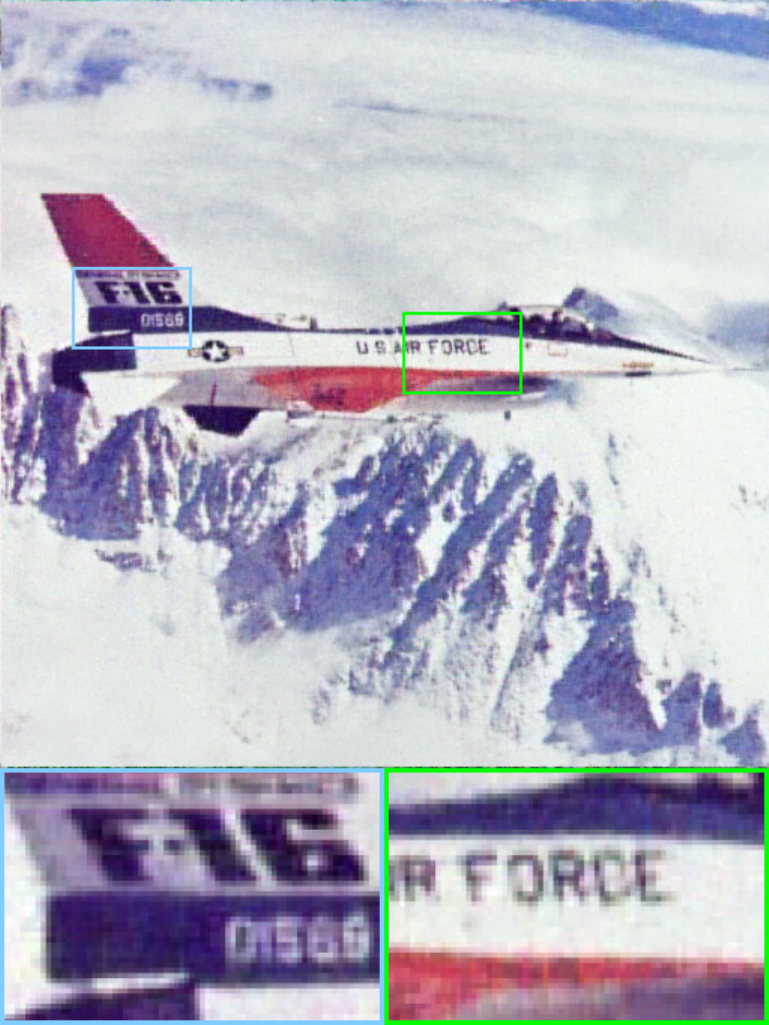}&
\includegraphics[width=0.106\textwidth]{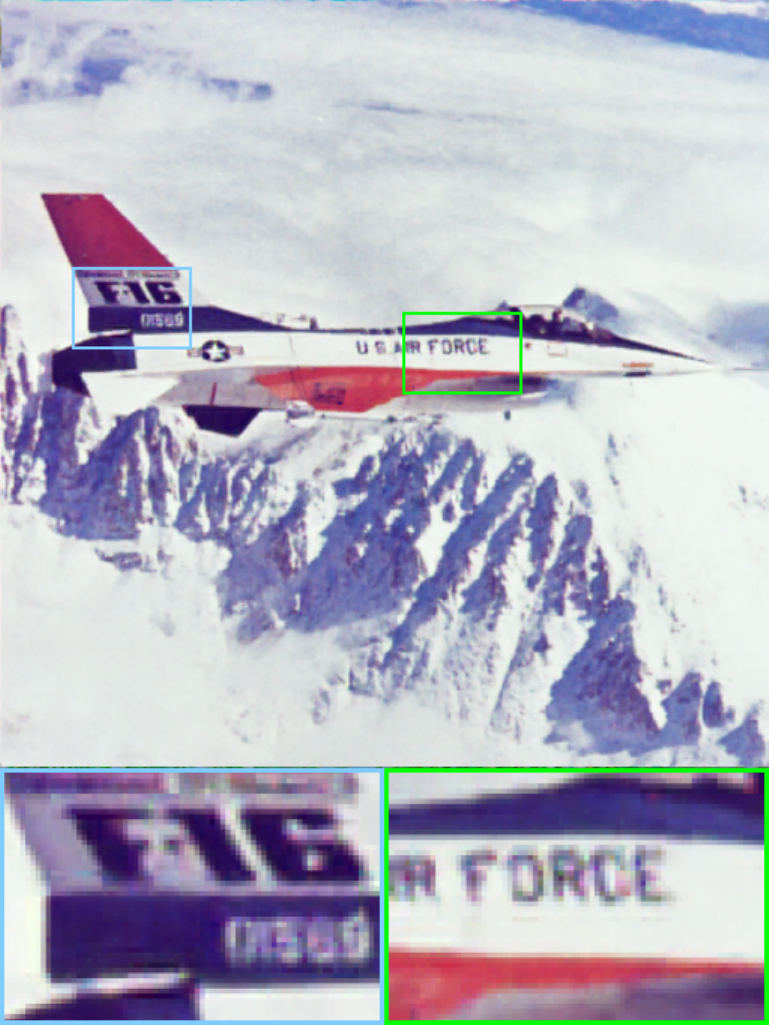}&
\includegraphics[width=0.106\textwidth]{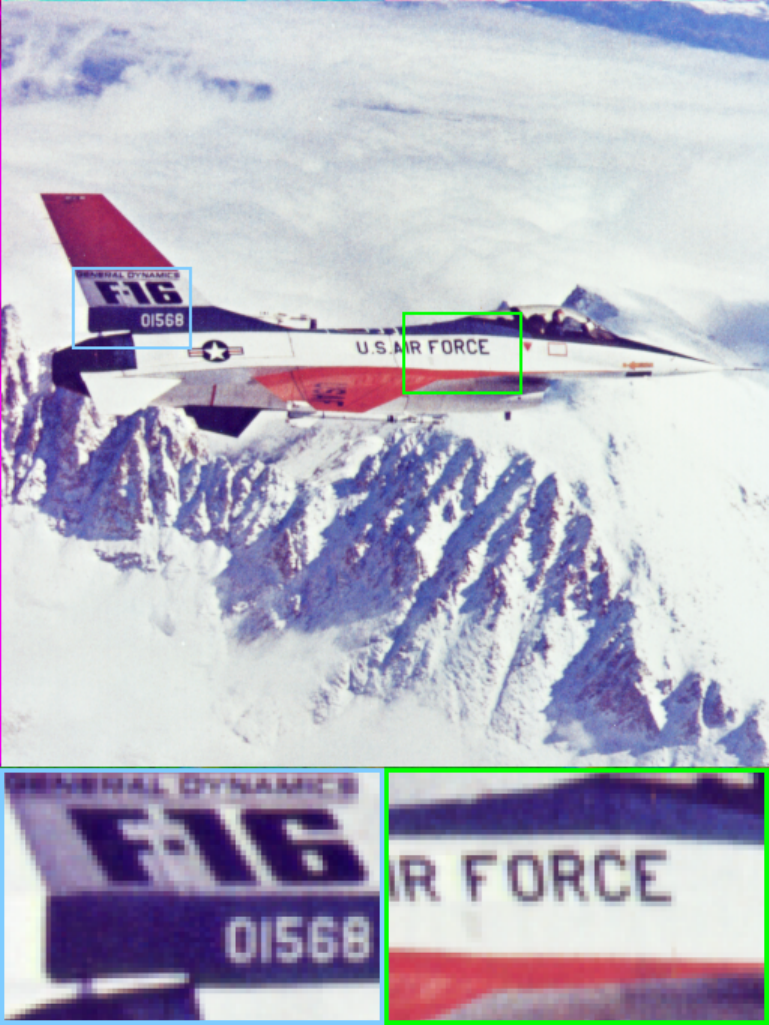}\\
PSNR 3.66 dB &
PSNR 24.88 dB &
PSNR 25.45 dB &
PSNR 28.37 dB &
PSNR 28.66 dB &
PSNR 26.34 dB &
PSNR 29.62 dB &
PSNR 30.77 dB &
PSNR Inf\\
\includegraphics[width=0.106\textwidth]{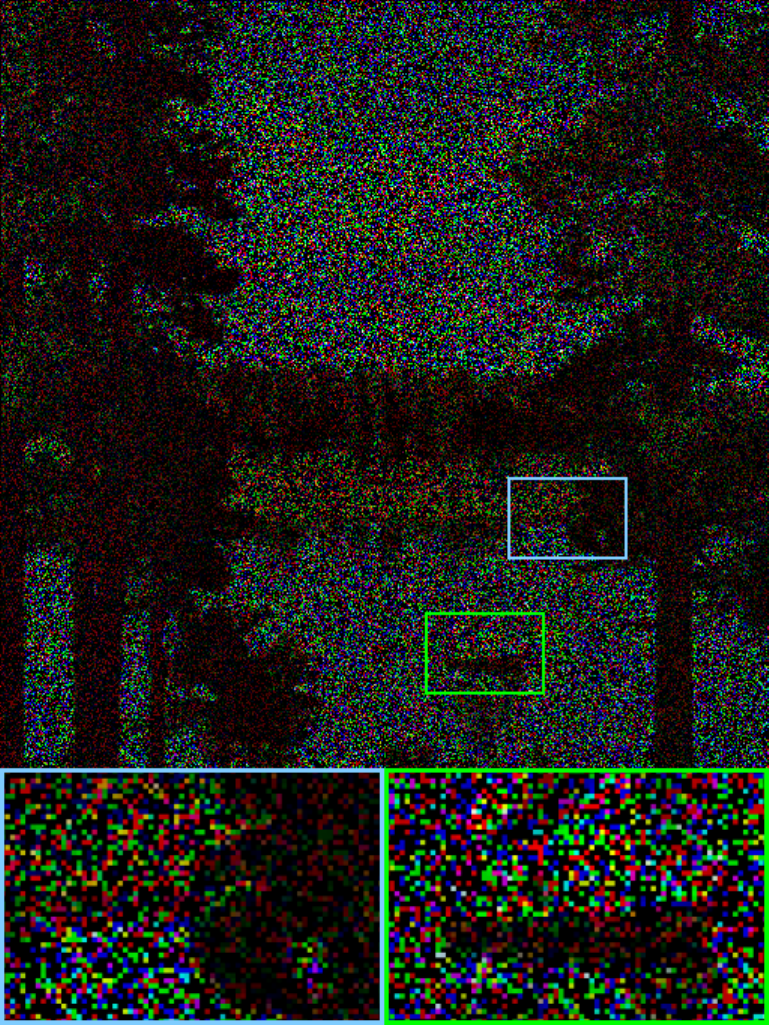}&
\includegraphics[width=0.106\textwidth]{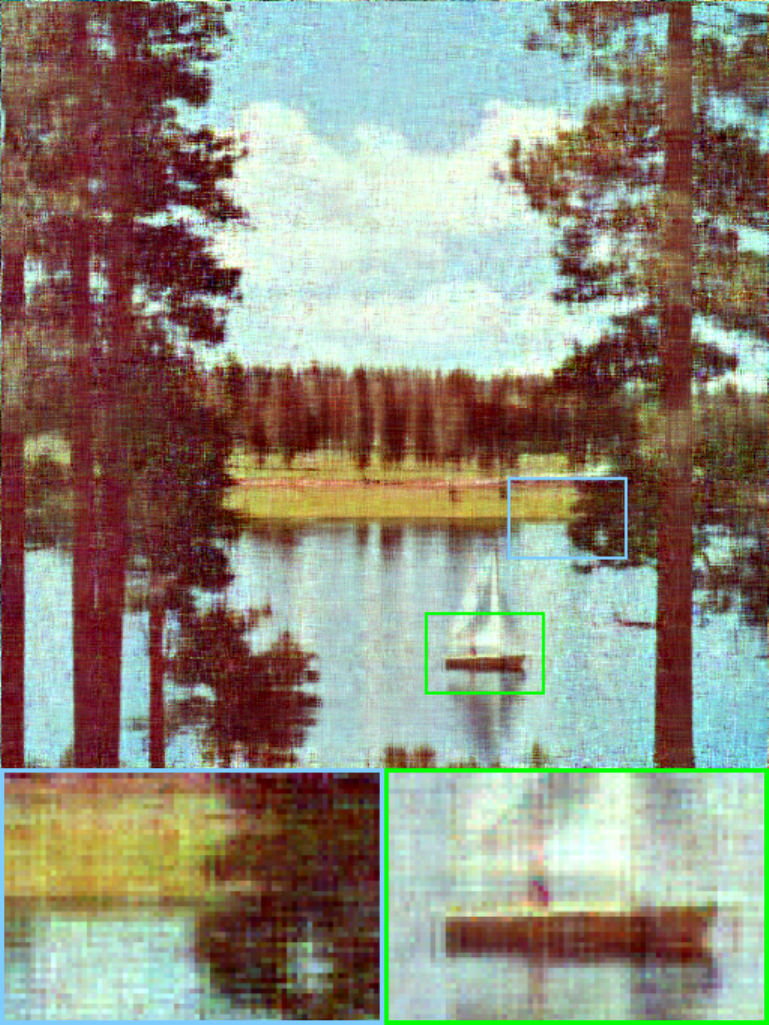}&
\includegraphics[width=0.106\textwidth]{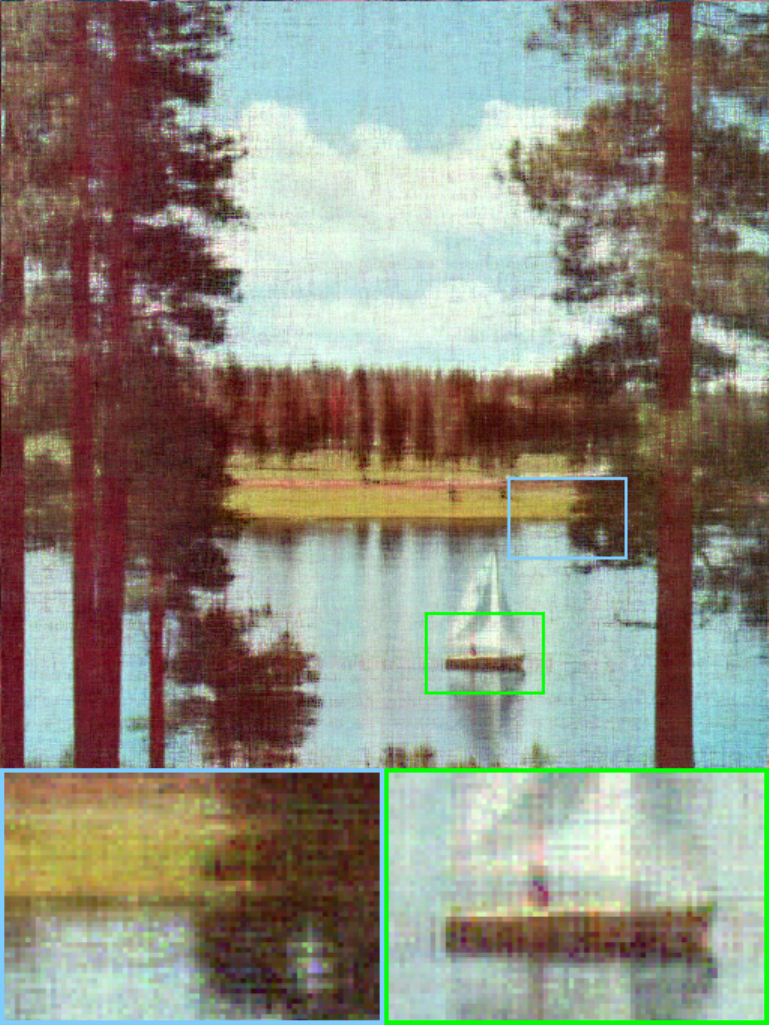}&
\includegraphics[width=0.106\textwidth]{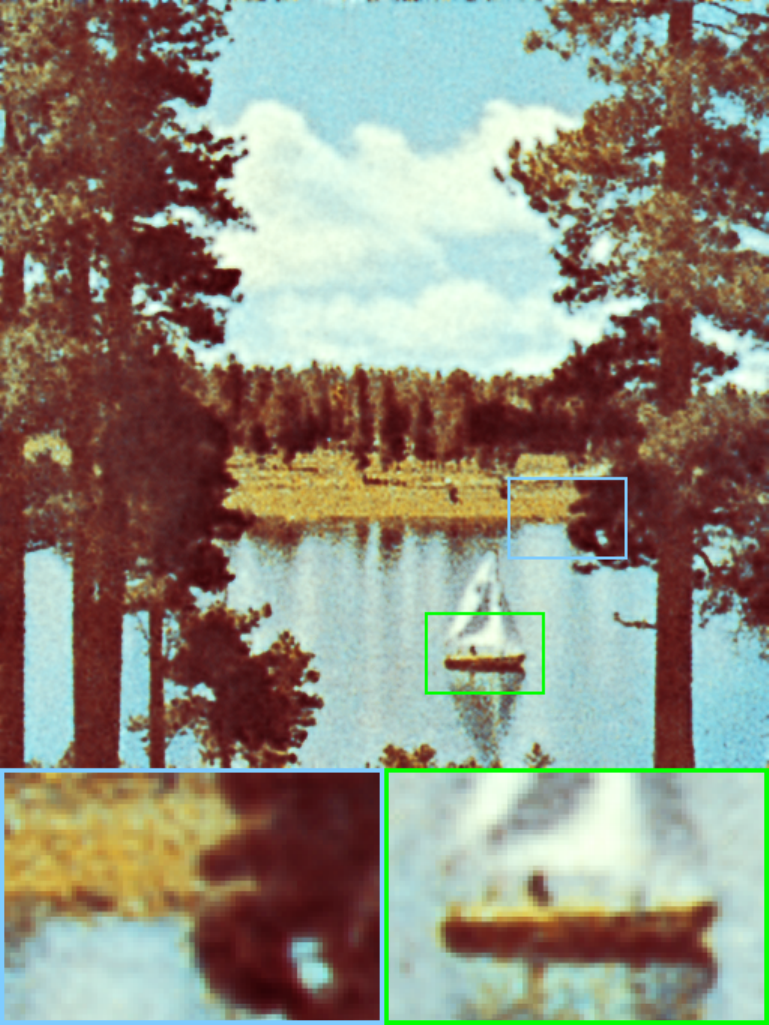}&
\includegraphics[width=0.106\textwidth]{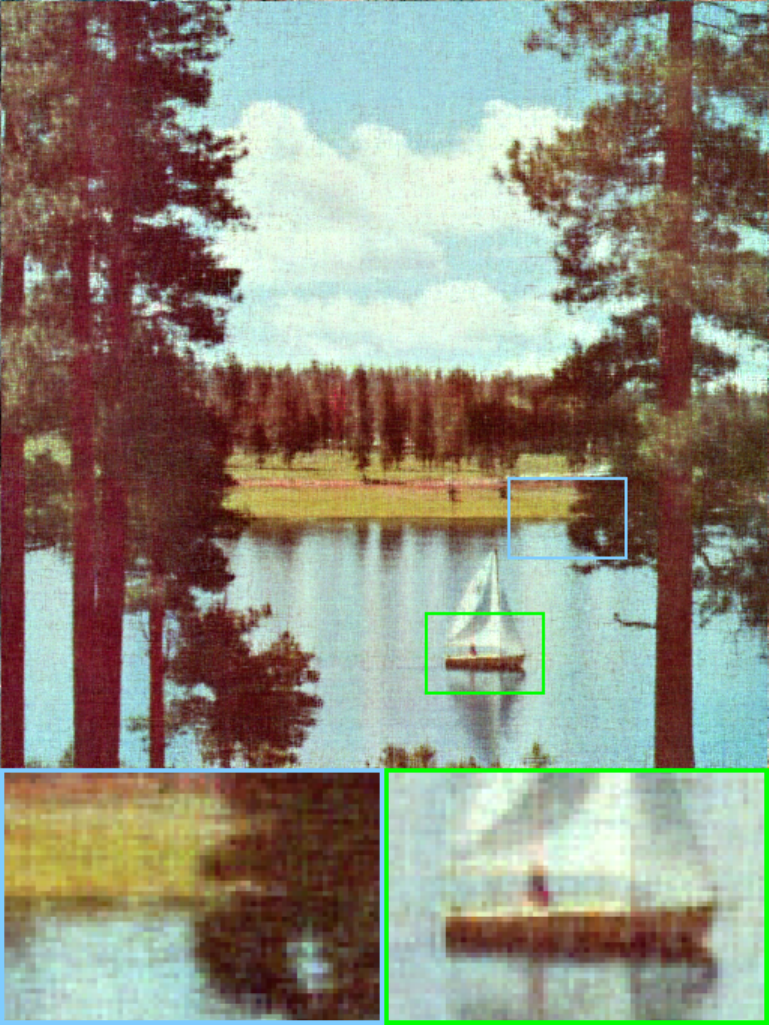}&
\includegraphics[width=0.106\textwidth]{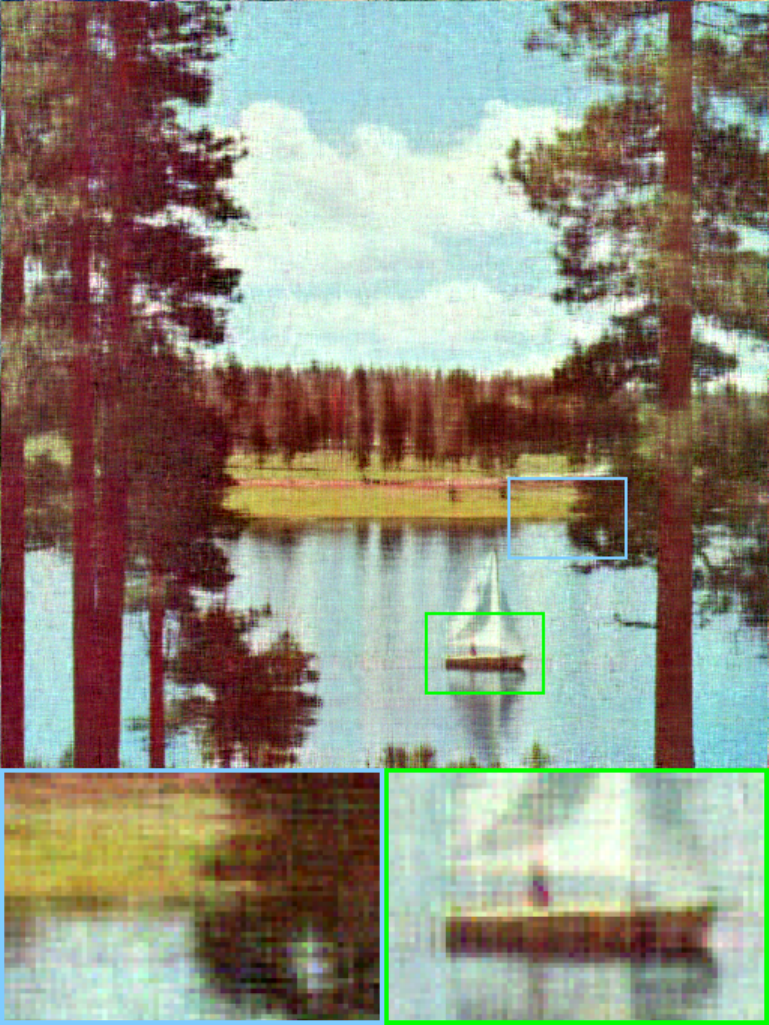}&
\includegraphics[width=0.106\textwidth]{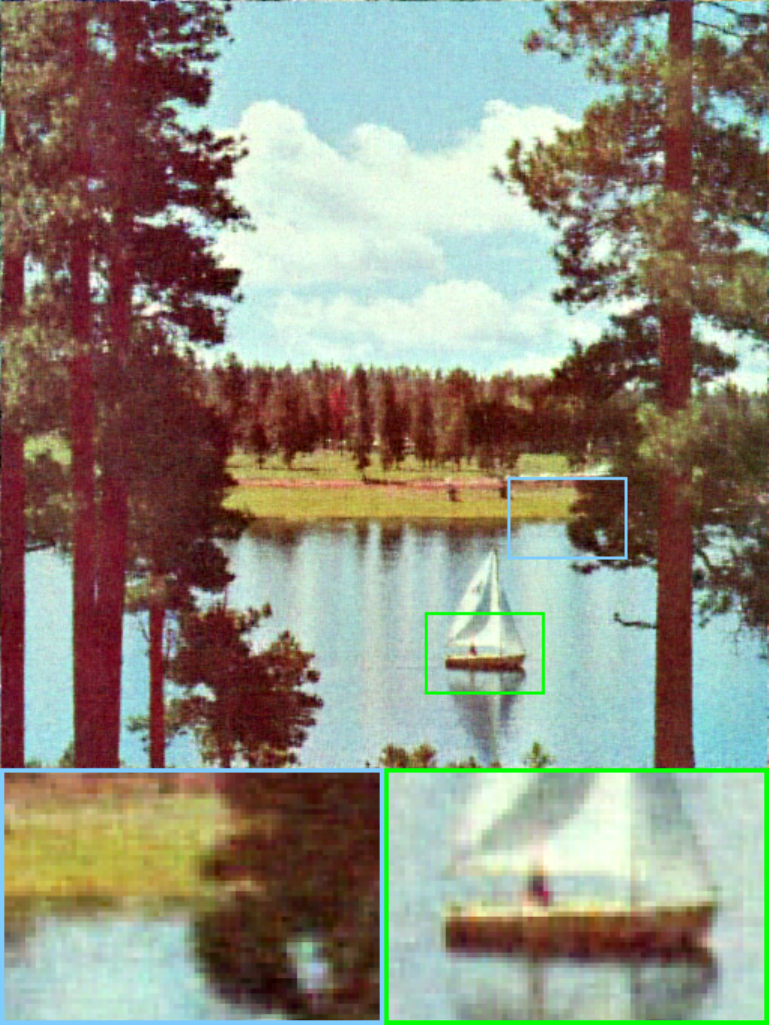}&
\includegraphics[width=0.106\textwidth]{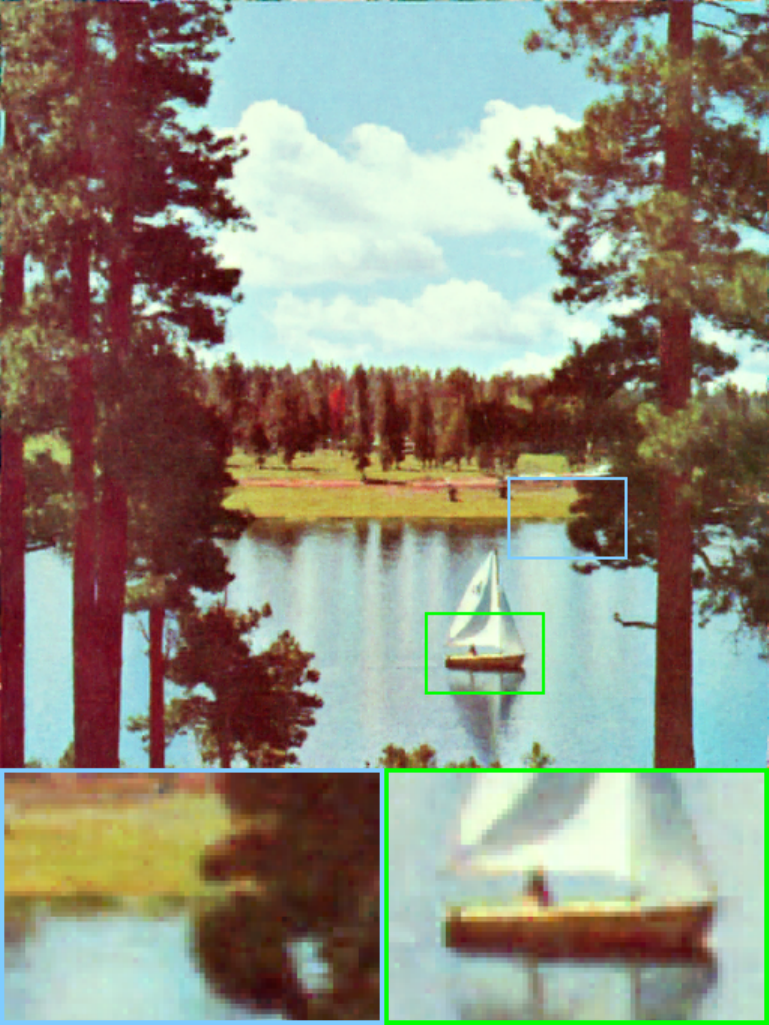}&
\includegraphics[width=0.106\textwidth]{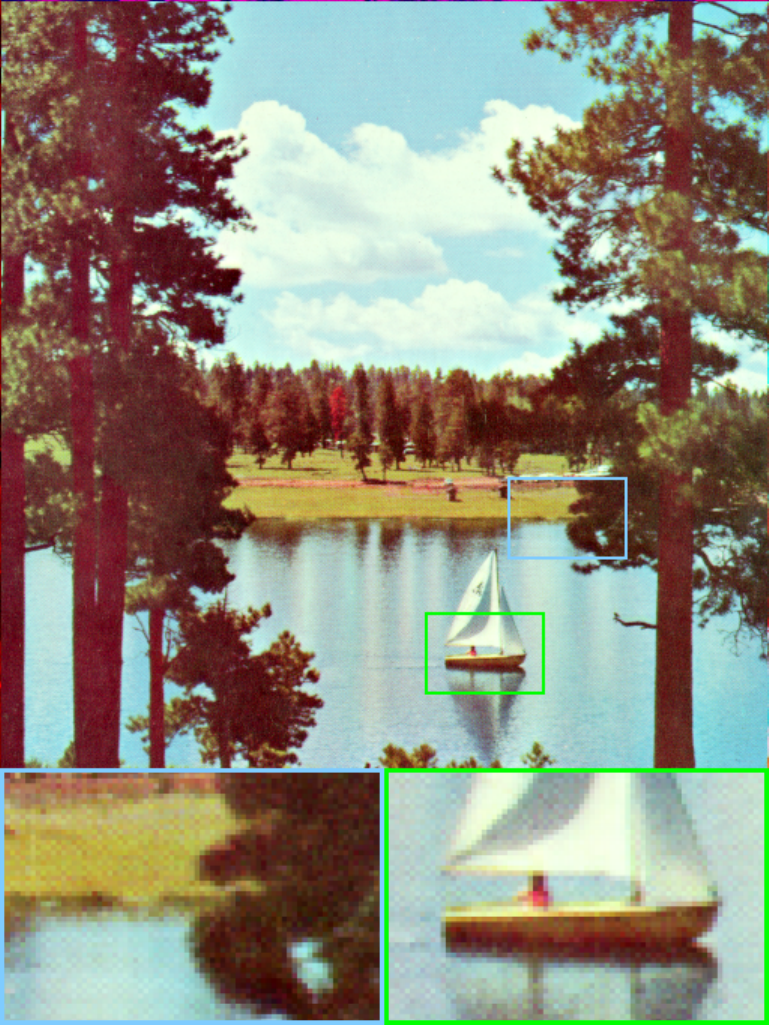}\\
PSNR 6.13 dB &
PSNR 21.78 dB &
PSNR 22.42 dB &
PSNR 24.92 dB &
PSNR 24.52 dB &
PSNR 22.87 dB &
PSNR 25.58 dB &
PSNR 26.64 dB &
PSNR Inf\\
\includegraphics[width=0.106\textwidth]{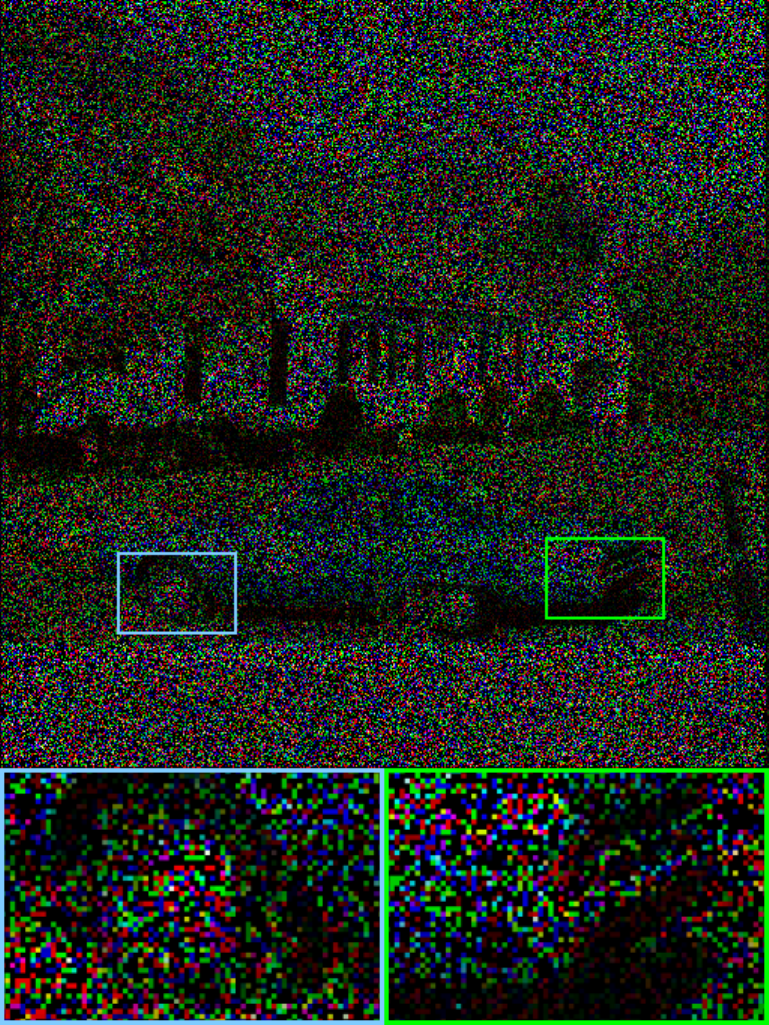}&
\includegraphics[width=0.106\textwidth]{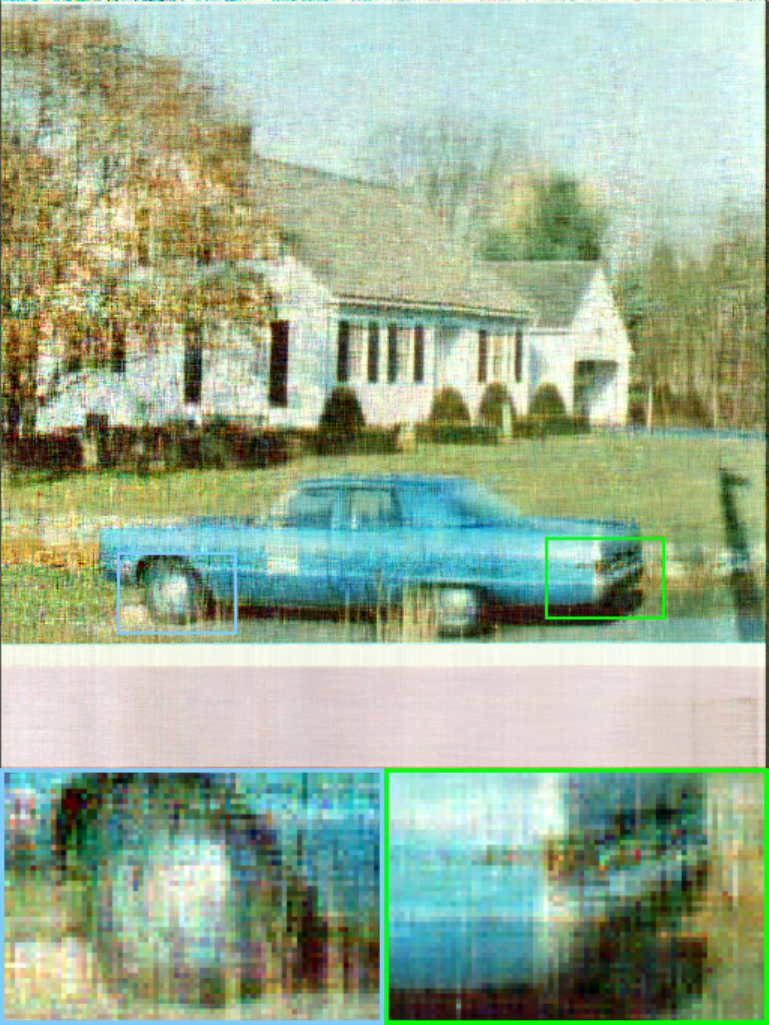}&
\includegraphics[width=0.106\textwidth]{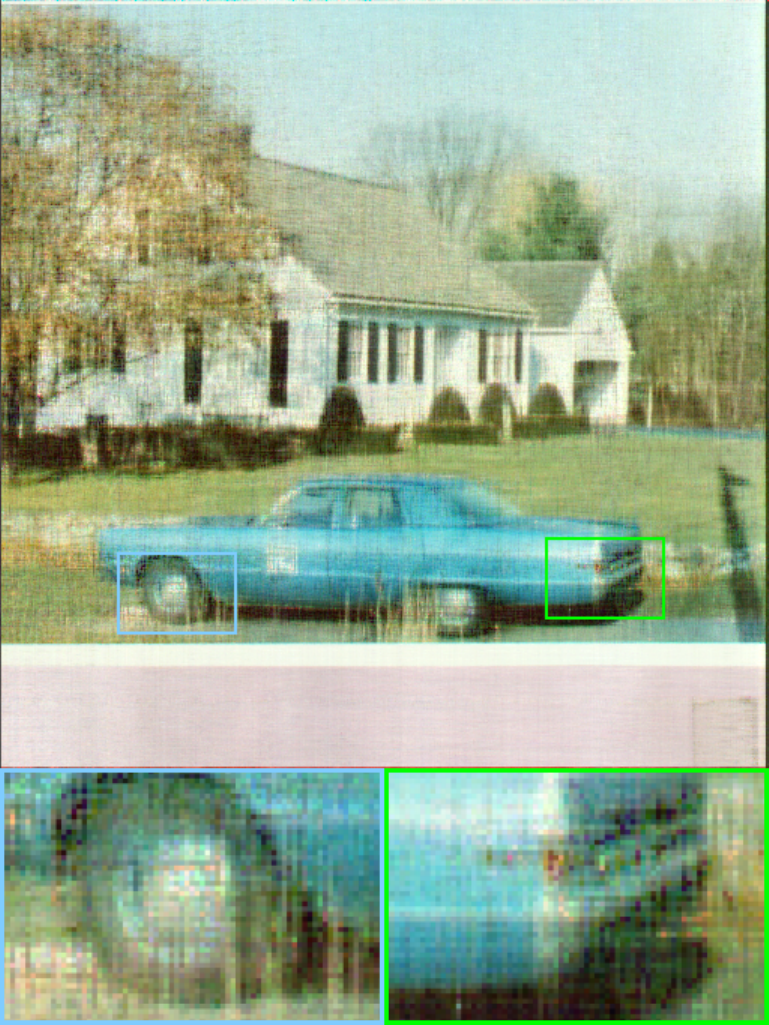}&
\includegraphics[width=0.106\textwidth]{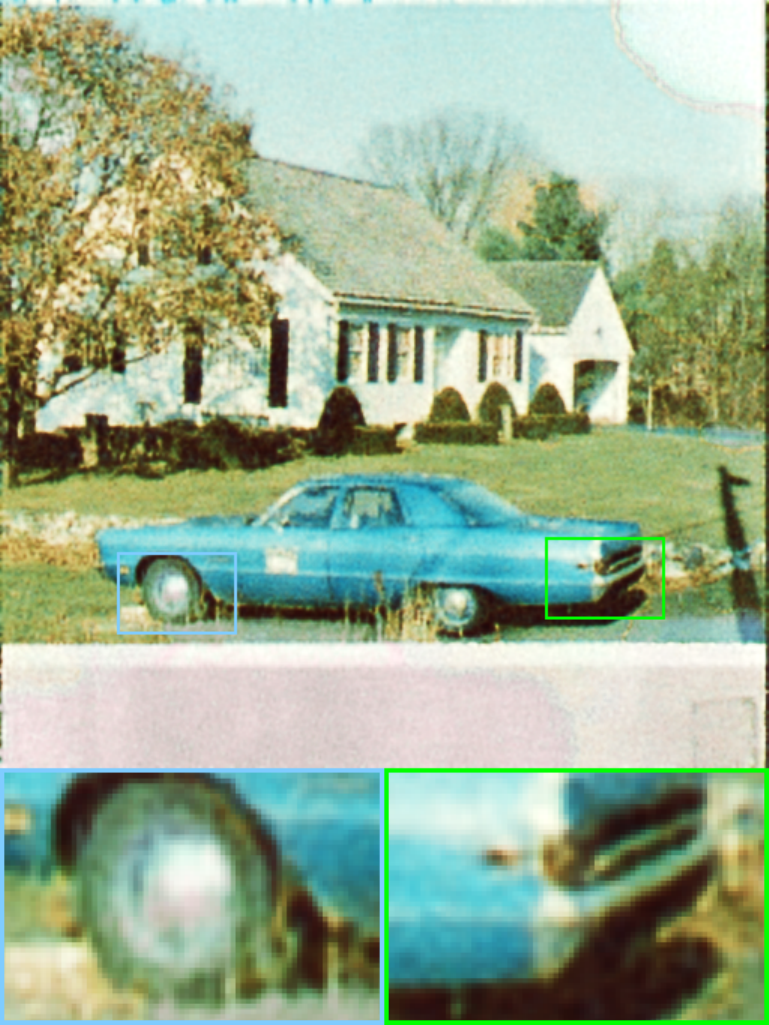}&
\includegraphics[width=0.106\textwidth]{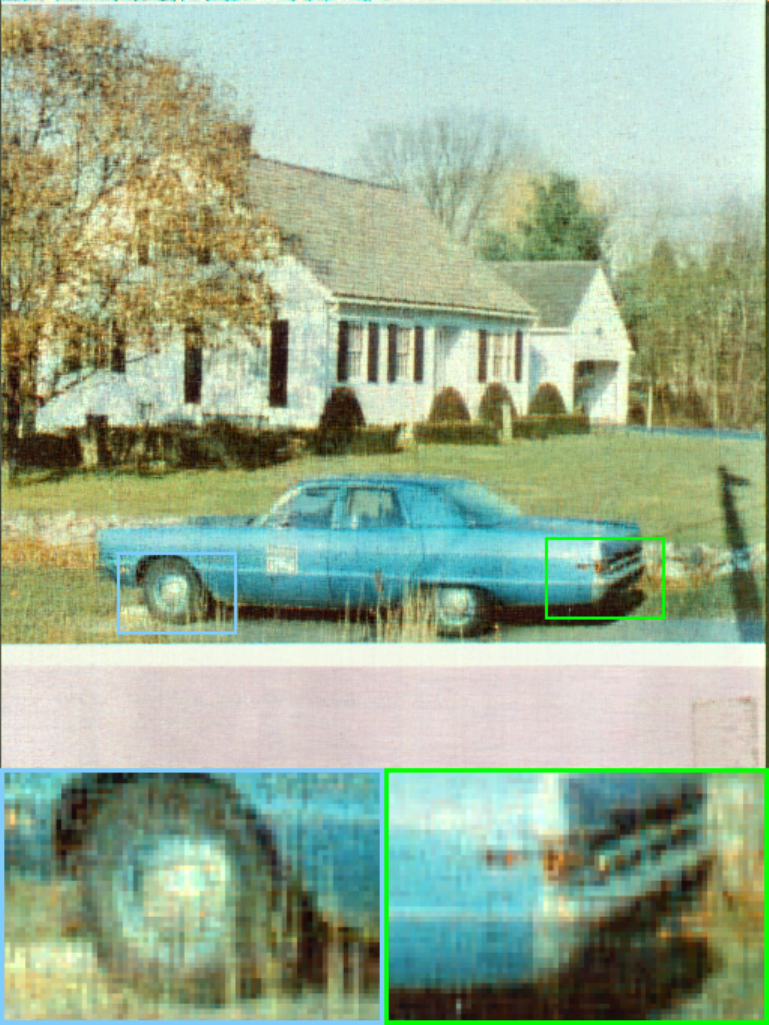}&
\includegraphics[width=0.106\textwidth]{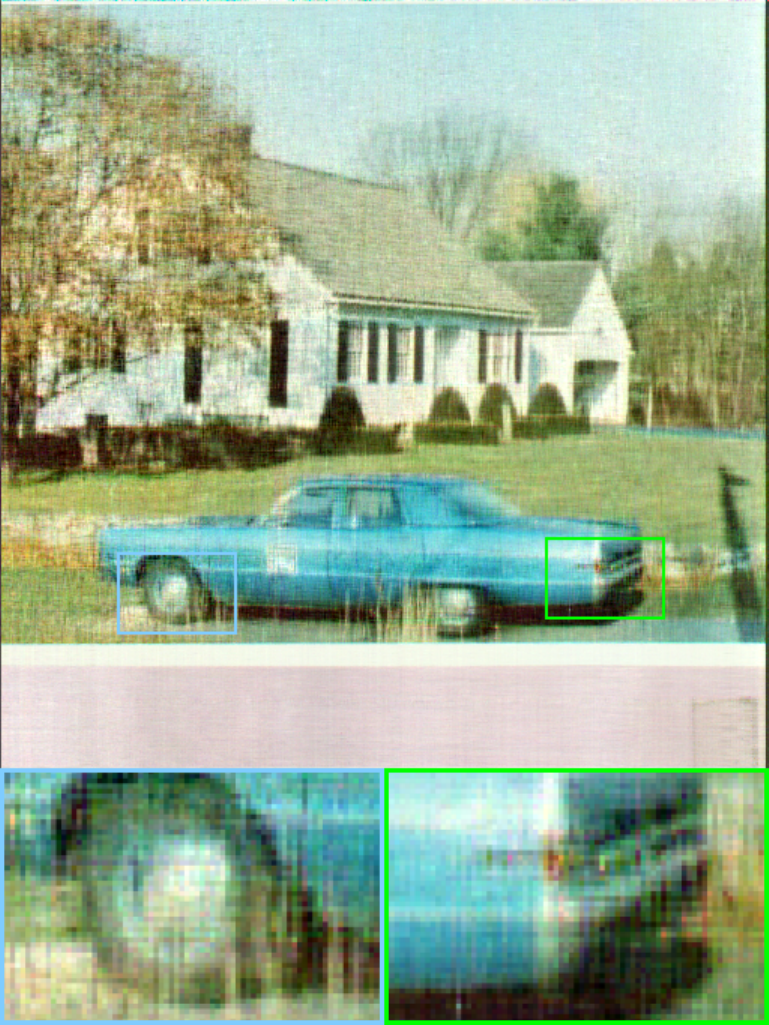}&
\includegraphics[width=0.106\textwidth]{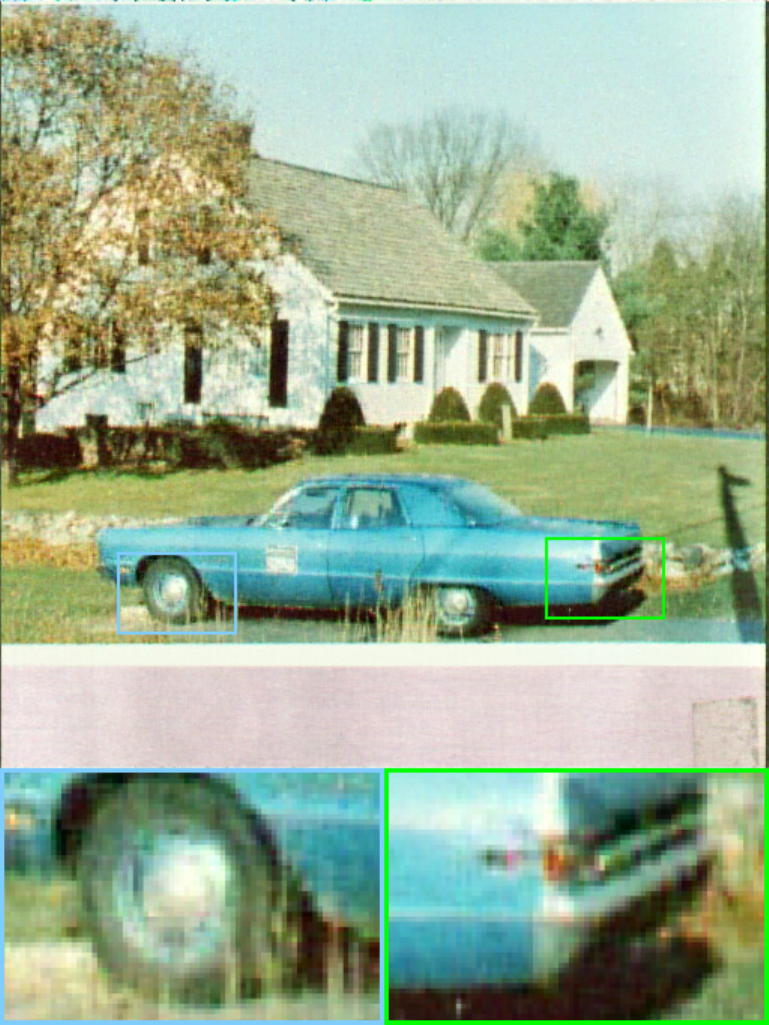}&
\includegraphics[width=0.106\textwidth]{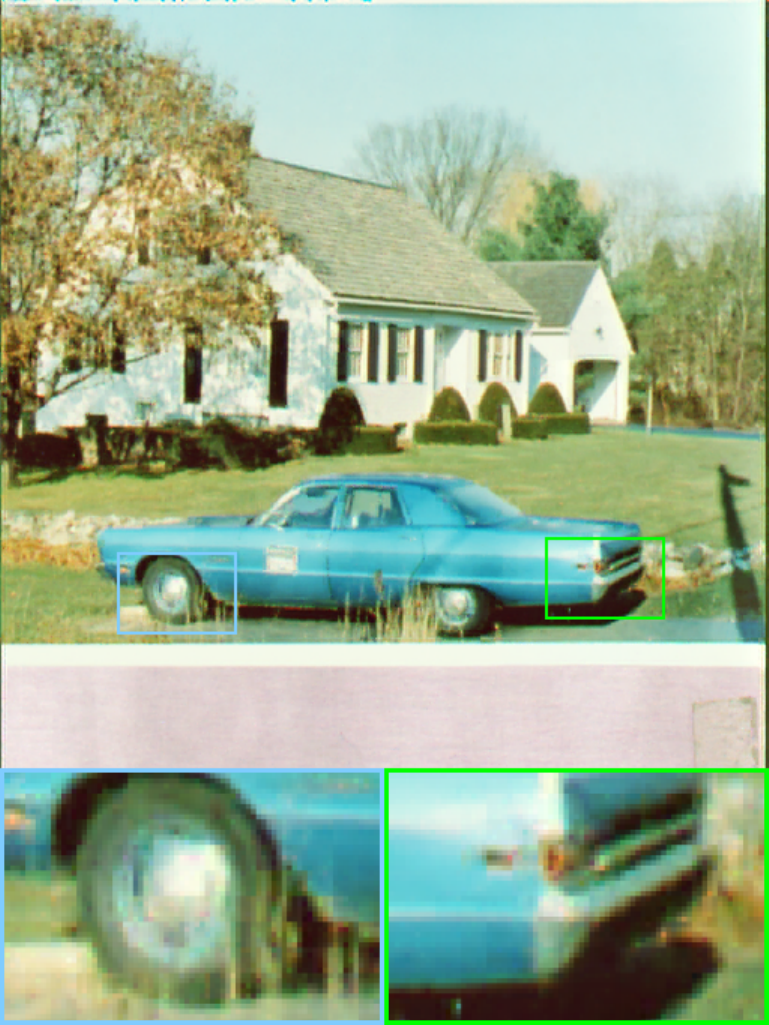}&
\includegraphics[width=0.106\textwidth]{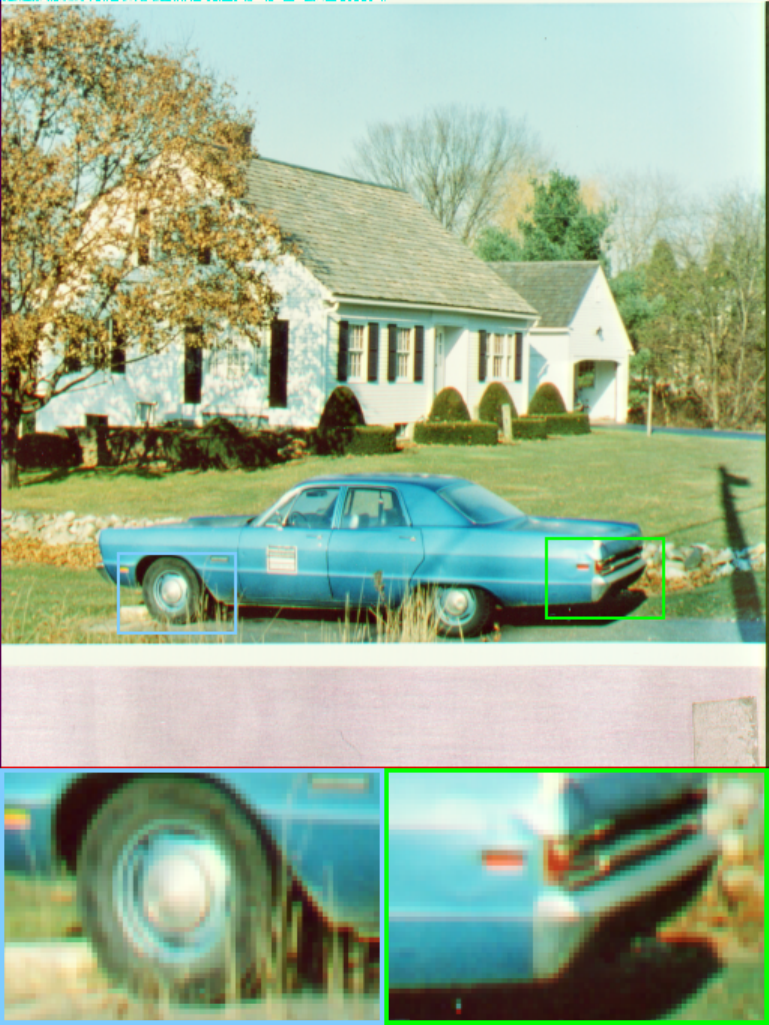}\\
PSNR 4.79 dB &
PSNR 22.29 dB &
PSNR 23.78 dB &
PSNR 24.63 dB &
PSNR 24.91 dB &
PSNR 23.85 dB &
PSNR 26.20 dB &
PSNR 26.54 dB &
PSNR Inf\\
			Observed &TRLRF \cite{TRLRF}&FTNN \cite{FTNN}&DIP\cite{DIP}&HLRTF \cite{HLRTF}&t-$\epsilon$-LogDet \cite{LogDet}&TCTV \cite{TCTV}& CRNL&Original\\
		\end{tabular}
	\end{center}
	\vspace{-0.3cm}
	\caption{From upper to lower: The results of image inpainting by different methods on color images {\it Peppers}, {\it Plane}, {\it Sailboat}, and {\it House} with sampling rate 0.2.\label{fig_inpainting_1}}
	\vspace{-0.5cm}
\end{figure*}
\begin{table*}[!h]
	\caption{The average quantitative results by different methods for image inpainting. The {\bf best} and \underline{second-best} values are highlighted. (PSNR $\uparrow$, SSIM $\uparrow$, and NRMSE $\downarrow$)\label{tab_inpainting}}\vspace{-0.4cm}
	\begin{center}
		\scriptsize
		\setlength{\tabcolsep}{2pt}
		\begin{spacing}{1.0}
			\begin{tabular}{clcccccccccccccccc}
				\toprule
				\multicolumn{2}{c}{Sampling rate}&\multicolumn{3}{c}{0.05}&\multicolumn{3}{c}{0.1}&\multicolumn{3}{c}{0.15}&\multicolumn{3}{c}{0.2}&\multicolumn{3}{c}{0.25}&\;\;\multirow{3}*{\tabincell{c}{\vspace{-0.05cm}Average \\ time (s)}}\\
				\cmidrule{1-17}
				Dataset&Method&PSNR &SSIM &NRMSE \;\; &PSNR &SSIM &NRMSE \;\; &PSNR &SSIM &NRMSE \;\; &PSNR &SSIM&NRMSE \;\;&PSNR &SSIM&NRMSE &~\\
				\midrule
				\multirow{8}*{\tabincell{c}{
						Color images\\ {\it Peppers}\\{\it Plane}\\{\it Sailboat}\\{\it House}\\{(512$\times$512$\times$3)}}}
&Observed&{4.78}&{0.031}&{0.975}\;\;&{5.01}&{0.043}&{0.949}\;\;&{5.26}&{0.054}&{0.922}\;\;&{5.52}&{0.065}&{0.895}\;\;&{5.81}&{0.075}&{0.866} &\--\--\\
&TRLRF&{13.88}&{0.190}&{0.352}\;\;&{16.94}&{0.345}&{0.251}\;\;&{20.30}&{0.556}&{0.171}\;\;&{22.89}&{0.701}&{0.127}\;\;&{24.70}&{0.786}&{0.104}&147\\
&FTNN&{13.66}&{0.309}&{0.356}\;\;&{19.57}&{0.586}&{0.186}\;\;&{21.96}&{0.697}&{0.143}\;\;&{23.62}&{0.766}&{0.119}\;\;&{25.03}&{0.814}&{0.102}&40 \\
&DIP&\underline{23.06}&\underline{0.776}&\underline{0.122}\;\;&\underline{24.95}&\underline{0.839}&\underline{0.098}\;\;&\underline{26.10}&\underline{0.869}&\underline{0.087}\;\;&{26.43}&{0.878}&{0.083}\;\;&{27.22}&{0.893}&{0.076}&44\\
&HLRTF&{18.93}&{0.538}&{0.199}\;\;&{22.47}&{0.719}&{0.133}\;\;&{24.23}&{0.779}&{0.109}\;\;&{25.66}&{0.829}&{0.095}\;\;&{26.28}&{0.846}&{0.088} &11\\
&t-$\epsilon$-LogDet&{17.98}&{0.433}&{0.221}\;\;&{20.92}&{0.605}&{0.157}\;\;&{22.86}&{0.709}&{0.126}\;\;&{24.25}&{0.774}&{0.108}\;\;&{25.36}&{0.819}&{0.095}& 7\\
&TCTV&{21.55}&{0.719}&{0.146}\;\;&{24.58}&{0.823}&{0.102}\;\;&{26.02}&{0.860}&{0.088}\;\;&\underline{27.05}&\underline{0.884}&\underline{0.079}\;\;&\underline{27.91}&\underline{0.902}&\underline{0.072} &107\\
&CRNL&\bf{23.39}&\bf{0.788}&\bf{0.116}\;\;&\bf{25.90}&\bf{0.871}&\bf{0.089}\;\;&\bf{27.18}&\bf{0.898}&\bf{0.077}\;\;&\bf{28.08}&\bf{0.915}&\bf{0.071}\;\;&\bf{28.68}&\bf{0.926}&\bf{0.067}&16 \\
	\midrule
\multirow{8}*{\tabincell{c}{MSIs\\ {\it Toys}\\{\it Cloth}\\{(256$\times$256$\times$31)}}}
&Observed&{16.12}&{0.221}&{0.975}\;\;&{16.36}&{0.253}&{0.949}\;\;&{16.60}&{0.285}&{0.922}\;\;&{16.86}&{0.316}&{0.895}\;\;&{17.15}&{0.346}&{0.866}&\--\--\\

&TRLRF&{28.71}&{0.684}&{0.220}\;\;&{33.05}&{0.841}&{0.148}\;\;&{35.24}&{0.902}&{0.114}\;\;&{36.10}&{0.916}&{0.104}\;\;&{36.56}&{0.923}&{0.099}&171\\

&FTNN&{32.81}&{0.876}&{0.149}\;\;&{37.18}&{0.945}&{0.094}\;\;&{40.21}&{0.968}&{0.070}\;\;&{42.68}&{0.979}&{0.055}\;\;&{44.77}&{0.986}&{0.045}& 233\\
&DIP&\underline{34.74}&\underline{0.919}&\underline{0.115}\;\;&\underline{39.73}&\underline{0.969}&\underline{0.066}\;\;&{40.08}&{0.972}&{0.061}\;\;&{42.61}&{0.985}&{0.046}\;\;&{43.94}&{0.986}&{0.040}&40\\
&HLRTF&{34.53}&{0.906}&{0.123}\;\;&{39.49}&{0.966}&{0.070}\;\;&\underline{42.86}&\underline{0.983}&\underline{0.049}\;\;&\underline{45.31}&\underline{0.990}&\underline{0.037}\;\;&\underline{47.22}&\underline{0.993}&\underline{0.031} &14\\

&t-$\epsilon$-LogDet&{31.11}&{0.800}&{0.181}\;\;&{35.80}&{0.917}&{0.109}\;\;&{38.81}&{0.954}&{0.080}\;\;&{40.98}&{0.970}&{0.064}\;\;&{42.78}&{0.979}&{0.053}&16\\

&TCTV&{34.73}&{0.917}&{0.119}\;\;&{38.68}&{0.960}&{0.079}\;\;&{41.50}&{0.977}&{0.058}\;\;&{43.76}&{0.985}&{0.046}\;\;&{45.69}&{0.989}&{0.038} &319\\

&CRNL&\bf{36.71}&\bf{0.945}&\bf{0.097}\;\;&\bf{41.12}&\bf{0.979}&\bf{0.058}\;\;&\bf{44.06}&\bf{0.988}&\bf{0.042}\;\;&\bf{46.40}&\bf{0.993}&\bf{0.032}\;\;&\bf{48.82}&\bf{0.995}&\bf{0.027}&74\\
				\midrule
				\multirow{8}*{\tabincell{c}{
						Videos\\ {\it Foreman}\\{\it Carphone}\\{(144$\times$176$\times$100)}}}
&Observed&{4.97}&{0.028}&{0.975}\;\;&{5.20}&{0.044}&{0.949}\;\;&{5.45}&{0.059}&{0.922}\;\;&{5.71}&{0.073}&{0.894}\;\;&{6.00}&{0.087}&{0.866} &\--\--\\

&TRLRF&{22.36}&{0.709}&{0.133}\;\;&{24.44}&{0.803}&{0.104}\;\;&{25.21}&{0.829}&{0.095}\;\;&{25.70}&{0.845}&{0.090}\;\;&{26.11}&{0.858}&{0.086}&217\\

&FTNN&{24.00}&{0.814}&{0.110}\;\;&{26.23}&{0.877}&{0.086}\;\;&{27.69}&{0.908}&{0.073}\;\;&{28.96}&{0.929}&{0.063}\;\;&{30.04}&{0.944}&\underline{0.056} &236\\
&DIP&{22.24}&{0.694}&{0.135}\;\;&{24.31}&{0.787}&{0.107}\;\;&{26.13}&{0.844}&{0.087}\;\;&{27.01}&{0.869}&{0.079}\;\;&{27.64}&{0.884}&{0.073}&14\\
&HLRTF&{22.53}&{0.690}&{0.131}\;\;&{24.64}&{0.769}&{0.103}\;\;&{26.26}&{0.831}&{0.086}\;\;&{28.07}&{0.883}&{0.070}\;\;&{29.39}&{0.910}&{0.060}&13 \\

&t-$\epsilon$-LogDet&{17.29}&{0.434}&{0.242}\;\;&{24.90}&{0.784}&{0.100}\;\;&{26.51}&{0.840}&{0.083}\;\;&{27.83}&{0.876}&{0.072}\;\;&{28.98}&{0.903}&{0.063}&16 \\

&TCTV&\underline{26.50}&\underline{0.883}&\underline{0.084}\;\;&\underline{28.32}&\underline{0.917}&\underline{0.068}\;\;&\underline{29.61}&\underline{0.935}&\underline{0.059}\;\;&\underline{30.71}&\underline{0.948}&\underline{0.052}\;\;&\bf{31.65}&\underline{0.957}&\bf{0.047}& 264\\

&CRNL&\bf{26.61}&\bf{0.893}&\bf{0.082}\;\;&\bf{28.80}&\bf{0.928}&\bf{0.063}\;\;&\bf{29.99}&\bf{0.943}&\bf{0.056}\;\;&\bf{30.96}&\bf{0.956}&\bf{0.050}\;\;&\underline{31.58}&\bf{0.961}&\bf{0.047} &40\\
				\bottomrule
			\end{tabular}
		\end{spacing}
	\end{center}
	\vspace{-0.9cm}
\end{table*}    
\begin{figure*}[t]
	\scriptsize
	\setlength{\tabcolsep}{0.9pt}
	\begin{center}
		\begin{tabular}{ccccccccc}
\includegraphics[width=0.106\textwidth]{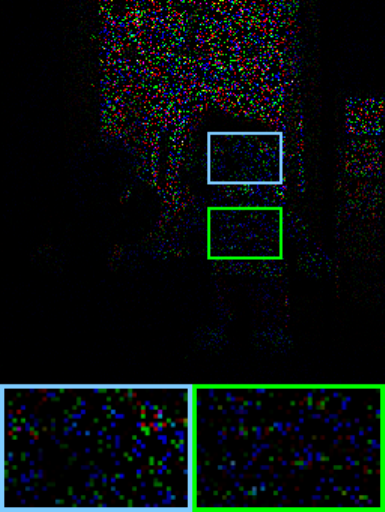}&
\includegraphics[width=0.106\textwidth]{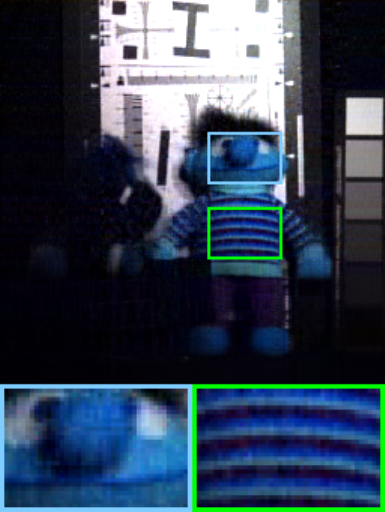}&
\includegraphics[width=0.106\textwidth]{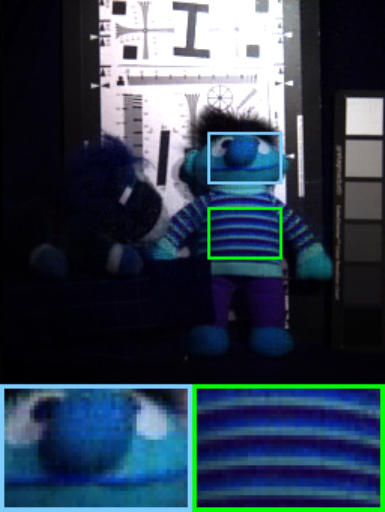}&
\includegraphics[width=0.106\textwidth]{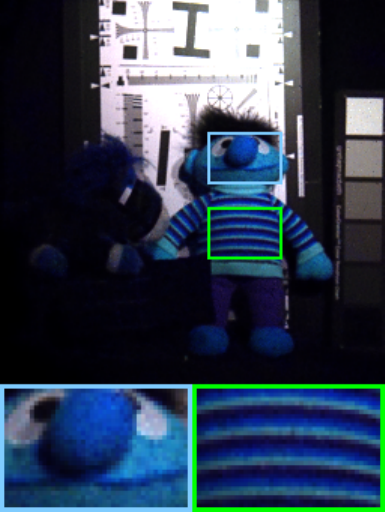}&
\includegraphics[width=0.106\textwidth]{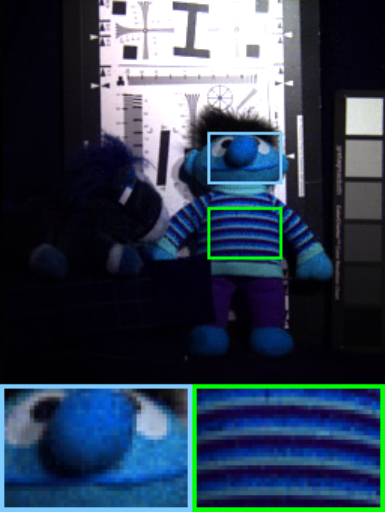}&
\includegraphics[width=0.106\textwidth]{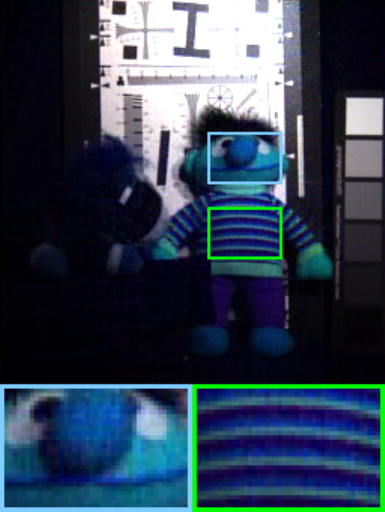}&
\includegraphics[width=0.106\textwidth]{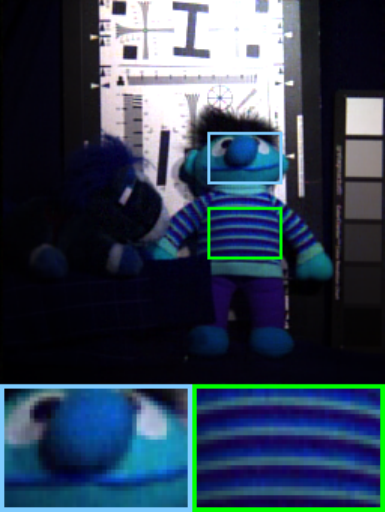}&
\includegraphics[width=0.106\textwidth]{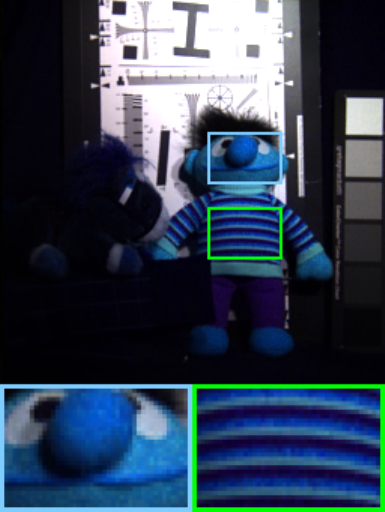}&
\includegraphics[width=0.106\textwidth]{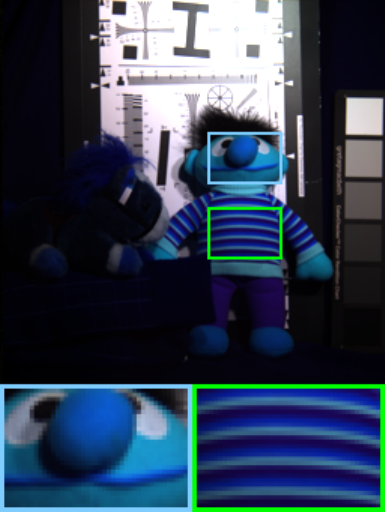}\\
PSNR 16.69 dB &
PSNR 38.38 dB &
PSNR 41.82 dB &
PSNR 43.39 dB &
PSNR 43.67 dB &
PSNR 40.54 dB &
PSNR 43.59 dB &
PSNR 45.20 dB &
PSNR Inf\\
\includegraphics[width=0.106\textwidth]{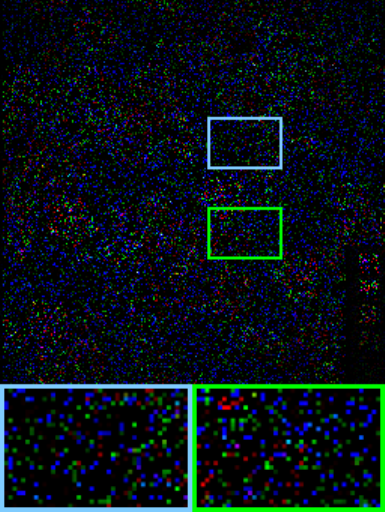}&
\includegraphics[width=0.106\textwidth]{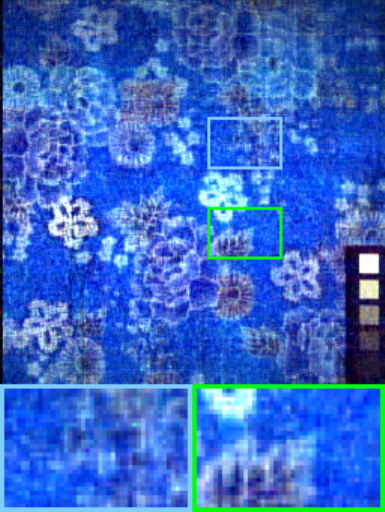}&
\includegraphics[width=0.106\textwidth]{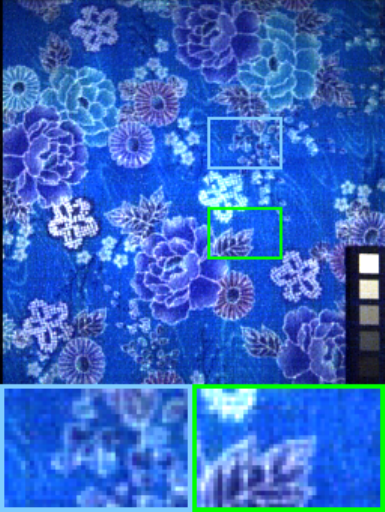}&
\includegraphics[width=0.106\textwidth]{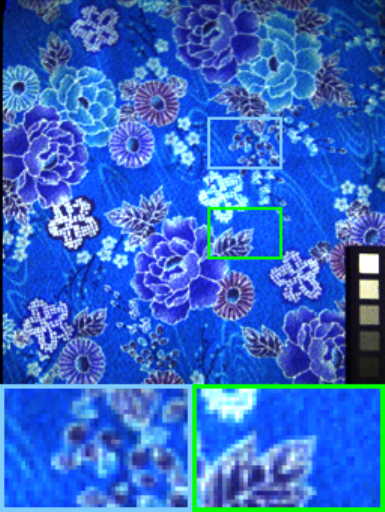}&
\includegraphics[width=0.106\textwidth]{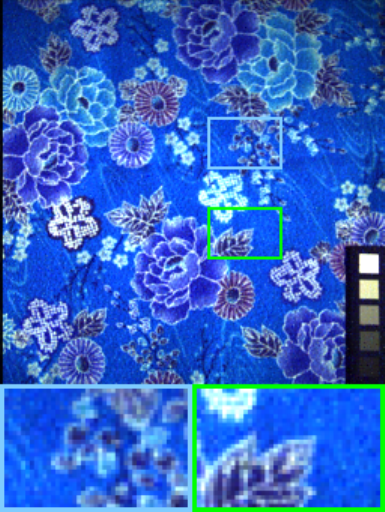}&
\includegraphics[width=0.106\textwidth]{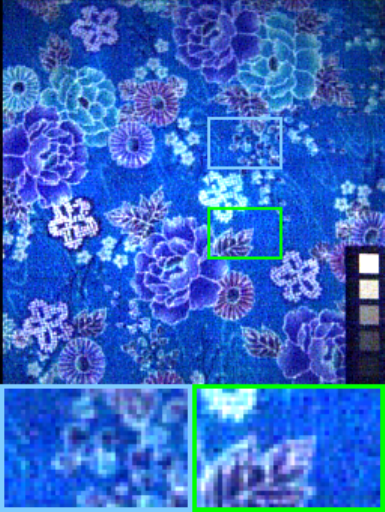}&
\includegraphics[width=0.106\textwidth]{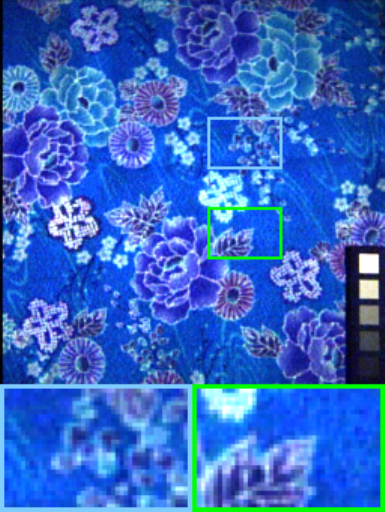}&
\includegraphics[width=0.106\textwidth]{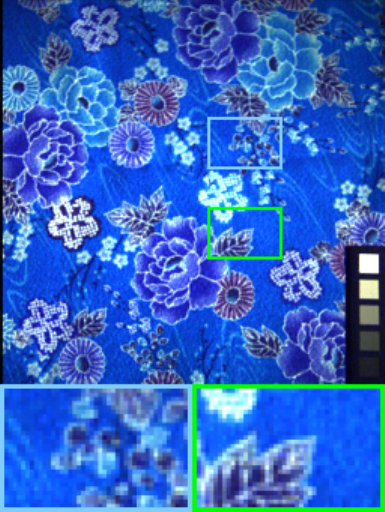}&
\includegraphics[width=0.106\textwidth]{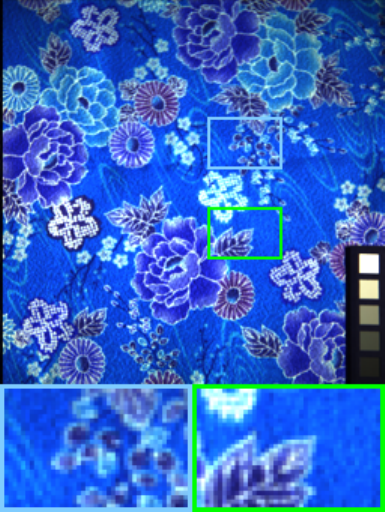}\\
PSNR 16.02 dB &
PSNR 27.73 dB &
PSNR 32.55 dB &
PSNR 36.08 dB &
PSNR 35.31 dB &
PSNR 31.05 dB &
PSNR 33.77 dB &
PSNR 37.04 dB &
PSNR Inf\\
\includegraphics[width=0.106\textwidth]{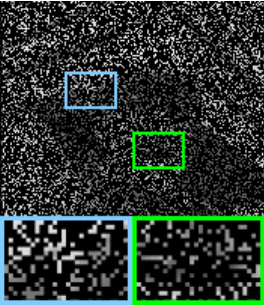}&
\includegraphics[width=0.106\textwidth]{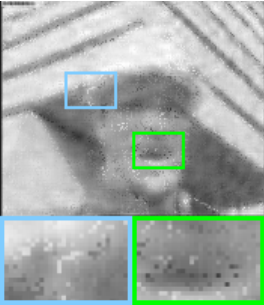}&
\includegraphics[width=0.106\textwidth]{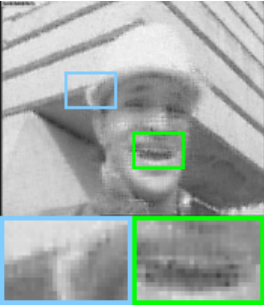}&
\includegraphics[width=0.106\textwidth]{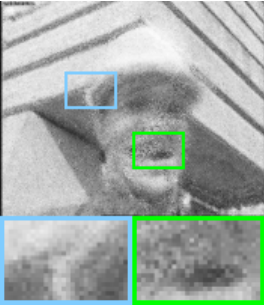}&
\includegraphics[width=0.106\textwidth]{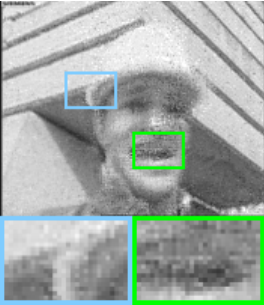}&
\includegraphics[width=0.106\textwidth]{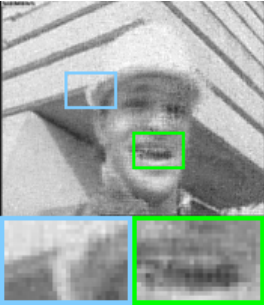}&
\includegraphics[width=0.106\textwidth]{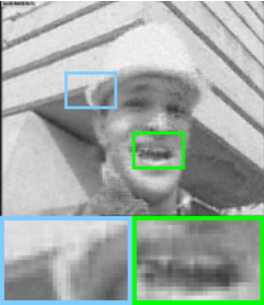}&
\includegraphics[width=0.106\textwidth]{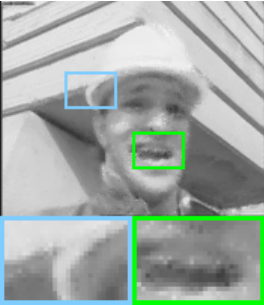}&
\includegraphics[width=0.106\textwidth]{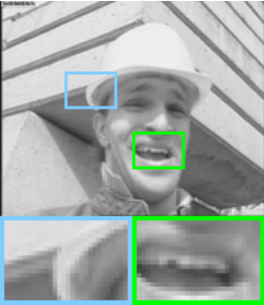}\\
PSNR 4.40 dB &
PSNR 24.89 dB &
PSNR 29.12 dB &
PSNR 26.68 dB &
PSNR 27.96 dB &
PSNR 28.21 dB &
PSNR 31.34 dB &
PSNR 31.37 dB &
PSNR Inf\\
\includegraphics[width=0.106\textwidth]{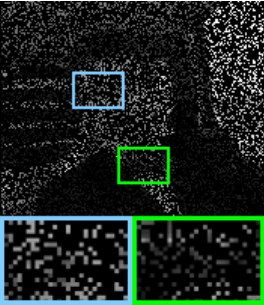}&
\includegraphics[width=0.106\textwidth]{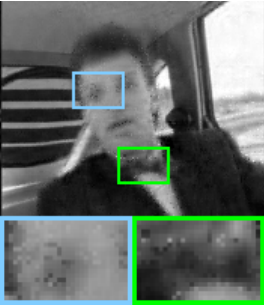}&
\includegraphics[width=0.106\textwidth]{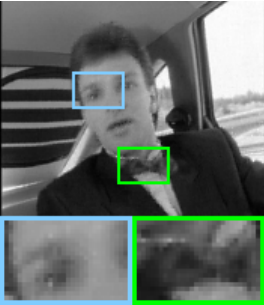}&
\includegraphics[width=0.106\textwidth]{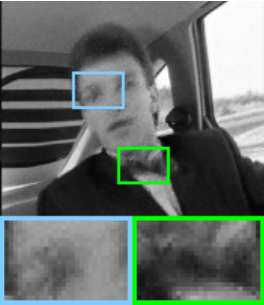}&
\includegraphics[width=0.106\textwidth]{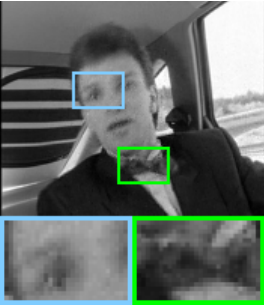}&
\includegraphics[width=0.106\textwidth]{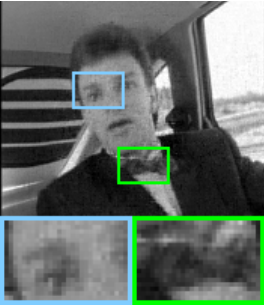}&
\includegraphics[width=0.106\textwidth]{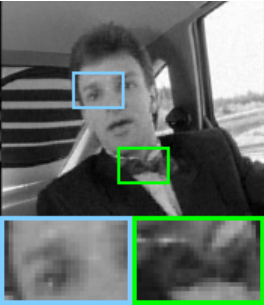}&
\includegraphics[width=0.106\textwidth]{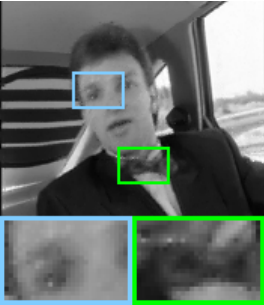}&
\includegraphics[width=0.106\textwidth]{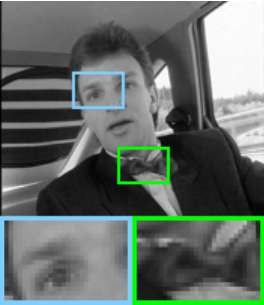}\\
PSNR 7.59 dB &
PSNR 27.34 dB &
PSNR 30.95 dB &
PSNR 28.60 dB &
PSNR 30.82 dB &
PSNR 29.76 dB &
PSNR 31.96 dB &
PSNR 31.80 dB &
PSNR Inf\\
			Observed &TRLRF \cite{TRLRF}&FTNN \cite{FTNN}&DIP\cite{DIP}&HLRTF \cite{HLRTF}&t-$\epsilon$-LogDet \cite{LogDet}&TCTV \cite{TCTV}& CRNL&Original\\
		\end{tabular}
	\end{center}
	\vspace{-0.3cm}
	\caption{From upper to lower: The results of image inpainting by different methods on MSIs {\it Toys} and {\it Cloth} with sampling rate 0.1, and videos {\it Foreman} and {\it Carphone} with sampling rate 0.25.\label{fig_inpainting_2}}
	\vspace{-0.5cm}
\end{figure*}
\section{Experiments}\label{sec_exp}
The proposed CRNL is versatile for data processing on and off-meshgrid. To validate such superiority, we conduct experiments with data processing tasks on-meshgrid (image inpainting and denoising) and off-meshgrid (multivariate regression problems, e.g., climate data prediction and point cloud recovery). We first introduce experimental settings of different tasks, and then present the experimental results. Our experiments are conducted on a computer with an i5-10400 CPU and an RTX 2080ti GPU.
\subsection{On Meshgrid Data}
\subsubsection{Image Inpainting} We first apply our method to the image inpainting task, which is a typical data recovery task on-meshgrid.
We collect three types of data, i.e., color images ({\it Peppers}, {\it Plane}, {\it Sailboat}, and {\it House}\footnote{\url{http://sipi.usc.edu/database/database.php}}), multispectral images (MSIs) ({\it Toys} and {\it Cloth} in the CAVE dataset\footnote{\url{https://www.cs.columbia.edu/CAVE/databases/multispectral/}} \cite{CAVE}), and videos ({\it Foreman} and {\it Carphone}\footnote{\url{http://trace.eas.asu.edu/yuv/}}) as testing data. We consider sampling rates (SRs) 0.05, 0.1, 0.15, 0.2, and 0.25 to generate incompleted images for testing. We compare our method with state-of-the-art unsupervised methods against this task, i.e., TRLRF\cite{TRLRF}, FTNN \cite{FTNN}, DIP \cite{DIP}, HLRTF\cite{HLRTF}, t-$\epsilon$-LogDet\cite{LogDet}, and TCTV\cite{TCTV}. We use peak-signal-to-noisy ratio (PSNR), structural similarity (SSIM), and normalized root mean square error (NRMSE) to evaluate the quality of recovered results.
\subsubsection{Image Denoising} Then we consider the image denoising, which is also a typical data recovery task on-meshgrid. We adopt multispectral images and hyperspectral images (HSIs) as testing data. Specifically, we consider four sub-images of the {\it WDC} and {\it Pavia U} HSI datasets\footnote{\url{http://sipi.usc.edu/database/database.php}}, denoted by {\it WDC-1}, {\it WDC-2}, {\it Pavia U-1}, and {\it Pavia U-2}. Meanwhile, we adopt two MSIs {\it Cups} and {\it Fruits} from the CAVE dataset \cite{CAVE} as testing data. We consider Gaussian noise with standard deviations 0.1, 0.15, 0.2, 0.25, and 0.3 to generate noisy images for testing. We compare our method with state-of-the-art model-based methods (LRTDTV \cite{LRTDTV}, LTDL\cite{LTDL}, RCTV\cite{RCTV}, and WNLRATV\cite{WNLRATV}) and deep learning-based methods (HSID-CNN\cite{HSIDCNN} and SDeCNN\cite{SDeCNN}). Here, LTDL and WNLRATV are nonlocal low-rank-based denoising methods. We use PSNR, SSIM, and NRMSE to evaluate the quality of denoising results.
\subsection{Beyond Meshgrid Data}
Since our CRNL learns a continuous representation, it is versatile for both on-meshgrid and off-meshgrid data processing, while classical NSS-based methods are discrete representations that are solely suitable for meshgrid data. To validate such superiority, we apply our method to multivariate regression problems beyond meshgrid. For these multivariate regression tasks, we report the NRMSE and R-Square results. 
\subsubsection{Synthetic Data} First, we employ our CRNL to two-dimensional regression problems by using the following functions (referred to \cite{Lee_1997}) to construct synthetic data:
\begin{equation}
	\begin{split}
		f_1(x,y) = &\frac{1}{3}\exp\big{(}-\frac{81}{4}((x-1/2)^2+(y-1/2)^2)\big{)},\\
		f_2(x,y) = &\frac{1.25+\cos(5.4y)}{6+6(3x-1)^2},\\
		f_3(x,y) = &\frac{1}{9}(\tanh(9-9x-9y)+1),\\
		f_4(x,y) = &2\exp\big{(}-30((x-\frac{1}{3})^2+(y-\frac{1}{3})^2)\big{)}\\
		&\;\;\;\;\;\quad-\exp\big{(}-20((x-\frac{2}{3})^2+(y-\frac{2}{3})^2)\big{)}.
	\end{split}
\end{equation}
Specifically, we randomly sample 3000 spatial points in the space and calculate the corresponding function values $f_1(x,y),\cdots,f_4(x,y)$. We split these points into training/testing datasets with split ratios 2/8, 1.5/8.5, and 1/9 respectively. Then we train our model with the training dataset and test our method on the testing dataset. We report the average regression results of the three split ratios. We compare our method with standard regression methods including support vector regressor (SVR), K-neighbors regressor (KNR), decision tree (DT), and random forest (RF). Meanwhile, we include FSA-HTF \cite{TSP_CP}, a tensor decomposition-based method, into the comparison of multivariate regression.
\subsubsection{Weather Data Prediction} To further test the effectiveness of our method beyond meshgrid, we apply it to more complicated real-world unordered datasets. Specifically, we consider the precipitation climate data\footnote{Provided in \url{https://daac.ornl.gov/cgi-bin/dsviewer.pl?ds_id=2130}} to test our algorithm. We collect four precipitation datasets from the North America located at (55$^\circ$N, 117$^\circ$W), (53$^\circ$N, 111$^\circ$W), (50$^\circ$N, 106$^\circ$W), and (47$^\circ$N, 100$^\circ$W), respectively. These datasets contain precipitation values at some spatial points around the selected areas. The climate data prediction problem refers to inferring the precipitation value of the given spatial point, provided that some training datasets are given. Hence, it is also a multivariate regression problem. The training/testing data split ratio is set to 2/8, 1.5/8.5, and 1/9 respectively and we report the average predication results of the three cases.
\subsubsection{Point Cloud Recovery} We further evaluate our method on higher-order point cloud data processing. Specifically, we consider the point cloud recovery task, which aims to estimate the color information of the given point cloud. The original point cloud with $n$ points is represented by an $n$-by-6 matrix ${\bf P}\in{\mathbb R}^{n\times 6}$, where each point ${\bf P}_{(d,:)}$ ($d=1,2,\cdots,n$) is an ($x,y,z$)-(R,G,B) formed six-dimensional vector, which contains both coordinate and color information. We split the point cloud into training and testing datasets. The training dataset contains $n'$ pairs of ($x,y,z$) and (R,G,B) information, where the model takes ($x,y,z$) as input and is expected to output the color information (R,G,B). The testing dataset contains $n-n'$ pairs of ($x,y,z$) and (R,G,B) data to test the trained model. We use four color point cloud datasets in the SHOT website\footnote{\url{http://www.vision.deis.unibo.it/research/80-shot}}, named {\it Mario}, {\it Rabbit}, {\it Frog}, and {\it Duck}. We consider the training/testing data split ratio as 2/8, 1.5/8.5, and 1/9 respectively and report the average prediction results.
\begin{figure*}[t]
	\scriptsize
	\setlength{\tabcolsep}{0.9pt}
	\begin{center}
		\begin{tabular}{ccccccccc}
			\includegraphics[width=0.106\textwidth]{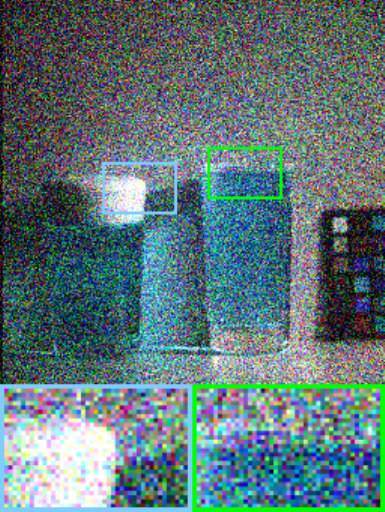}&
			\includegraphics[width=0.106\textwidth]{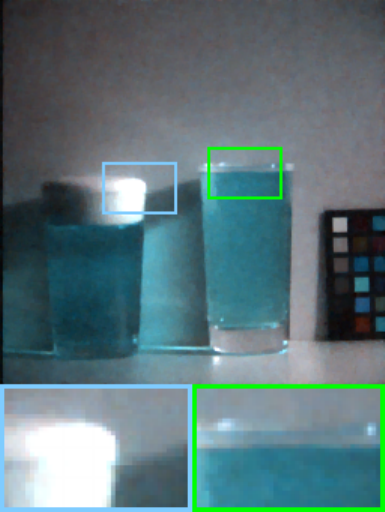}&
						\includegraphics[width=0.106\textwidth]{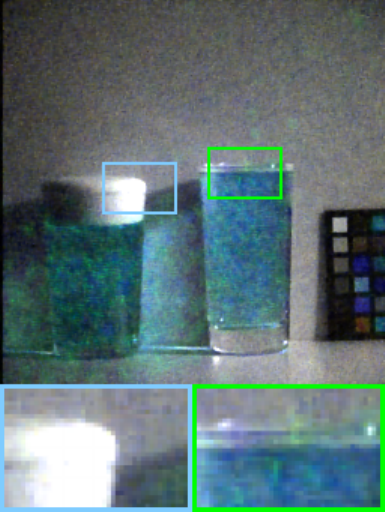}&
						\includegraphics[width=0.106\textwidth]{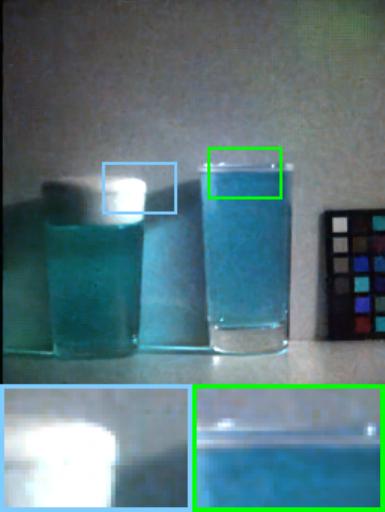}&
			\includegraphics[width=0.106\textwidth]{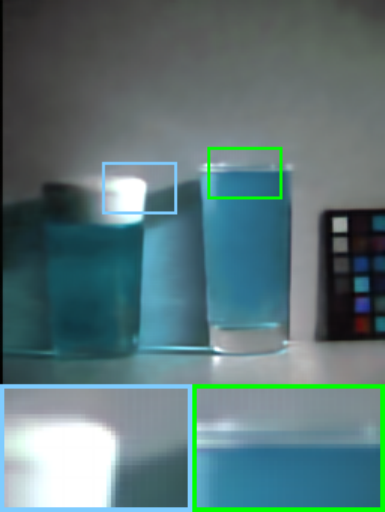}&
			\includegraphics[width=0.106\textwidth]{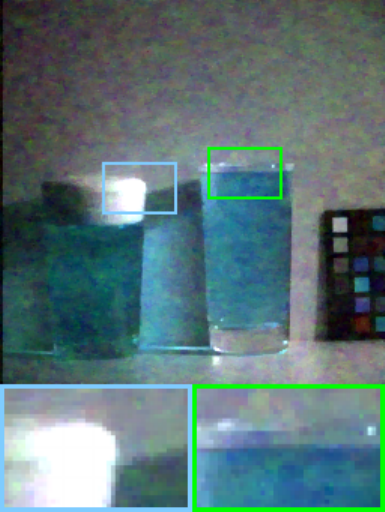}&
			\includegraphics[width=0.106\textwidth]{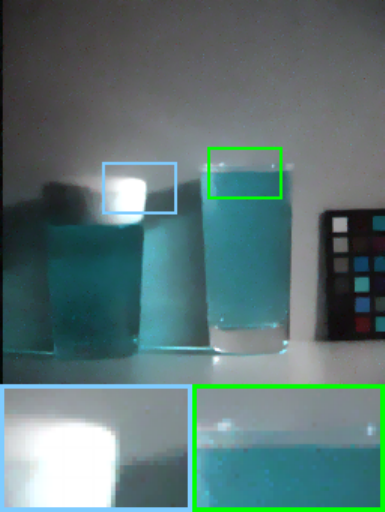}&
			\includegraphics[width=0.106\textwidth]{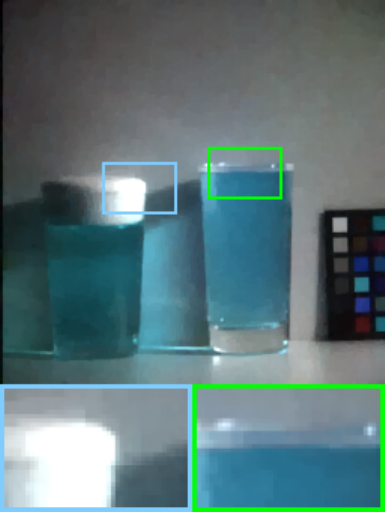}&
			\includegraphics[width=0.106\textwidth]{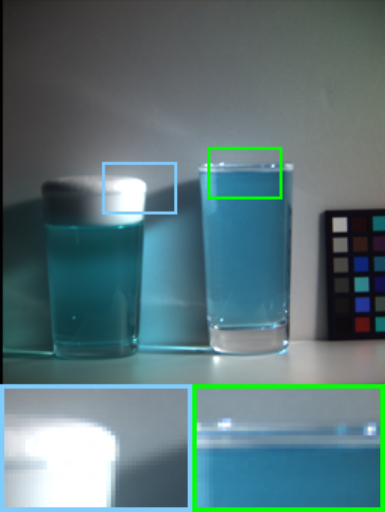}\\
PSNR 13.98 dB &
PSNR 33.93 dB &
PSNR 29.29 dB &
PSNR 34.07 dB &
PSNR 36.43 dB &
PSNR 30.42 dB &
PSNR 32.78 dB &
PSNR 37.24 dB &
PSNR Inf\\
			\includegraphics[width=0.106\textwidth]{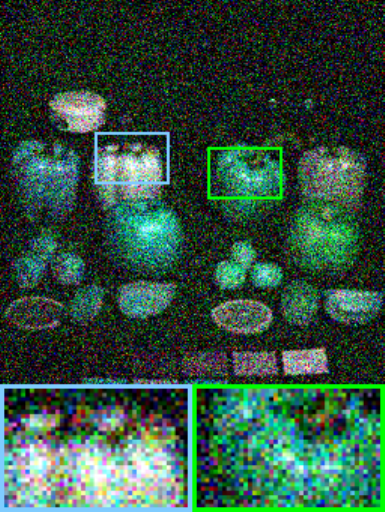}&
			\includegraphics[width=0.106\textwidth]{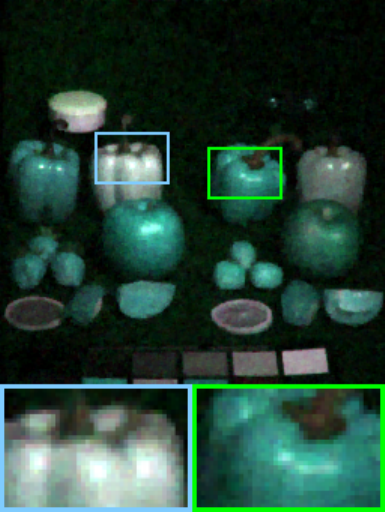}&
						\includegraphics[width=0.106\textwidth]{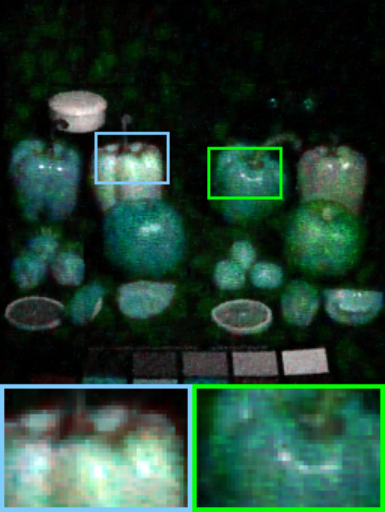}&
						\includegraphics[width=0.106\textwidth]{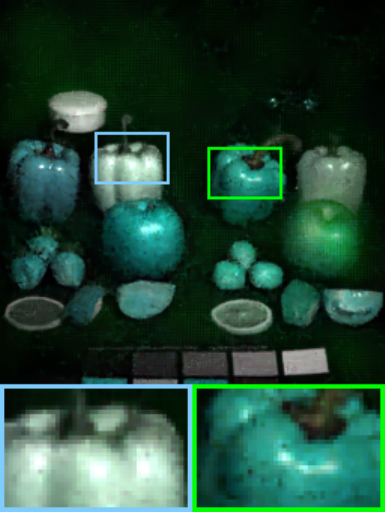}&
			\includegraphics[width=0.106\textwidth]{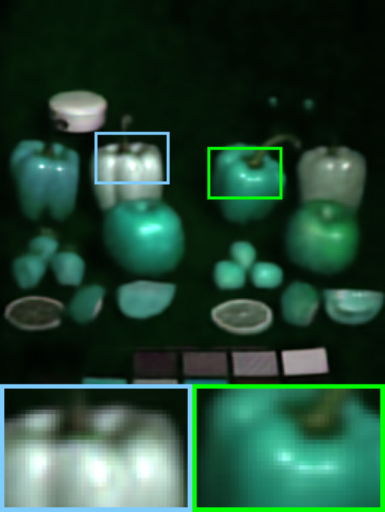}&
			\includegraphics[width=0.106\textwidth]{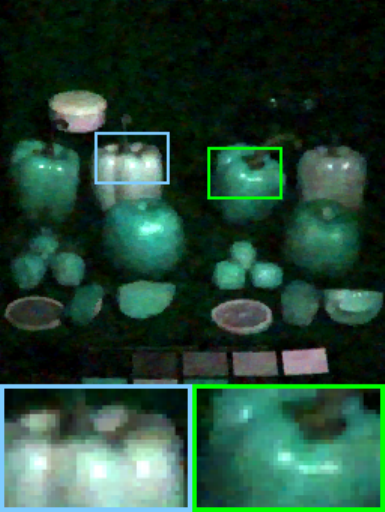}&
			\includegraphics[width=0.106\textwidth]{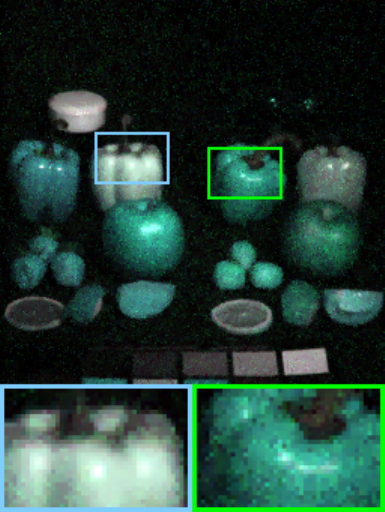}&
			\includegraphics[width=0.106\textwidth]{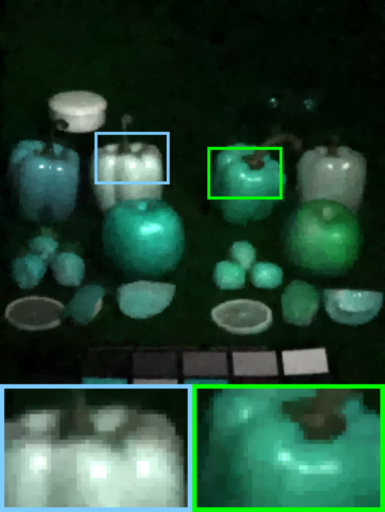}&
			\includegraphics[width=0.106\textwidth]{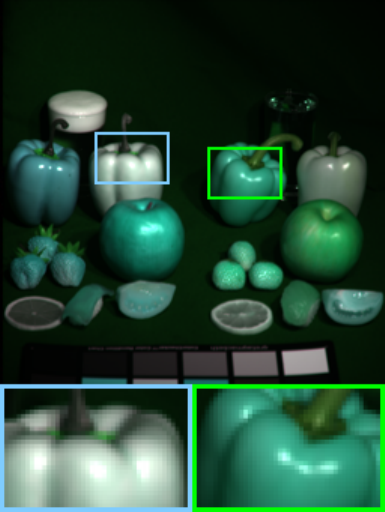}\\
PSNR 19.42 dB &
PSNR 38.41 dB &
PSNR 36.90 dB &
PSNR 40.09 dB &
PSNR 40.90 dB &
PSNR 35.56 dB &
PSNR 36.80 dB &
PSNR 41.22 dB &
PSNR Inf\\
			Observed &LRTDTV \cite{LRTDTV}&HSID-CNN\cite{HSIDCNN}&SDeCNN\cite{SDeCNN}&LTDL \cite{LTDL}&RCTV \cite{RCTV}&WNLRATV \cite{WNLRATV}& CRNL&Original\\
		\end{tabular}
	\end{center}
	\vspace{-0.3cm}
	\caption{From upper to lower: The results of multi-dimensional image denoising by different methods on MSIs {\it Cups} and {\it Fruits} with noisy level $\sigma =0.2$.\label{fig_denoising_2}}
	\vspace{-0.2cm}
\end{figure*}
\begin{table*}[t]
	\caption{The average quantitative results by different methods for multi-dimensional image denoising. The {\bf best} and \underline{second-best} values are highlighted. (PSNR $\uparrow$, SSIM $\uparrow$, and NRMSE $\downarrow$)\label{tab_denoising}}\vspace{-0.4cm}
	\begin{center}
		\scriptsize
		\setlength{\tabcolsep}{2pt}
		\begin{spacing}{1.0}
			\begin{tabular}{clcccccccccccccccc}
				\toprule
				\multicolumn{2}{c}{Noisy level}&\multicolumn{3}{c}{$\sigma=0.1$}&\multicolumn{3}{c}{$\sigma=0.15$}&\multicolumn{3}{c}{$\sigma=0.2$}&\multicolumn{3}{c}{$\sigma=0.25$}&\multicolumn{3}{c}{$\sigma=0.3$}&\;\;\multirow{3}*{\tabincell{c}{\vspace{-0.05cm}Average \\ time (s)}}\\
				\cmidrule{1-17}
				Dataset&Method&PSNR &SSIM &NRMSE \;\; &PSNR &SSIM &NRMSE \;\; &PSNR &SSIM &NRMSE \;\; &PSNR &SSIM&NRMSE \;\;&PSNR &SSIM&NRMSE&~\\
				\midrule
				\multirow{8}*{\tabincell{c}{MSIs\\ {\it Cups}\\{\it Fruits}\\{(256$\times$256$\times$31)}}}
				&Observed&{22.72}&{0.326}&{0.495}\;\;&{19.20}&{0.186}&{0.742}\;\;&{16.70}&{0.118}&{0.990}\;\;&{14.76}&{0.081}&{1.237}\;\;&{13.18}&{0.058}&{1.485}&\--\-- \\
				
				&LRTDTV&{38.97}&\underline{0.974}&{0.091}\;\;&{37.80}&{0.957}&{0.104}\;\;&{36.17}&{0.927}&{0.123}\;\;&{34.37}&{0.880}&{0.149}\;\;&{32.52}&\underline{0.816}&{0.182}& 49\\
				
				&HSID-CNN&{34.83}&{0.941}&{0.131}\;\;&{34.25}&{0.918}&{0.137}\;\;&{33.09}&{0.874}&{0.153}\;\;&{31.17}&{0.797}&{0.187}\;\;&{28.86}&{0.681}&{0.240}&199\\
				
				&SDeCNN&{39.26}&{0.969}&{0.085}\;\;&{38.89}&{0.967}&{0.088}\;\;&{37.08}&{0.940}&{0.106}\;\;&{33.32}&{0.843}&{0.156}\;\;&{29.90}&{0.708}&{0.218}&14\\
				
				&LTDL&\underline{42.34}&\bf{0.986}&\underline{0.055}\;\;&\underline{40.21}&\bf{0.979}&\underline{0.069}\;\;&\underline{38.66}&\bf{0.971}&\underline{0.083}\;\;&\underline{37.25}&\underline{0.962}&\underline{0.097}\;\;&\underline{36.28}&\bf{0.955}&\underline{0.108} &632\\
				
				&RCTV&{38.40}&{0.947}&{0.084}\;\;&{35.41}&{0.895}&{0.119}\;\;&{32.99}&{0.834}&{0.156}\;\;&{31.52}&{0.774}&{0.185}\;\;&{30.17}&{0.713}&{0.217}&5\\
				
				&WNLRATV&{37.29}&{0.938}&{0.106}\;\;&{36.21}&{0.925}&{0.125}\;\;&{34.79}&{0.886}&{0.157}\;\;&{32.59}&{0.819}&{0.197}\;\;&{30.27}&{0.728}&{0.257}& 234\\
				
				&CRNL&\bf{42.57}&\bf{0.986}&\bf{0.054}\;\;&\bf{40.66}&\underline{0.977}&\bf{0.068}\;\;&\bf{39.23}&\underline{0.968}&\bf{0.079}\;\;&\bf{38.22}&\bf{0.965}&\bf{0.089}\;\;&\bf{37.38}&\bf{0.955}&\bf{0.098}&33 \\
				\midrule
				\multirow{8}*{\tabincell{c}{
						HSIs\\ {\it WDC-1}\\{\it WDC-2}\\{(256$\times$256$\times$32)}}}
				&Observed&{21.22}&{0.468}&{0.529}\;\;&{17.70}&{0.298}&{0.793}\;\;&{15.20}&{0.202}&{1.057}\;\;&{13.26}&{0.143}&{1.321}\;\;&{11.68}&{0.105}&{1.585}&\--\-- \\
				
				&LRTDTV&{32.48}&{0.920}&{0.151}\;\;&{32.06}&{0.913}&{0.158}\;\;&{31.47}&{0.900}&{0.169}\;\;&{30.63}&{0.878}&{0.187}\;\;&{29.57}&{0.845}&{0.212}& 44\\
				
				&HSID-CNN&{32.63}&{0.927}&{0.175}\;\;&{31.80}&{0.913}&{0.185}\;\;&{30.55}&{0.884}&{0.205}\;\;&{28.78}&{0.834}&{0.240}\;\;&{26.66}&{0.760}&{0.296}&202\\
				
				&SDeCNN&{34.66}&{0.953}&\underline{0.119}\;\;&{34.14}&\bf{0.950}&{0.127}\;\;&{32.58}&\underline{0.927}&{0.153}\;\;&{29.87}&{0.860}&{0.206}\;\;&{27.32}&{0.775}&{0.270}&16\\
				
				&LTDL&\underline{36.63}&\bf{0.970}&\bf{0.093}\;\;&\underline{34.15}&\underline{0.947}&\underline{0.124}\;\;&\underline{32.52}&{0.923}&\underline{0.150}\;\;&\underline{31.24}&\underline{0.896}&\underline{0.175}\;\;&\underline{30.23}&\underline{0.868}&\underline{0.197}& 688\\
				
				&RCTV&{33.88}&{0.942}&{0.127}\;\;&{31.44}&{0.900}&{0.166}\;\;&{29.51}&{0.851}&{0.206}\;\;&{28.21}&{0.801}&{0.240}\;\;&{26.99}&{0.749}&{0.277}&5 \\
				
				&WNLRATV&{34.27}&{0.938}&{0.128}\;\;&{32.13}&{0.908}&{0.166}\;\;&{30.60}&{0.877}&{0.195}\;\;&{29.34}&{0.838}&{0.235}\;\;&{28.04}&{0.787}&{0.279}& 269\\
				
				&CRNL&\bf{36.72}&\underline{0.969}&\bf{0.093}\;\;&\bf{34.32}&\underline{0.947}&\bf{0.121}\;\;&\bf{32.91}&\bf{0.930}&\bf{0.144}\;\;&\bf{31.87}&\bf{0.913}&\bf{0.162}\;\;&\bf{31.03}&\bf{0.895}&\bf{0.178}&29 \\
				\midrule
				\multirow{8}*{\tabincell{c}{
						HSIs\\ {\it Pavia U-1}\\{\it Pavia U-2}\\{(256$\times$256$\times$32)}}}
				&Observed&{20.00}&{0.392}&{0.683}\;\;&{16.48}&{0.235}&{1.023}\;\;&{13.98}&{0.153}&{1.365}\;\;&{12.04}&{0.106}&{1.706}\;\;&{10.46}&{0.077}&{2.048}&\--\--\\
				&LRTDTV&{33.28}&{0.939}&{0.148}\;\;&{32.72}&{0.931}&{0.158}\;\;&{31.79}&{0.914}&{0.176}\;\;&{30.74}&{0.888}&{0.199}\;\;&{29.38}&{0.846}&{0.233}&44\\
				&HSID-CNN&{34.37}&{0.953}&{0.136}\;\;&{33.12}&{0.936}&{0.155}\;\;&{31.34}&{0.901}&{0.188}\;\;&{29.04}&{0.839}&{0.244}\;\;&{26.53}&{0.750}&{0.324}&201\\
				&SDeCNN&{35.46}&{0.957}&{0.118}\;\;&{34.09}&{0.946}&\underline{0.133}\;\;&{32.95}&\underline{0.934}&\underline{0.153}\;\;&{29.91}&{0.856}&{0.230}\;\;&{26.80}&{0.751}&{0.321}&15\\
				&LTDL&\underline{36.39}&\underline{0.965}&\underline{0.107}\;\;&\underline{34.35}&\underline{0.949}&\underline{0.133}\;\;&\underline{32.98}&\underline{0.934}&{0.154}\;\;&\underline{31.62}&\underline{0.914}&\underline{0.179}\;\;&\underline{30.88}&\underline{0.902}&\underline{0.195}&753\\
				&RCTV&{33.91}&{0.941}&{0.140}\;\;&{31.56}&{0.903}&{0.182}\;\;&{29.59}&{0.856}&{0.228}\;\;&{27.98}&{0.803}&{0.275}\;\;&{26.81}&{0.750}&{0.317}&5\\
				&WNLRATV&{34.23}&{0.938}&{0.141}\;\;&{32.45}&{0.913}&{0.177}\;\;&{31.11}&{0.886}&{0.214}\;\;&{29.38}&{0.837}&{0.263}\;\;&{28.52}&{0.802}&{0.317}&252\\
				&CRNL&\bf{36.68}&\bf{0.967}&\bf{0.104}\;\;&\bf{34.70}&\bf{0.952}&\bf{0.128}\;\;&\bf{33.31}&\bf{0.938}&\bf{0.149}\;\;&\bf{32.30}&\bf{0.927}&\bf{0.167}\;\;&\bf{31.58}&\bf{0.917}&\bf{0.181}&23\\
				\bottomrule
			\end{tabular}
		\end{spacing}
	\end{center}
	\vspace{-0.7cm}
\end{table*}    
\begin{figure*}[t]
	\scriptsize
	\setlength{\tabcolsep}{0.9pt}
	\begin{center}
			\begin{tabular}{ccccccccc}
			\includegraphics[width=0.106\textwidth]{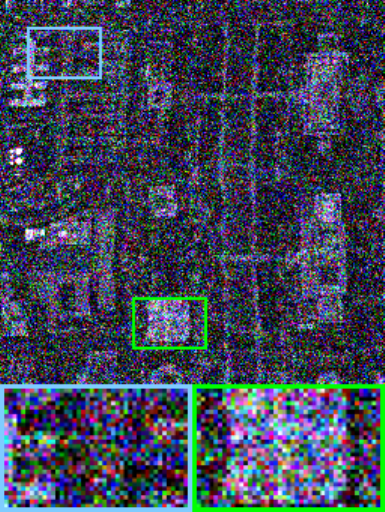}&
			\includegraphics[width=0.106\textwidth]{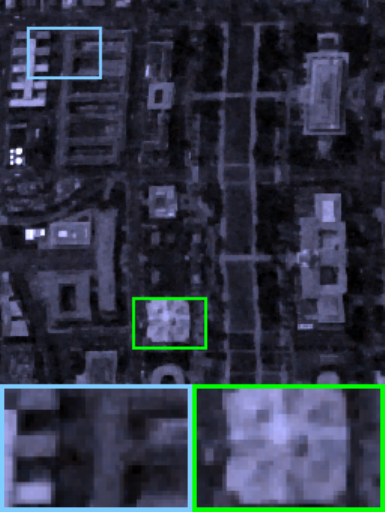}&
						\includegraphics[width=0.106\textwidth]{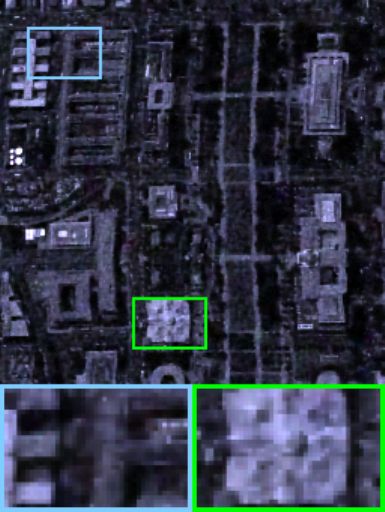}&
						\includegraphics[width=0.106\textwidth]{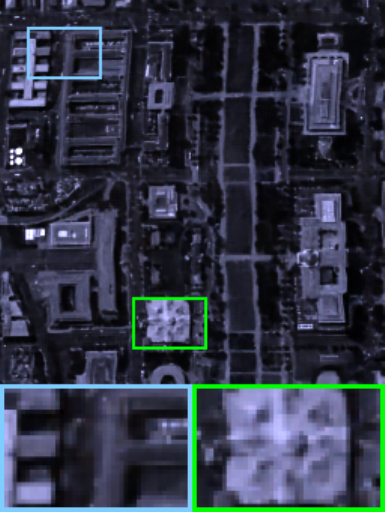}&
			\includegraphics[width=0.106\textwidth]{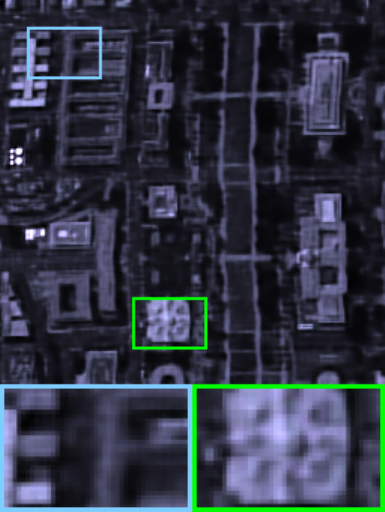}&
			\includegraphics[width=0.106\textwidth]{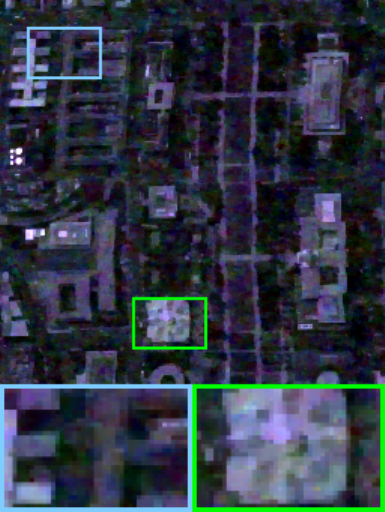}&
			\includegraphics[width=0.106\textwidth]{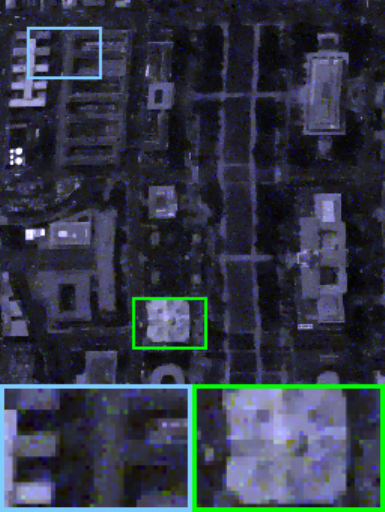}&
			\includegraphics[width=0.106\textwidth]{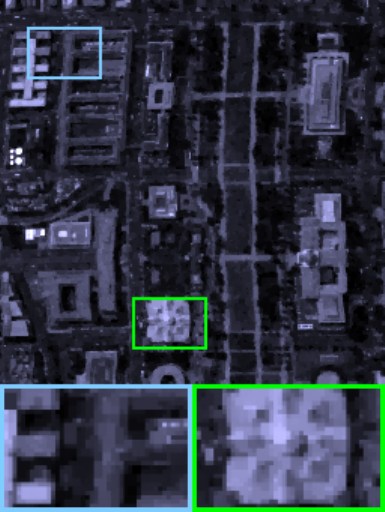}&
			\includegraphics[width=0.106\textwidth]{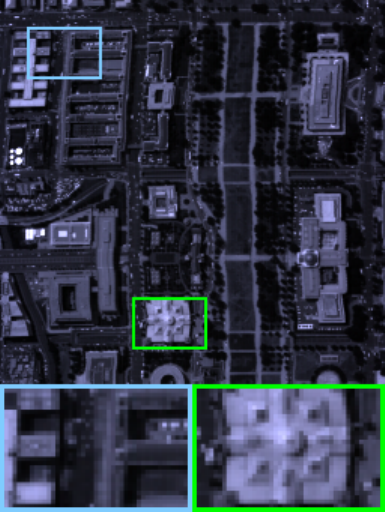}\\
PSNR 13.98 dB &
PSNR 30.42 dB &
PSNR 31.00 dB &
PSNR 31.84 dB &
PSNR 31.53 dB & 
PSNR 28.77 dB & 
PSNR 29.43 dB & 
PSNR 32.28 dB & 
PSNR Inf\\
			\includegraphics[width=0.106\textwidth]{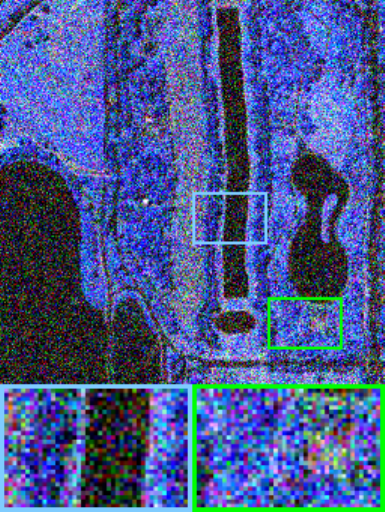}&
			\includegraphics[width=0.106\textwidth]{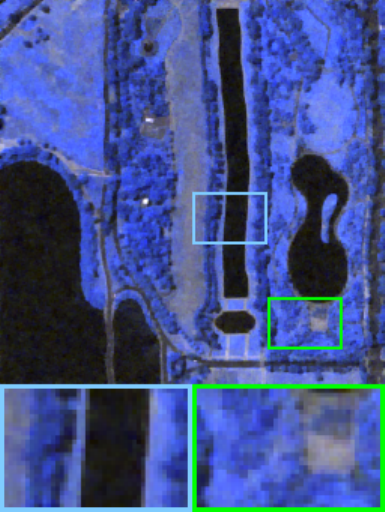}&
						\includegraphics[width=0.106\textwidth]{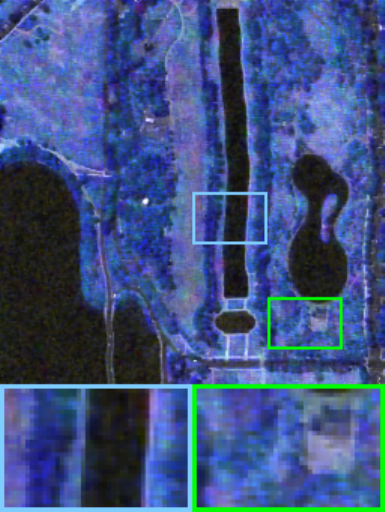}&
						\includegraphics[width=0.106\textwidth]{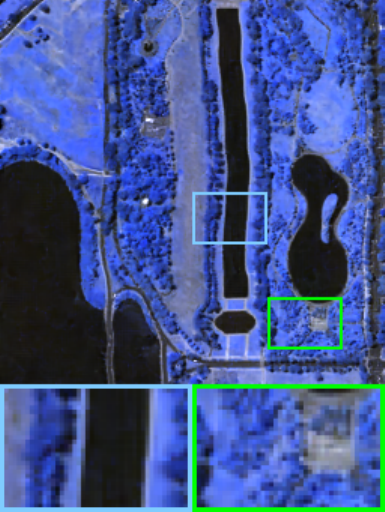}&
			\includegraphics[width=0.106\textwidth]{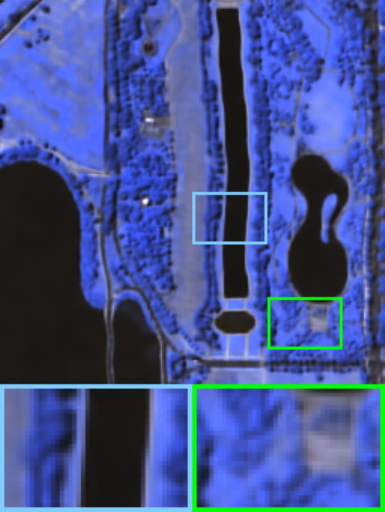}&
			\includegraphics[width=0.106\textwidth]{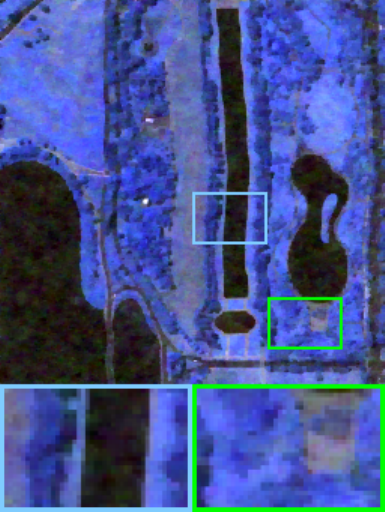}&
			\includegraphics[width=0.106\textwidth]{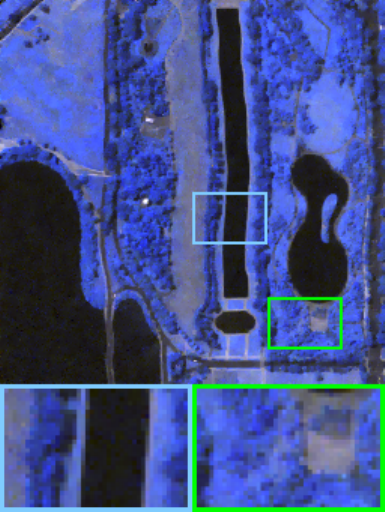}&
			\includegraphics[width=0.106\textwidth]{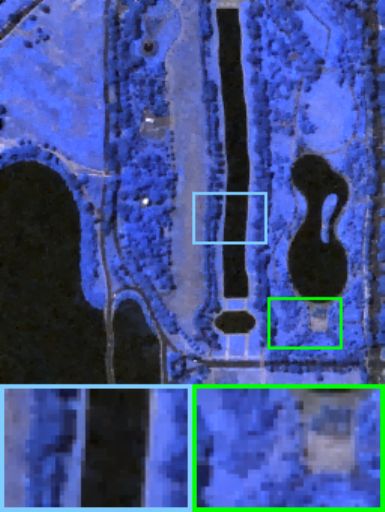}&
			\includegraphics[width=0.106\textwidth]{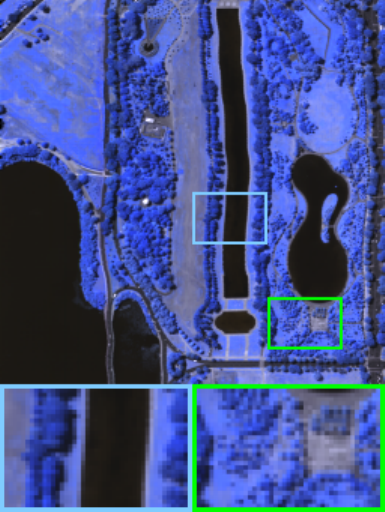}\\
PSNR 16.43 dB &
PSNR 32.52 dB &
PSNR 30.09 dB &
PSNR 33.32 dB &
PSNR 33.52 dB & PSNR 30.25 dB & PSNR 31.77 dB & PSNR 33.55 dB&
PSNR Inf\\
			\includegraphics[width=0.106\textwidth]{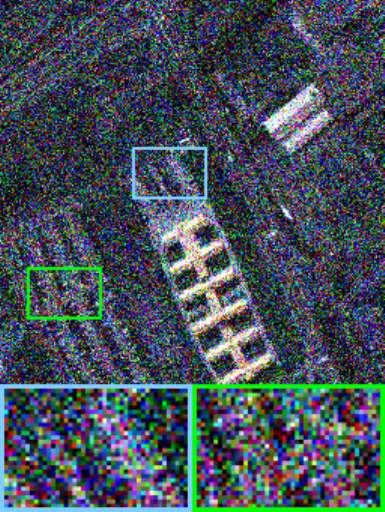}&
			\includegraphics[width=0.106\textwidth]{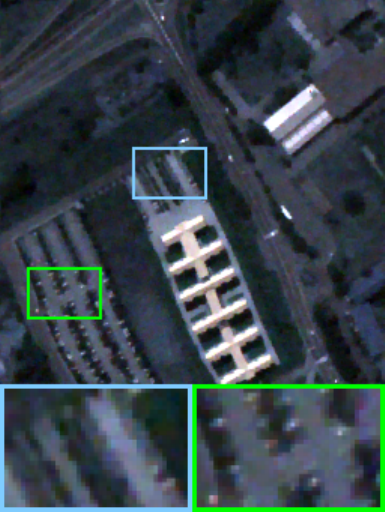}&
						\includegraphics[width=0.106\textwidth]{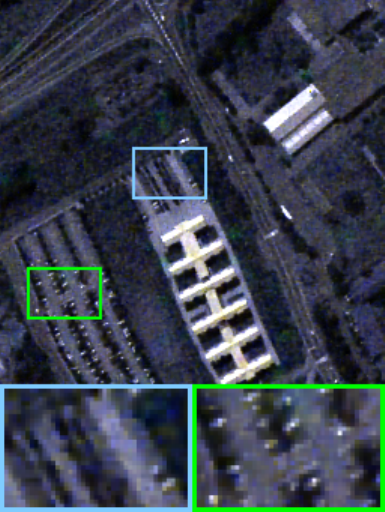}&
						\includegraphics[width=0.106\textwidth]{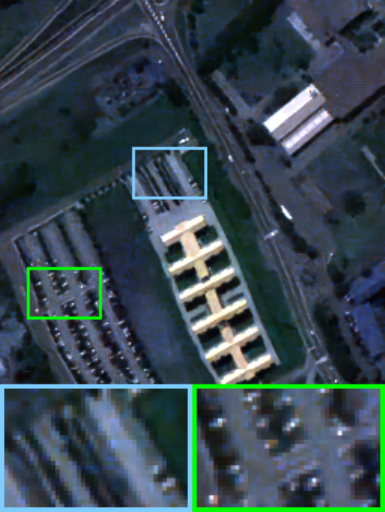}&
			\includegraphics[width=0.106\textwidth]{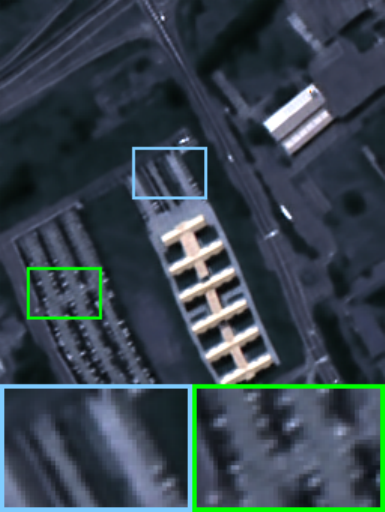}&
			\includegraphics[width=0.106\textwidth]{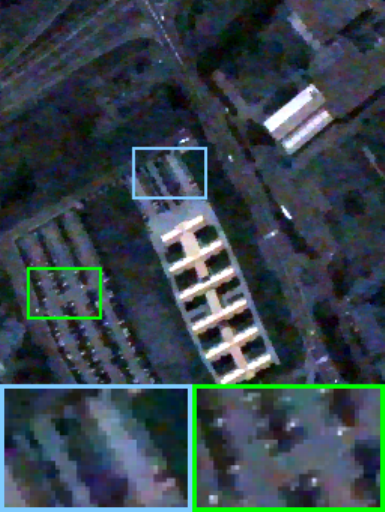}&
			\includegraphics[width=0.106\textwidth]{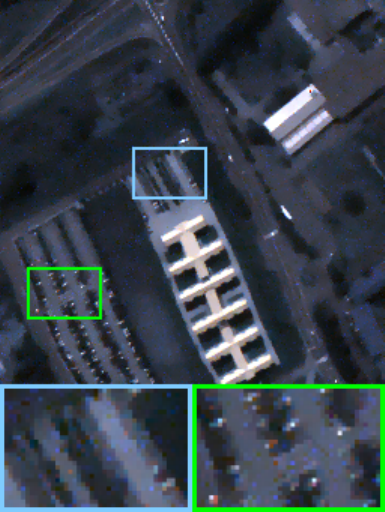}&
			\includegraphics[width=0.106\textwidth]{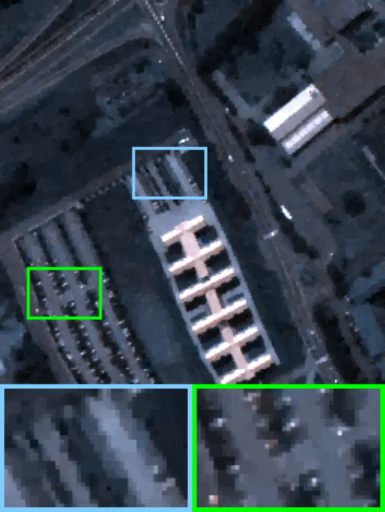}&
			\includegraphics[width=0.106\textwidth]{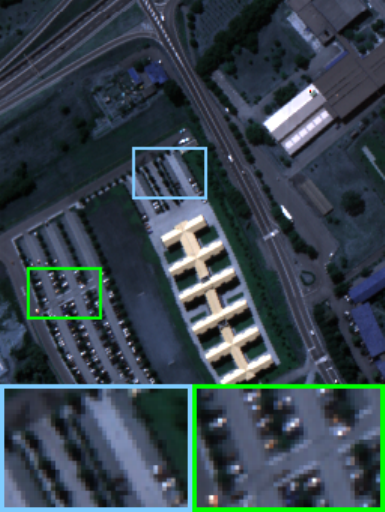}\\
PSNR 13.98 dB &
PSNR 31.44 dB &
PSNR 31.05 dB &
PSNR 33.13 dB &
PSNR 32.84 dB & PSNR 29.50 dB & PSNR 30.81 dB&  PSNR 33.14 dB&
PSNR Inf\\
			\includegraphics[width=0.106\textwidth]{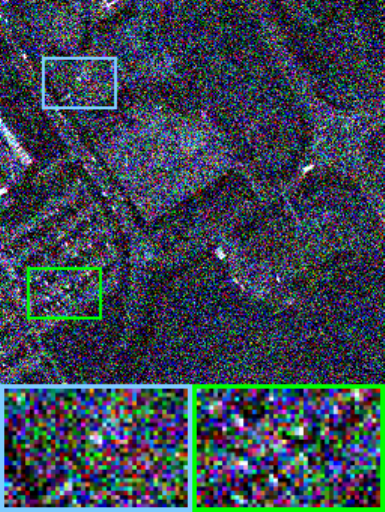}&
			\includegraphics[width=0.106\textwidth]{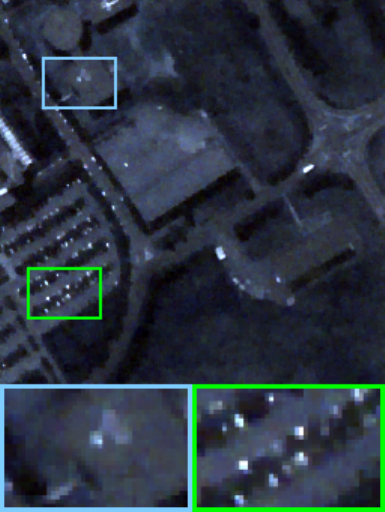}&
						\includegraphics[width=0.106\textwidth]{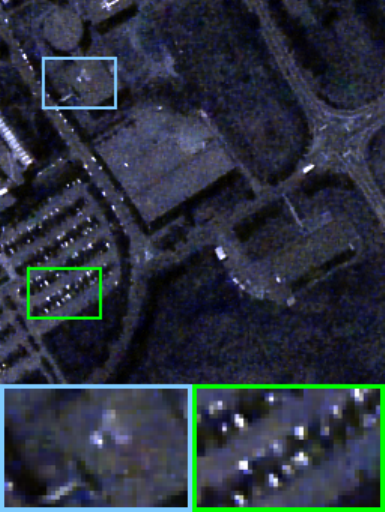}&
						\includegraphics[width=0.106\textwidth]{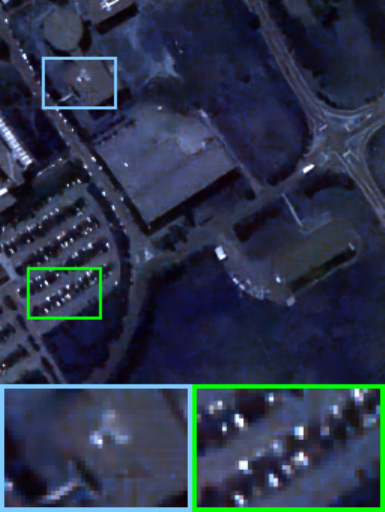}&
			\includegraphics[width=0.106\textwidth]{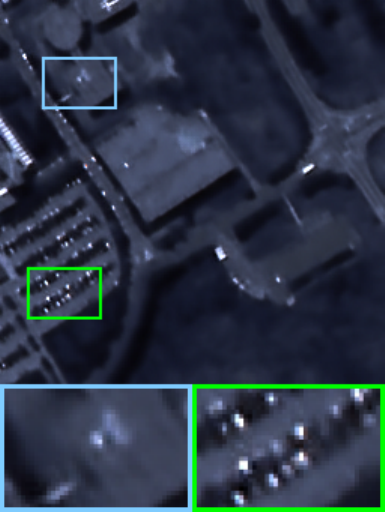}&
			\includegraphics[width=0.106\textwidth]{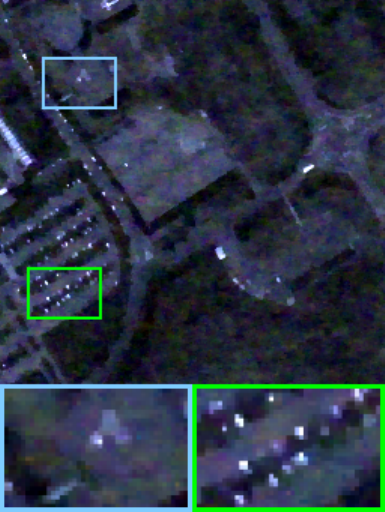}&
			\includegraphics[width=0.106\textwidth]{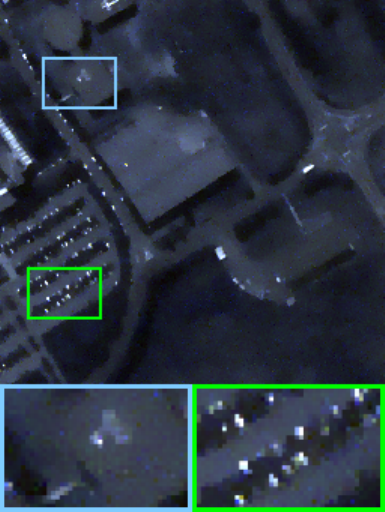}&
			\includegraphics[width=0.106\textwidth]{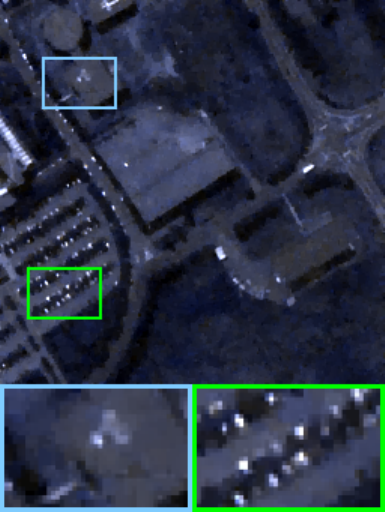}&
			\includegraphics[width=0.106\textwidth]{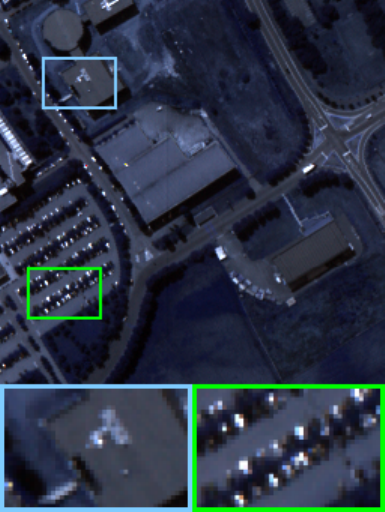}\\
PSNR 13.98 dB &
PSNR 32.15 dB &
PSNR 31.63 dB &
PSNR 32.77 dB &
PSNR 33.13 dB &PSNR 29.67 dB &PSNR 31.42 dB &PSNR 33.47 dB&
PSNR Inf\\
				Observed &LRTDTV \cite{LRTDTV}&HSID-CNN\cite{HSIDCNN}&SDeCNN\cite{SDeCNN}&LTDL \cite{LTDL}&RCTV \cite{RCTV}&WNLRATV \cite{WNLRATV}& CRNL&Original\\
		\end{tabular}
	\end{center}
	\vspace{-0.3cm}
	\caption{From upper to lower: The results of multi-dimensional image denoising by different methods on HSIs {\it WDC-1}, {\it WDC-2}, {\it Pavia U-1}, and {\it Pavia U-2} with noisy level $\sigma =0.2$.\label{fig_denoising_1}}
	\vspace{-0.4cm}
\end{figure*}
\begin{figure*}[t]
	\scriptsize
	\setlength{\tabcolsep}{0.9pt}
	\begin{center}
		\begin{tabular}{ccccccc}
			\includegraphics[width=0.135\textwidth]{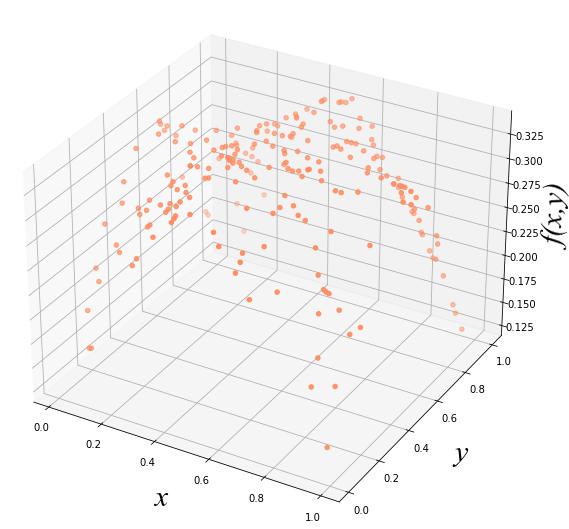}&
			\includegraphics[width=0.135\textwidth]{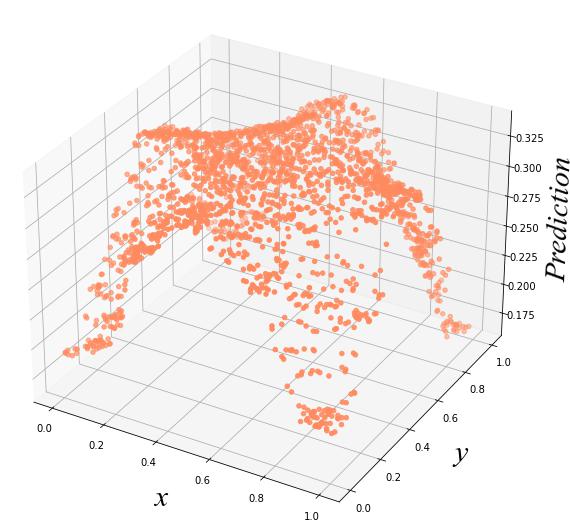}&
			\includegraphics[width=0.135\textwidth]{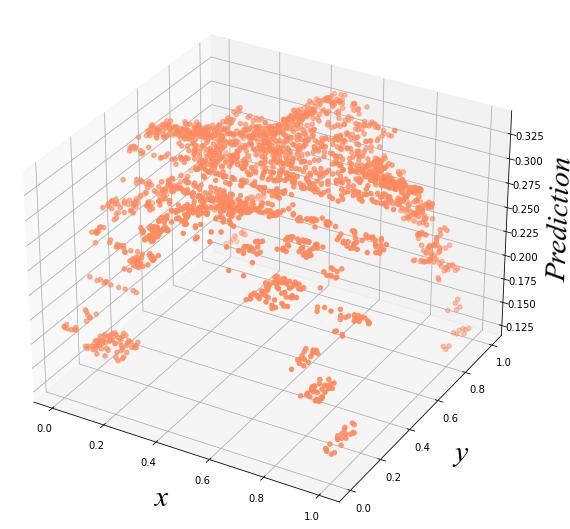}&
			\includegraphics[width=0.135\textwidth]{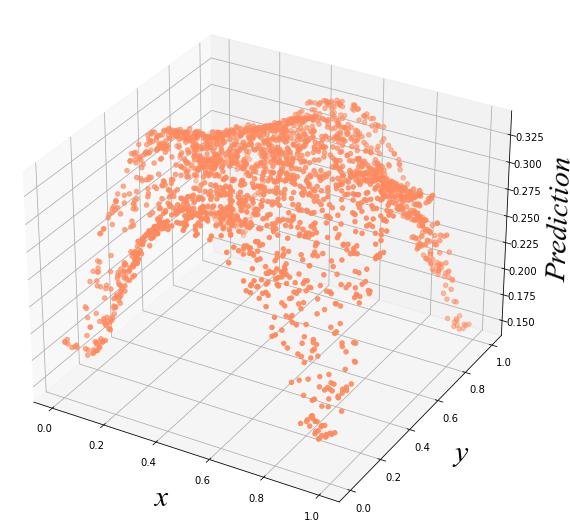}&		
			\includegraphics[width=0.135\textwidth]{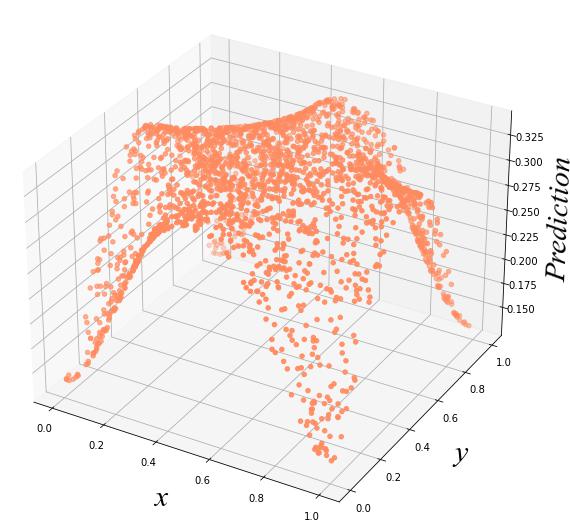}&
			\includegraphics[width=0.135\textwidth]{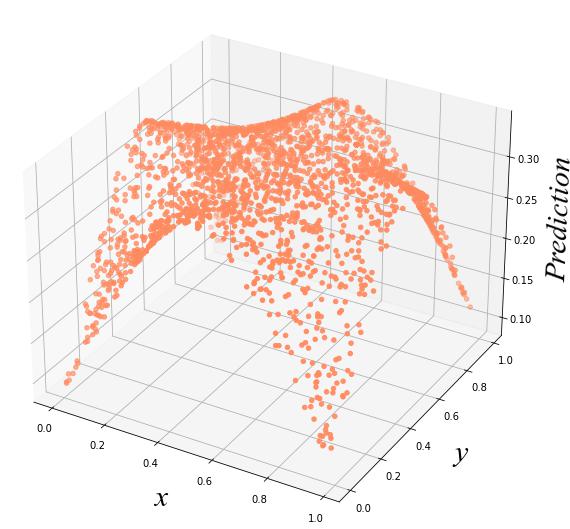}&				
			\includegraphics[width=0.135\textwidth]{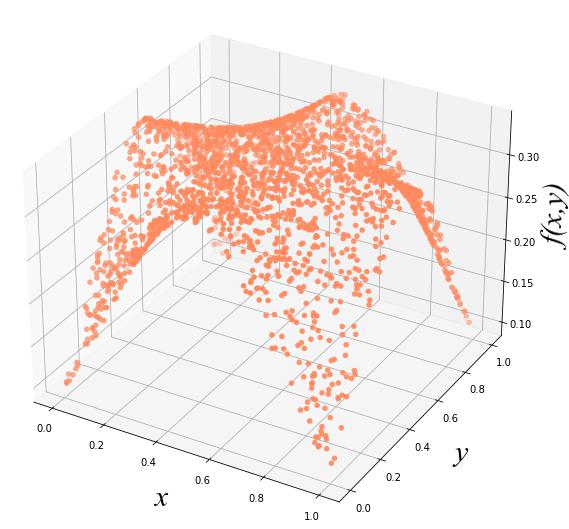}\\
			\includegraphics[width=0.135\textwidth]{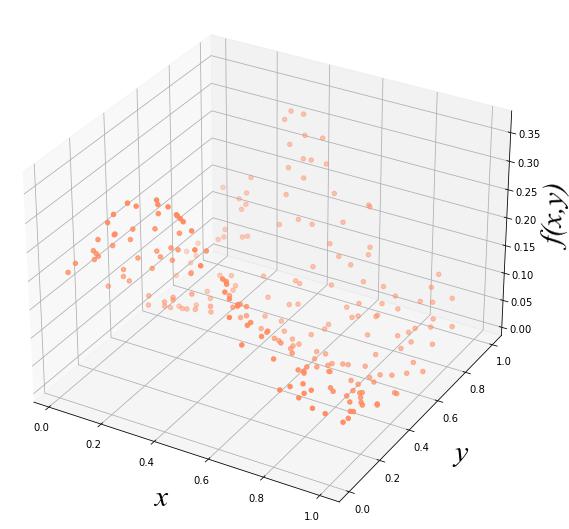}&
			\includegraphics[width=0.135\textwidth]{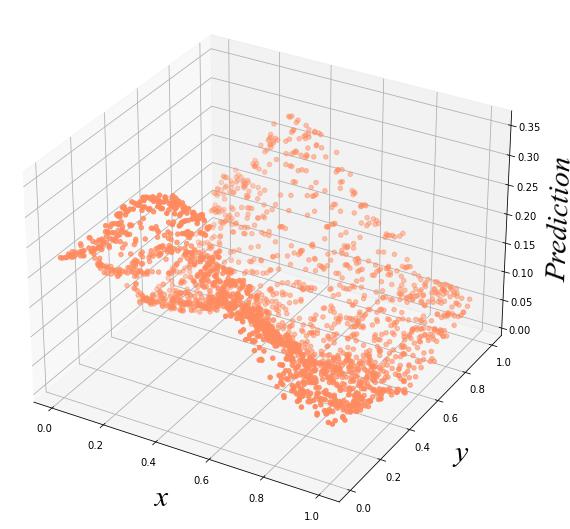}&
			\includegraphics[width=0.135\textwidth]{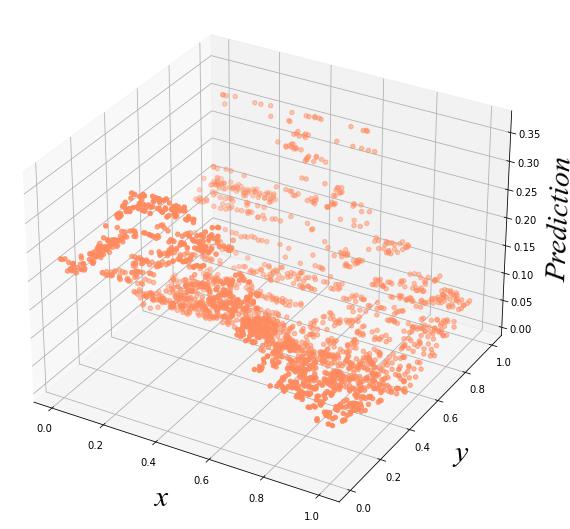}&
			\includegraphics[width=0.135\textwidth]{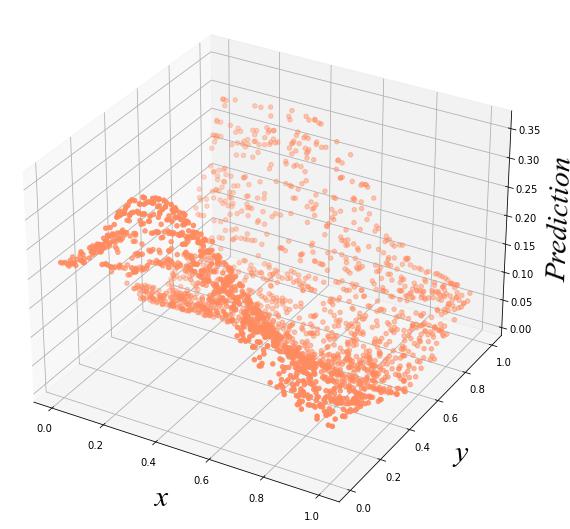}&		
			\includegraphics[width=0.135\textwidth]{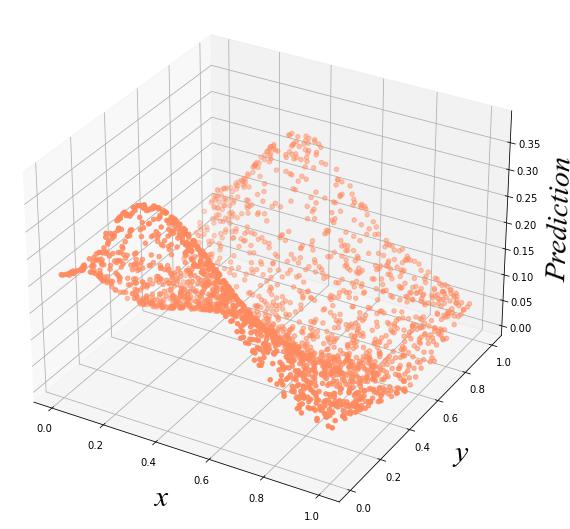}&
			\includegraphics[width=0.135\textwidth]{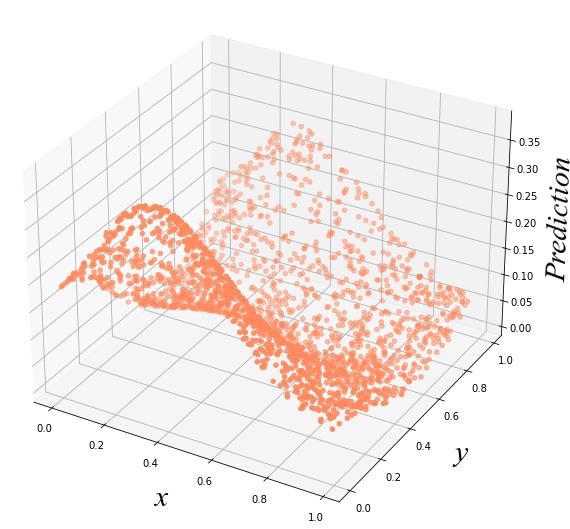}&				
			\includegraphics[width=0.135\textwidth]{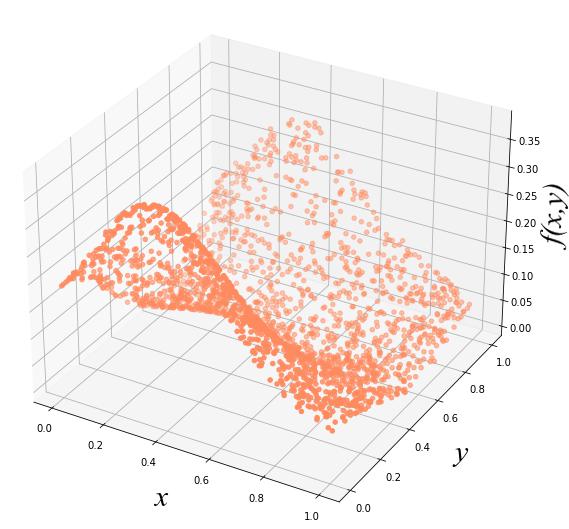}\\
			\includegraphics[width=0.135\textwidth]{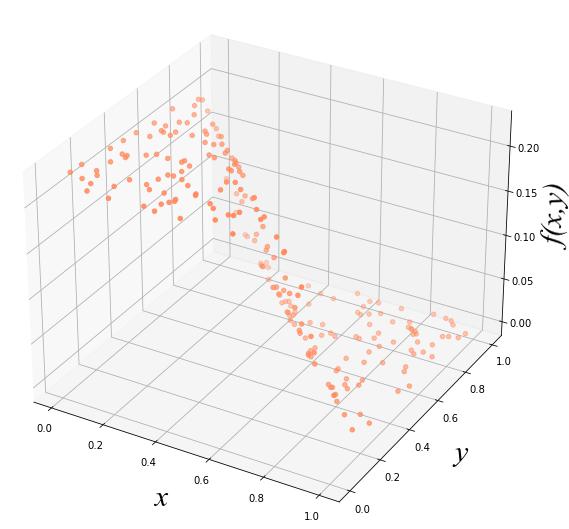}&
			\includegraphics[width=0.135\textwidth]{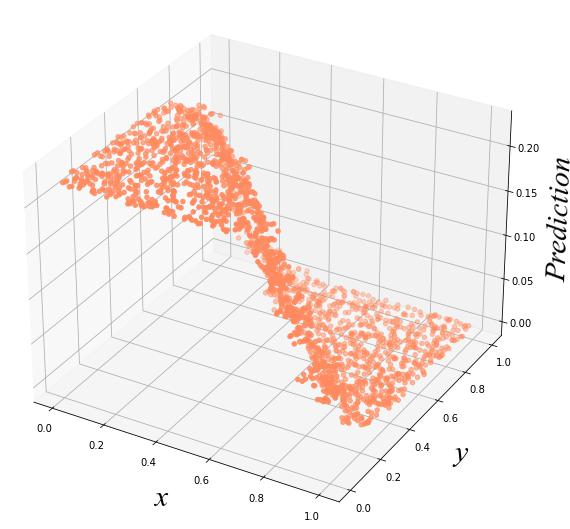}&
			\includegraphics[width=0.135\textwidth]{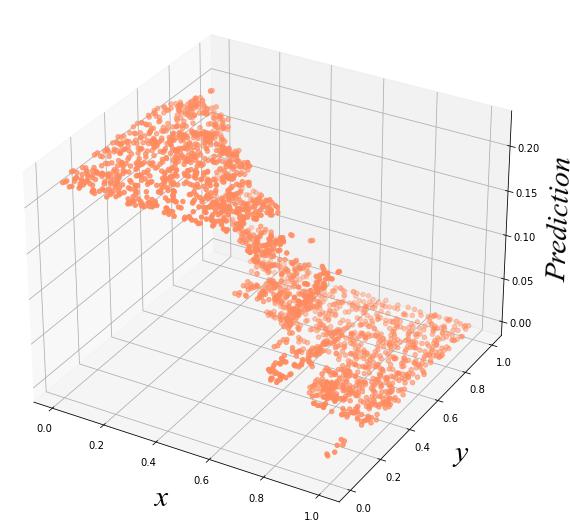}&
			\includegraphics[width=0.135\textwidth]{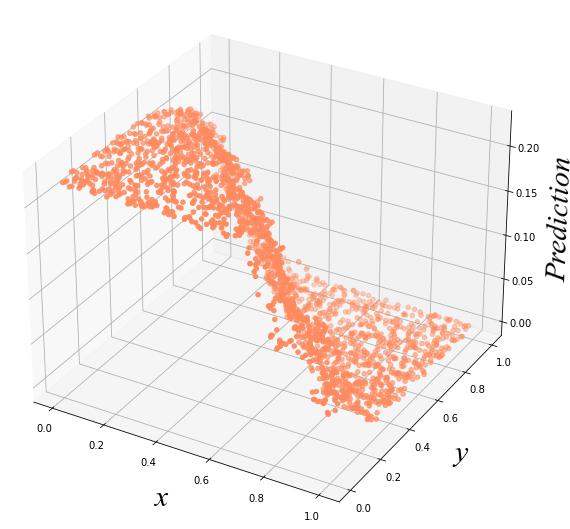}&		
			\includegraphics[width=0.135\textwidth]{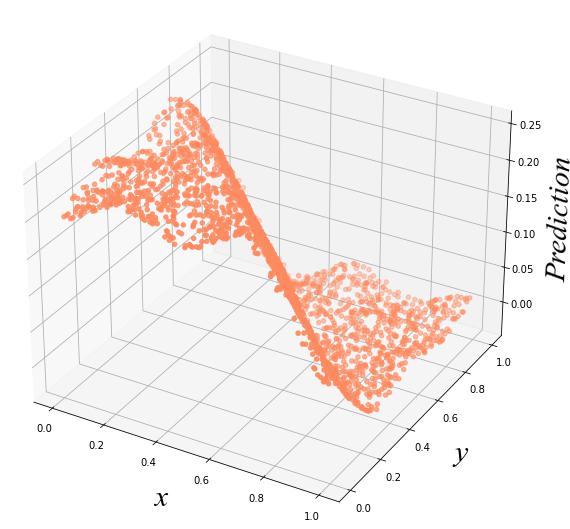}&
			\includegraphics[width=0.135\textwidth]{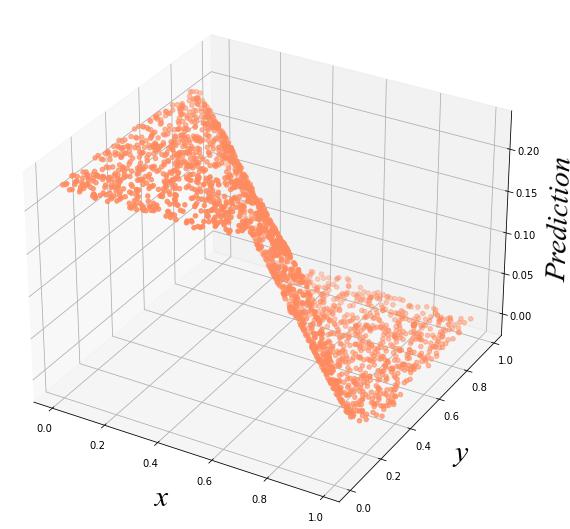}&				
			\includegraphics[width=0.135\textwidth]{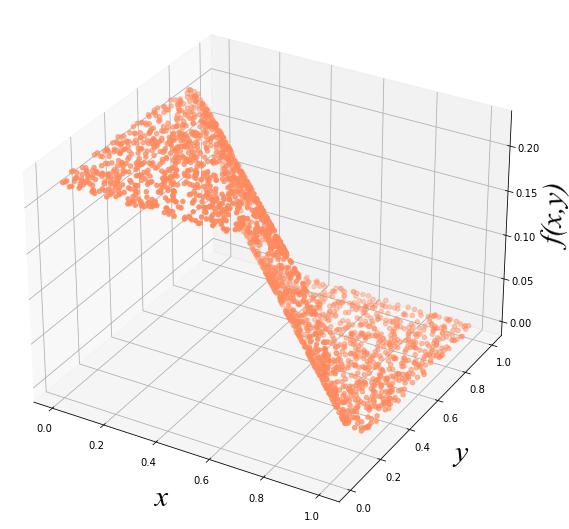}\\
			\includegraphics[width=0.135\textwidth]{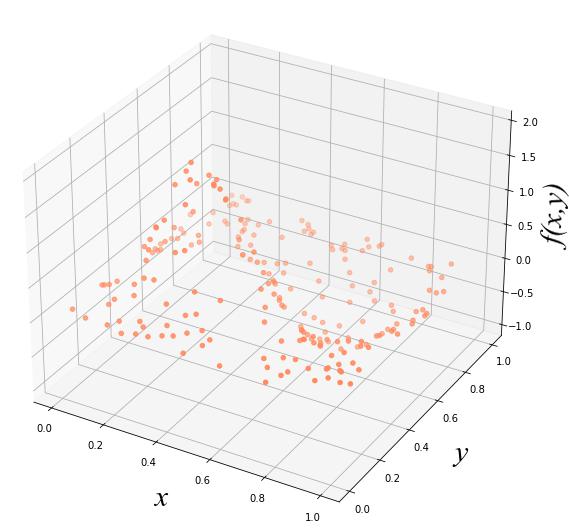}&
			\includegraphics[width=0.135\textwidth]{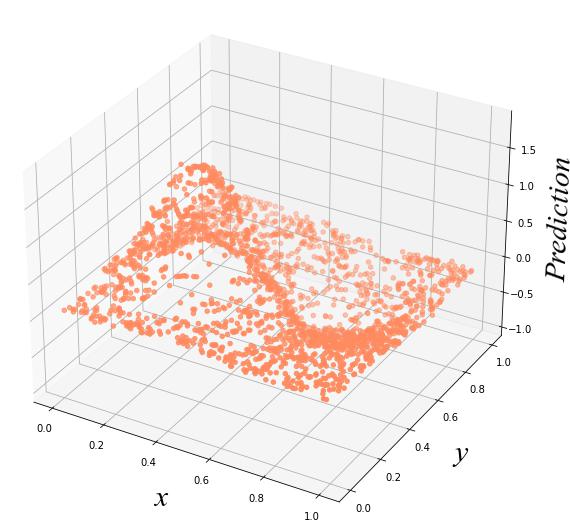}&
			\includegraphics[width=0.135\textwidth]{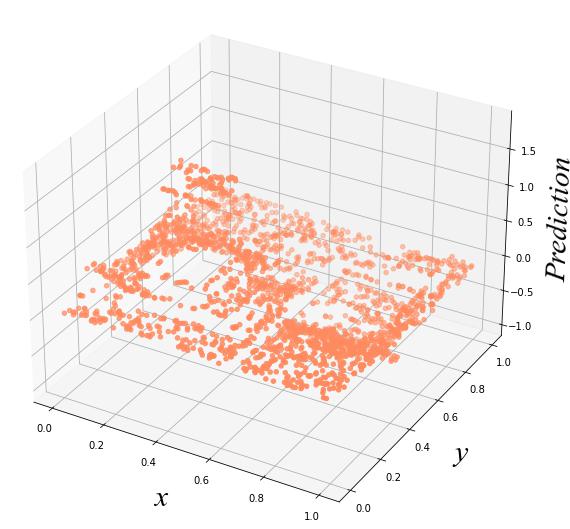}&
			\includegraphics[width=0.135\textwidth]{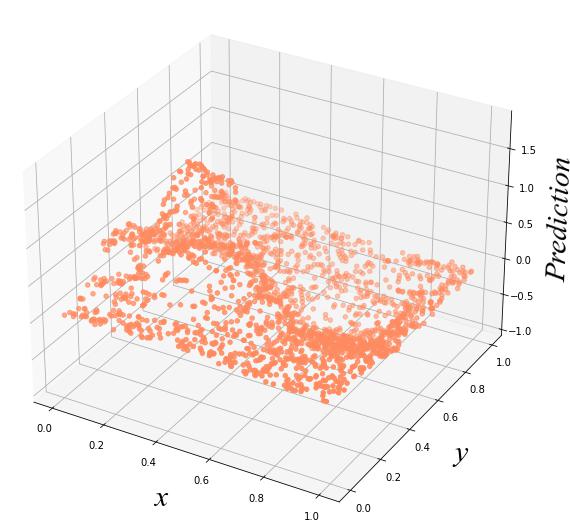}&		
			\includegraphics[width=0.135\textwidth]{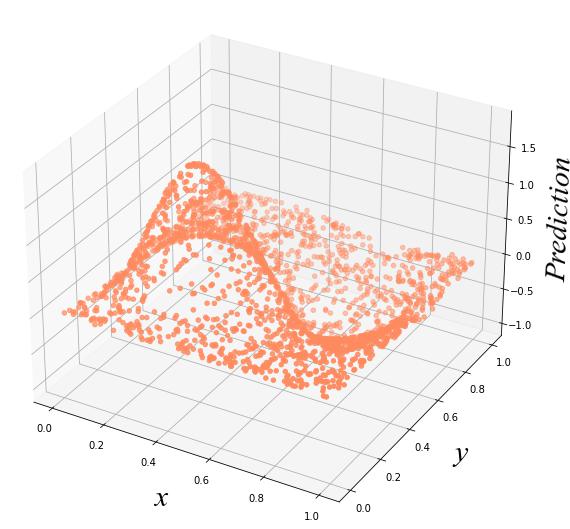}&
			\includegraphics[width=0.135\textwidth]{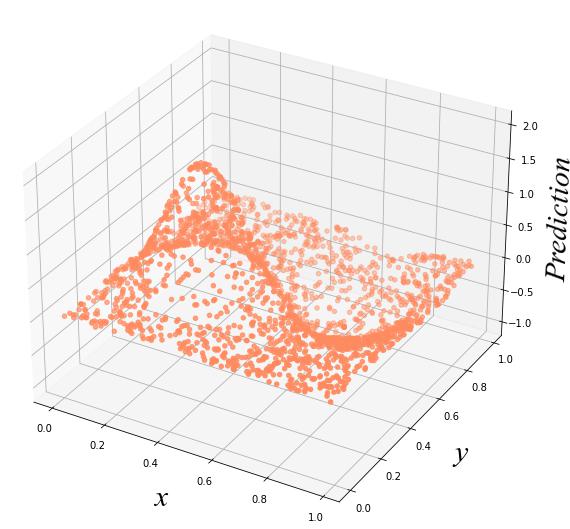}&				
			\includegraphics[width=0.135\textwidth]{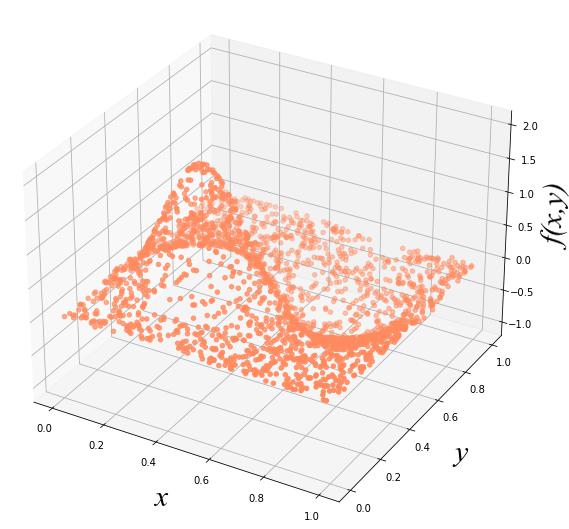}\\
			Observed &KNR&DT&RF&FSA-HTF&CRNL&Original\\
		\end{tabular}
	\end{center}
	\vspace{-0.3cm}
	\caption{From upper to lower: The results of multivariate regression by different methods on $f_1(x,y)$, $f_2(x,y)$, $f_3(x,y)$, and $f_4(x,y)$.\label{fig_regression_1}}
	\vspace{-0.2cm}
\end{figure*}
\begin{table*}[t]
	\caption{The average quantitative results by different methods for multivariate regression. The {\bf best} and \underline{second-best} values are highlighted. (NRMSE $\downarrow$ and R-Square $\uparrow$)\label{tab_regression}}\vspace{-0.4cm}
	\begin{center}
		\scriptsize
		\setlength{\tabcolsep}{1.5pt}
		\begin{spacing}{1.0}
			\begin{tabular}{lccccccccccccccccc}
				\toprule
				\multirow{2}*{\vspace{-0.1cm}Dataset}&\multicolumn{8}{c}{Explicit functions}&\;\;&\multicolumn{8}{c}{Weather data}\\
				%\cmidrule{3-10}\cmidrule{12-19}
				\cmidrule{2-9}\cmidrule{11-18}
				~&\multicolumn{2}{c}{$f_1(x,y)$}&\multicolumn{2}{c}{$f_2(x,y)$}&\multicolumn{2}{c}{$f_3(x,y)$}&\multicolumn{2}{c}{$f_4(x,y)$}&\;\;&\multicolumn{2}{c}{(55$^\circ$N, 117$^\circ$W)}&\multicolumn{2}{c}{(53$^\circ$N, 111$^\circ$W)}&\multicolumn{2}{c}{(50$^\circ$N, 106$^\circ$W)}&\multicolumn{2}{c}{(47$^\circ$N, 100$^\circ$W)}\\
				%\cmidrule{3-10}\cmidrule{12-19}
				\midrule
				Method\;&NRMSE&R-Square\;  &NRMSE&R-Square\; &NRMSE&R-Square\; &NRMSE&R-Square\;&\;\;&NRMSE&R-Square\; &NRMSE&R-Square\; &NRMSE&R-Square\; &NRMSE&R-Square \\
				\midrule
				SVR&{0.267}&{0.243}&{0.499}&{0.911}&{0.603}&{0.940}&{0.248}&{0.946}&\;\;&{0.076}&{0.595}&{0.040}&{0.442}&{0.053}&{0.179}&{0.051}&{0.549}
				\\
				KNR&{0.044}&{0.940}&{0.112}&{0.963}&{0.097}&{0.979}&{0.189}&{0.964}&\;\;&\underline{0.018}&\underline{0.949}&\bf{0.021}&\bf{0.696}&\underline{0.027}&\underline{0.731}&{0.023}&{0.806}
				\\
				DT&{0.072}&{0.820}&{0.152}&{0.938}&{0.190}&{0.919}&{0.315}&{0.896}&\;\;&{0.023}&{0.906}&{0.028}&{0.533}&{0.030}&{0.682}&{0.030}&{0.722}
				\\
				RF&{0.045}&{0.930}&{0.114}&{0.965}&{0.118}&{0.970}&{0.221}&{0.953}&\;\;&{0.018}&{0.945}&{0.024}&{0.674}&{0.025}&{0.760}&\underline{0.022}&\underline{0.817}\\
				FSA-HTF&\underline{0.018}&\underline{0.989}&\underline{0.035}&\underline{0.997}&\underline{0.069}&\underline{0.989}&\underline{0.103}&\underline{0.989}&\;\;&{0.027}&{0.878}&{0.038}&{0.207}&{0.030}&{0.683}&{0.028}&{0.727}
				\\
				CRNL&\bf{0.005}&\bf{1.000}&\bf{0.006}&\bf{1.000}&\bf{0.018}&\bf{0.999}&\bf{0.013}&\bf{1.000}&\;\;&\bf{0.015}&\bf{0.961}&\underline{0.022}&\underline{0.686}&\bf{0.024}&\bf{0.779}&\bf{0.021}&\bf{0.838}\\
				\bottomrule
			\end{tabular}
		\end{spacing}
	\end{center}
	\vspace{-0.7cm}
\end{table*}    
\begin{figure*}[!h]
	\scriptsize
	\setlength{\tabcolsep}{0.9pt}
	\begin{center}
		\begin{tabular}{ccccccc}
			\includegraphics[width=0.135\textwidth]{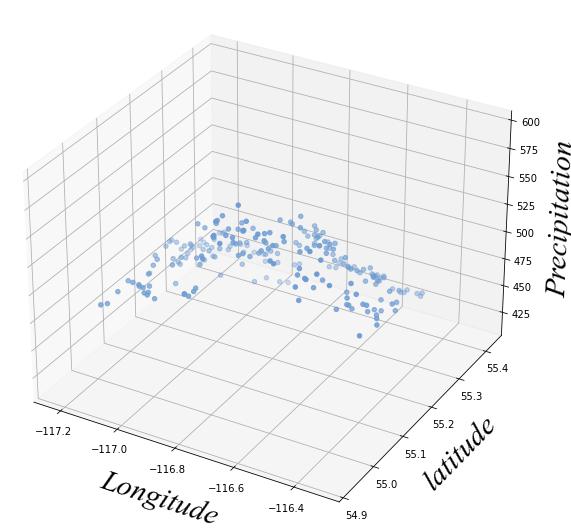}&
			\includegraphics[width=0.135\textwidth]{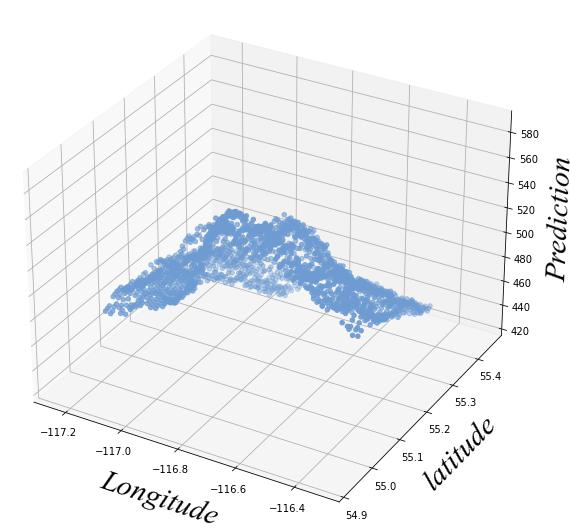}&
			\includegraphics[width=0.135\textwidth]{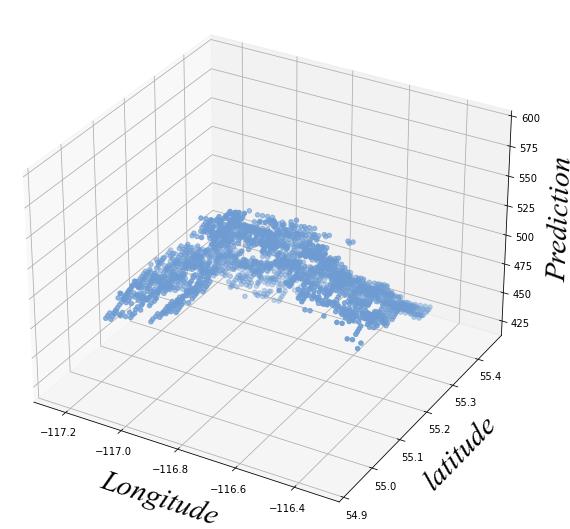}&
			\includegraphics[width=0.135\textwidth]{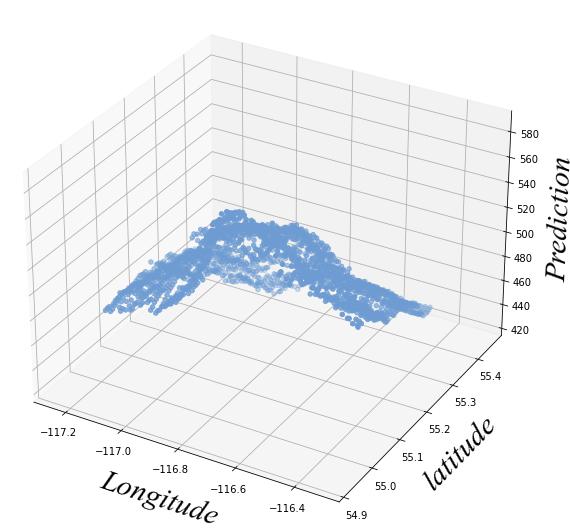}&		
			\includegraphics[width=0.135\textwidth]{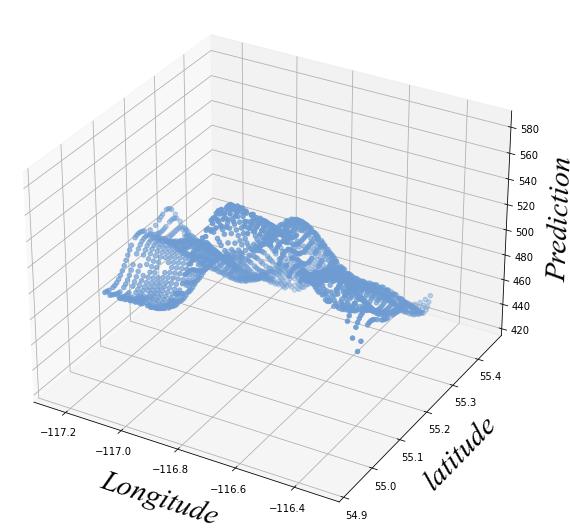}&
			\includegraphics[width=0.135\textwidth]{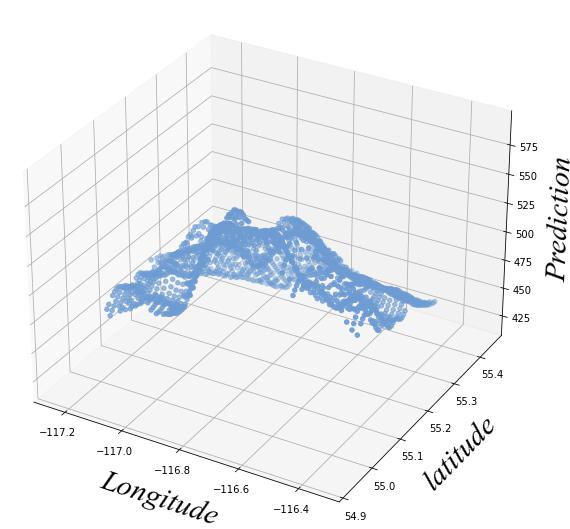}&				
			\includegraphics[width=0.135\textwidth]{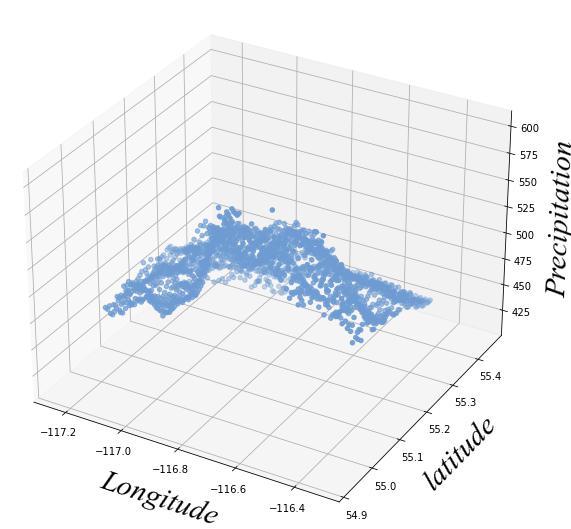}\\
			\includegraphics[width=0.135\textwidth]{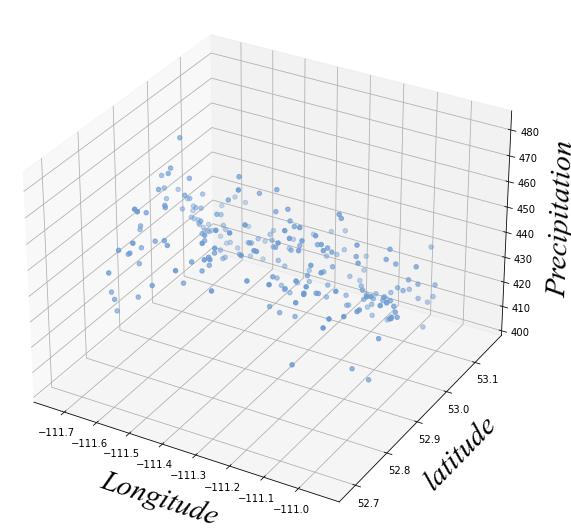}&
			\includegraphics[width=0.135\textwidth]{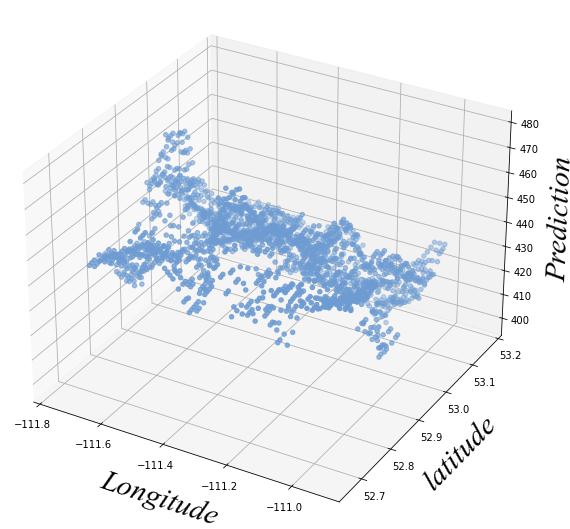}&
			\includegraphics[width=0.135\textwidth]{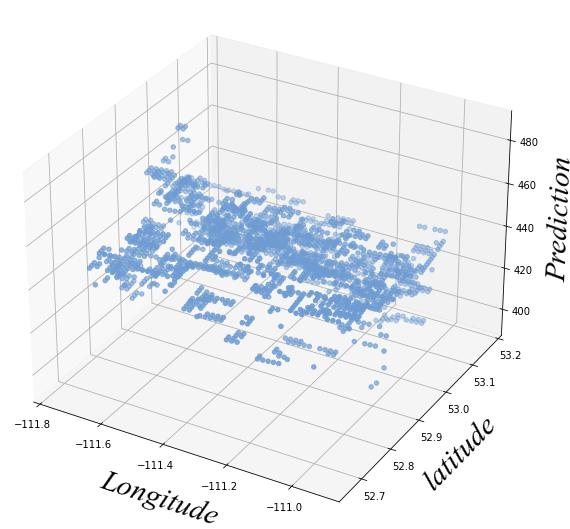}&
			\includegraphics[width=0.135\textwidth]{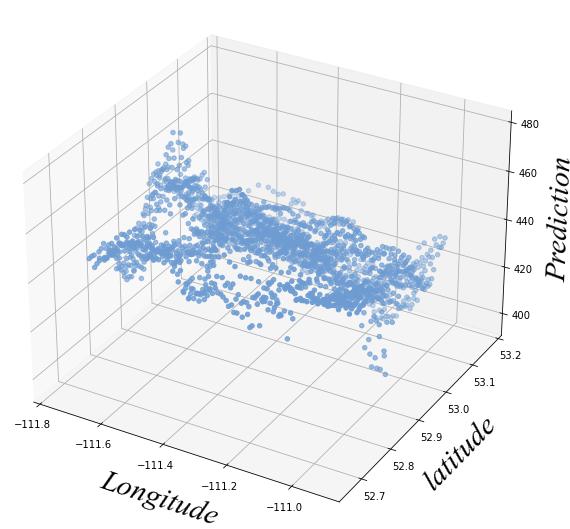}&		
			\includegraphics[width=0.135\textwidth]{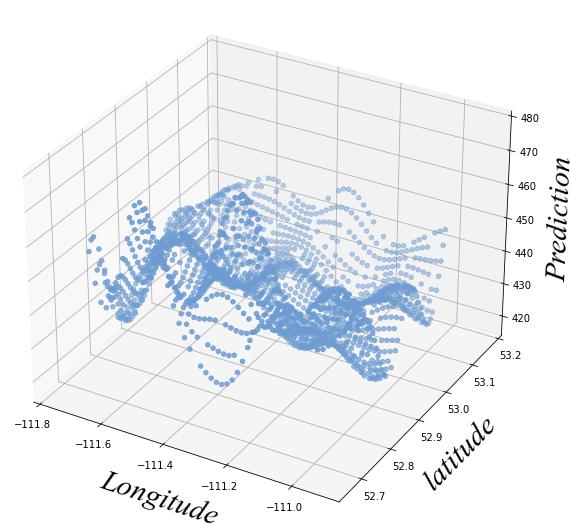}&
			\includegraphics[width=0.135\textwidth]{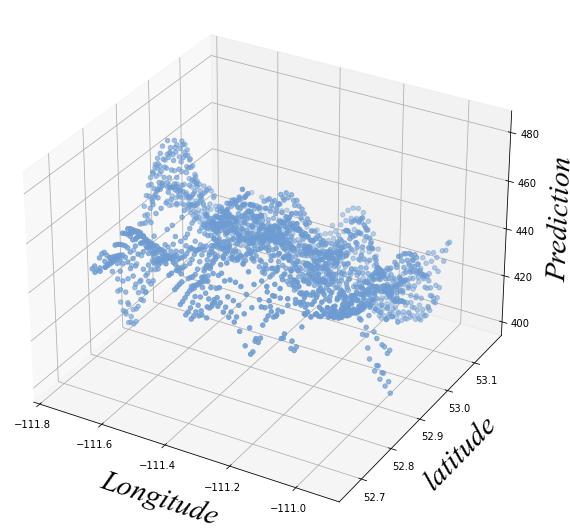}&				
			\includegraphics[width=0.135\textwidth]{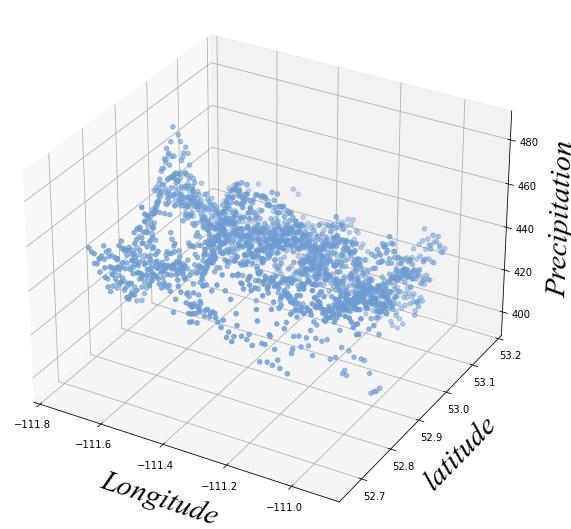}\\
			\includegraphics[width=0.135\textwidth]{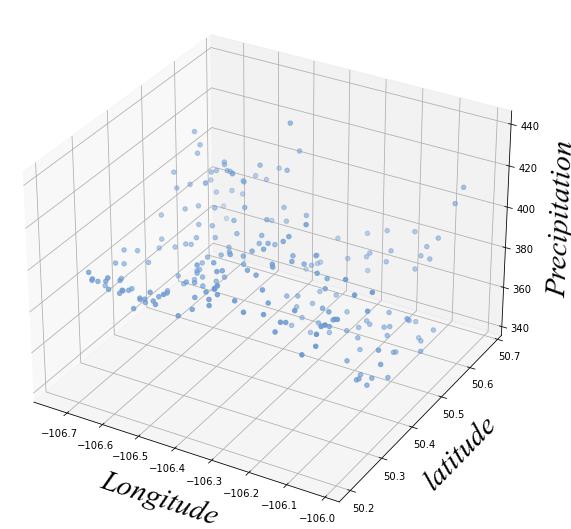}&
\includegraphics[width=0.135\textwidth]{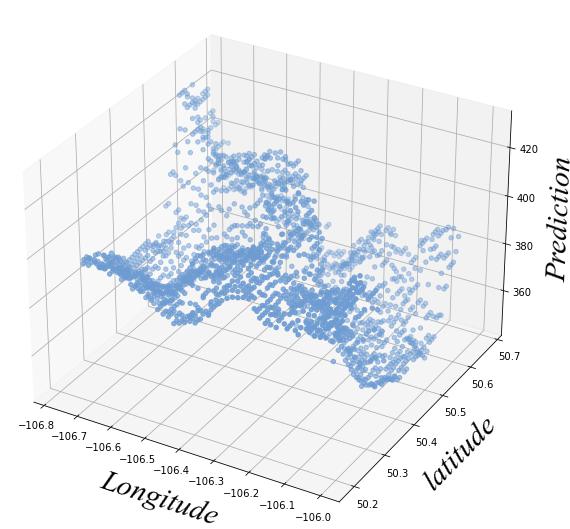}&
\includegraphics[width=0.135\textwidth]{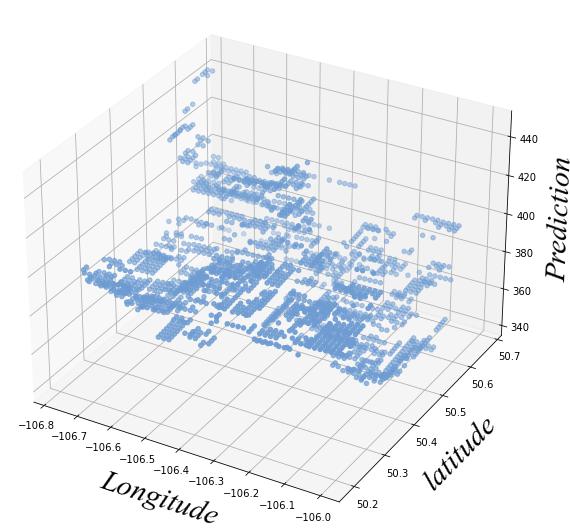}&
\includegraphics[width=0.135\textwidth]{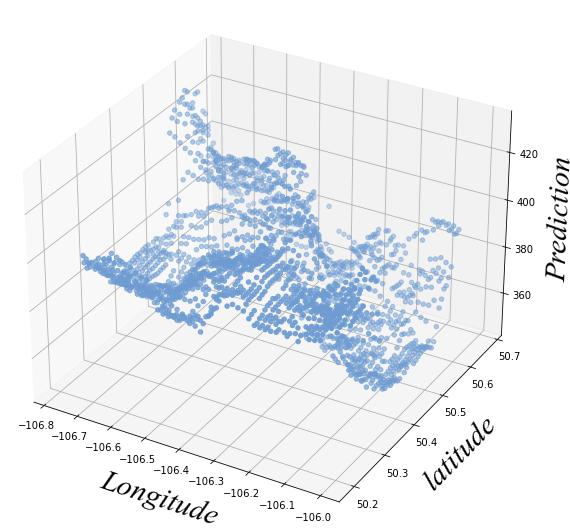}&		
\includegraphics[width=0.135\textwidth]{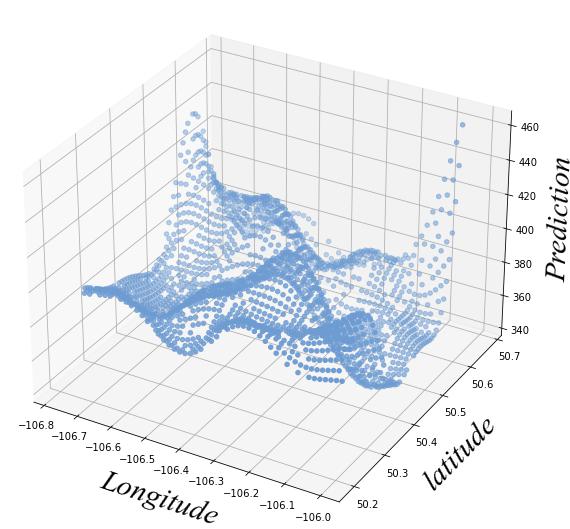}&
\includegraphics[width=0.135\textwidth]{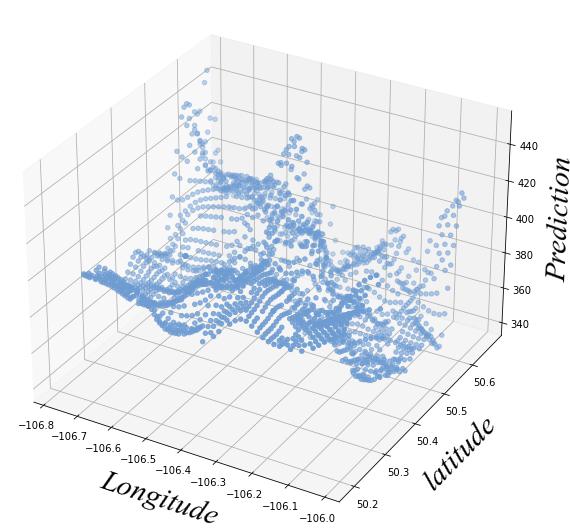}&				
\includegraphics[width=0.135\textwidth]{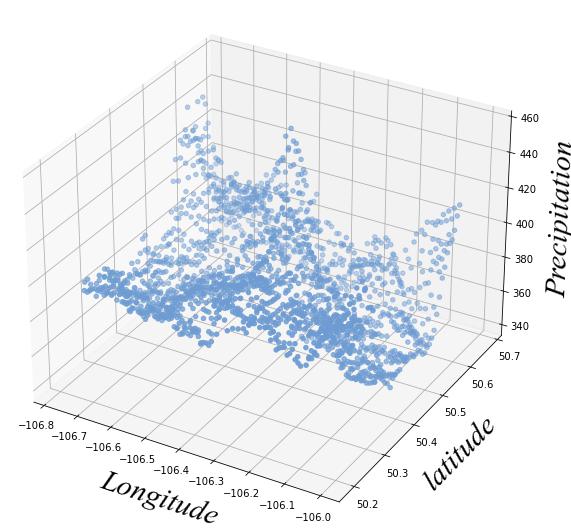}\\
\includegraphics[width=0.135\textwidth]{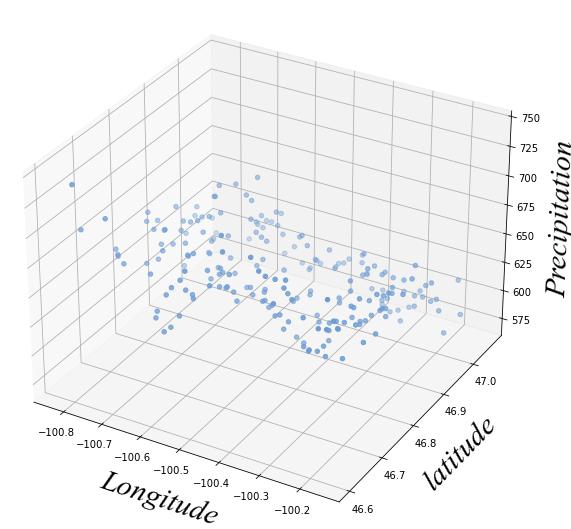}&
\includegraphics[width=0.135\textwidth]{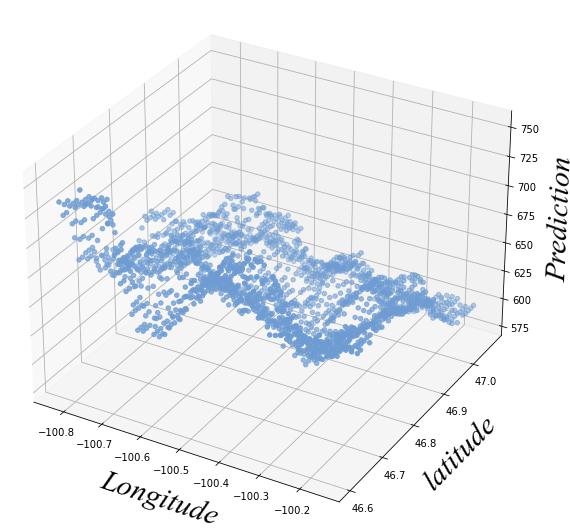}&
\includegraphics[width=0.135\textwidth]{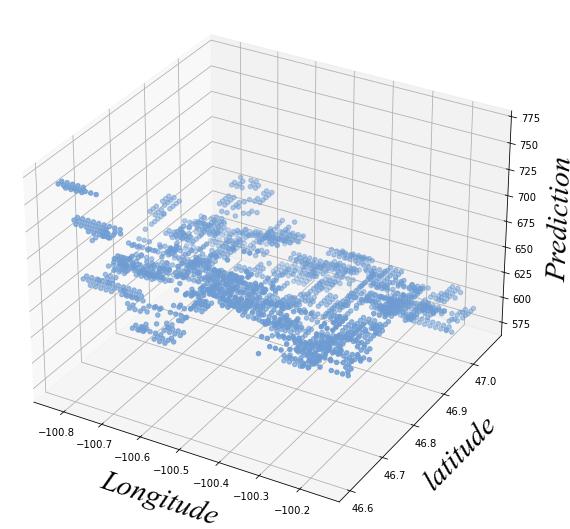}&
\includegraphics[width=0.135\textwidth]{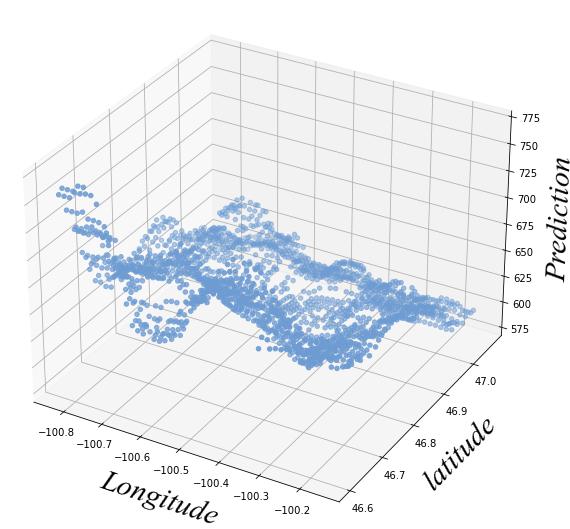}&		
\includegraphics[width=0.135\textwidth]{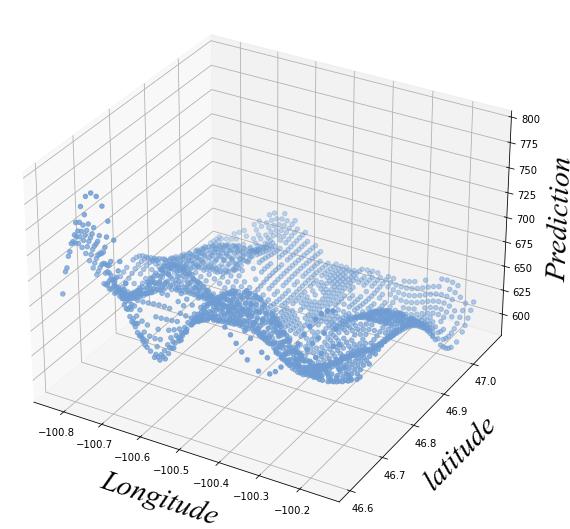}&
\includegraphics[width=0.135\textwidth]{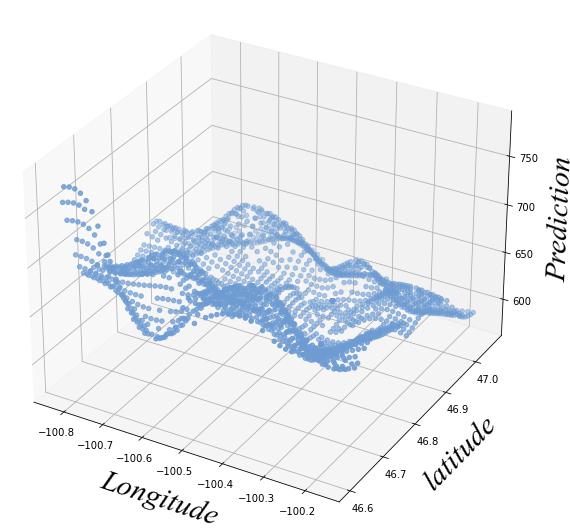}&				
\includegraphics[width=0.135\textwidth]{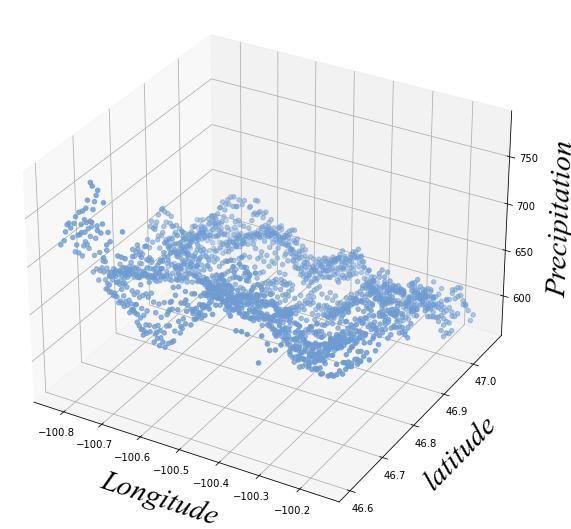}\\
			Observed &KNR&DT&RF&FSA-HTF&CRNL&Original\\
		\end{tabular}
	\end{center}
	\vspace{-0.3cm}
	\caption{From upper to lower: The results of climate data prediction by different methods on the precipitation data at (55$^\circ$N, 117$^\circ$W), (53$^\circ$N, 111$^\circ$W), (50$^\circ$N, 106$^\circ$W), and (47$^\circ$N, 100$^\circ$W).\label{fig_regression_2}}
	\vspace{-0.4cm}
\end{figure*}
\subsection{Hyperparameters Settings}
Next we introduce the hyperparameters setting. In our method, the hyperparameters are the rank of the coupled low-rank function factorization, i.e., $(r_1,r_2,\cdots,r_{N+1})$, the basic continuous cube size, i.e., $p$, and the number of similar cubes, i.e., $S$. We systematically try different hyperparameter values to obtain satisfactory performances. In consequence, we obtain the following hyperparameters settings for different tasks:
\begin{itemize}
	\item For image inpainting, the rank $(r_1,r_2,r_3,r_4)$ is set to $(6,6,3,5)$ for color images and $(6,6,{\rm int}(n_3/3),5)$ for multispectral images and videos, where $n_3$ denotes the size of the third dimension of the observed data. The cube size $p$ is set to 6 for all datasets. The number of similar cubes $S$ is set to 20 for all datasets. The trade-off parameter $\gamma$ of TV regularization is set to $10^{-6}$ for all datasets.
	\item For image denoising, the rank is set to $(6,6,8,1)$ for all datasets. The cube size $p$ is set to 6 for all datasets. The number of similar cubes $S$ is set to 20 for all datasets. The trade-off parameter $\gamma$ of TV regularization is set to $0.6\times 10^{-5}$ for hyperspectral images and $0.9\times 10^{-5}$ for multispectral images.
	\item For multivariate regression, the rank $(r_1,r_2,r_3)$ is set to $(15,15,15)$ for synthetic datasets $f_1(x,y),\cdots,f_4(x,y)$ and climate datasets. For point cloud datasets, the rank $(r_1,r_2,r_3,r_4,r_5)$ is set to $(15,15,15,3,10)$. The cube size $p$ is set to 20 for all datasets. The number of similar cubes $S$ is set to 10 for all datasets.
\end{itemize}
\subsection{Experimental Results}
\subsubsection{Image Inpainting Results}
The quantitative and qualitative results of image inpainting are shown in Table \ref{tab_inpainting} and Figs. \ref{fig_inpainting_1}-\ref{fig_inpainting_2}. We can observe that our CRNL generally obtains the best
quantitative results as compared with other methods, which verifies the effectiveness of our method for image inpainting. Our NSS-based method enjoys relatively short execution time as compared with other image inpainting methods, which validates the favorable computational efficiency of our method. Meanwhile, from the visual results, we can observe that our method can capture fine details of the images better than other methods, which verifies its strong representation abilities.
\subsubsection{Image Denoising Results}
The quantitative and qualitative results of image denoising are shown in Table \ref{tab_denoising} and Figs. \ref{fig_denoising_2}-\ref{fig_denoising_1}. We can observe that our method obtains satisfactory quantitative results as compared with other state-of-the-art methods, including nonlocal low-rank-based methods LTDL and WNLRATV. The superior performance of our CRNL can be rationally attributed to the nonlocal low-rankness and intrinsic similarity within each group and across different groups imposed by our coupled low-rank function factorization, which leads to better noise removal and details recovery. Meanwhile, from the running time comparisons, we can see that our CRNL is quite efficient, especially as compared with classical nonlocal-based methods, e.g., LTDL and WNLRATV, which verifies the favorable computational efficiency of our method due to the coupled function factorization. From the visual comparisons in Figs. \ref{fig_denoising_2}-\ref{fig_denoising_1}, we can see that our CRNL could generally recover fine details and color information of the images better than other methods, and also remove the noise more thoroughly, which validates the effectiveness of our method for image denoising. 
\subsubsection{Multivariate Regression Results}
The good performances of our CRNL on image inpainting and denoising have demonstrated the effectiveness of our method on classical meshgrid problems. Next, we illustrate the results of multivariate regression tasks. The results on synthetic datasets $f_1(x,y),\cdots,f_4(x,y)$ and precipitation climate datasets are shown in Table \ref{tab_regression} and Figs. \ref{fig_regression_1}-\ref{fig_regression_2}. We can observe that our CRNL obtains higher accuracy than other methods in most cases. From the visual results in Figs. \ref{fig_regression_1}-\ref{fig_regression_2}, we can see that the results of the proposed method are generally the closest to the true values, especially on the complex climate datasets in Fig. \ref{fig_regression_2}, where our method could effectively recover the complex structures of real-world data. These results verify the effectiveness and superiority of our CRNL for data processing beyond meshgrid.\par 
\begin{figure*}[!h]
	\scriptsize
	\setlength{\tabcolsep}{0.9pt}
	\begin{center}
		\begin{tabular}{ccccccc}
			\vspace{-0.1cm}
			\includegraphics[width=0.135\textwidth]{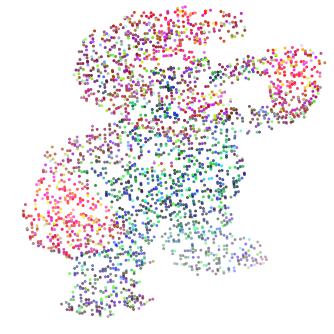}&
\includegraphics[width=0.135\textwidth]{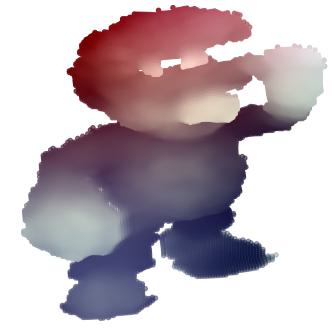}&
\includegraphics[width=0.135\textwidth]{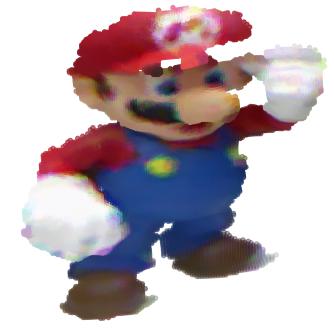}&
\includegraphics[width=0.135\textwidth]{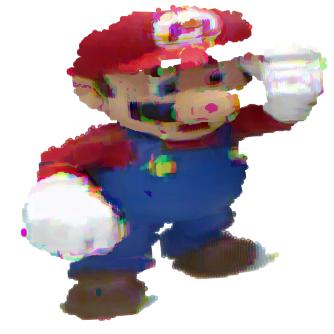}&
\includegraphics[width=0.135\textwidth]{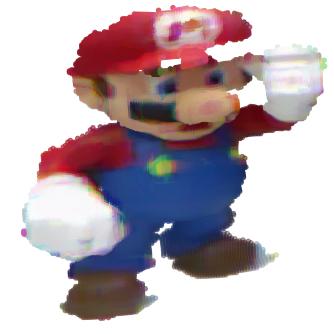}&
\includegraphics[width=0.135\textwidth]{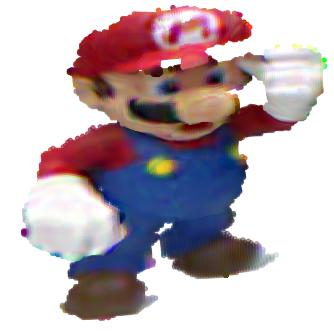}&
\includegraphics[width=0.135\textwidth]{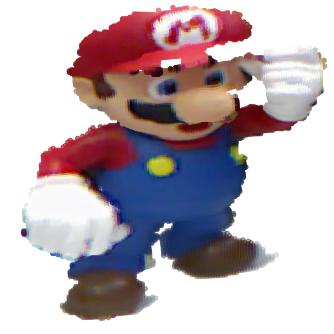}\\
\vspace{-0.1cm}
			\includegraphics[width=0.135\textwidth]{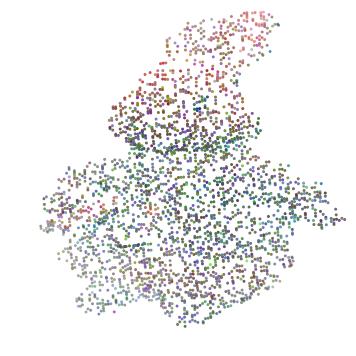}&
			\includegraphics[width=0.135\textwidth]{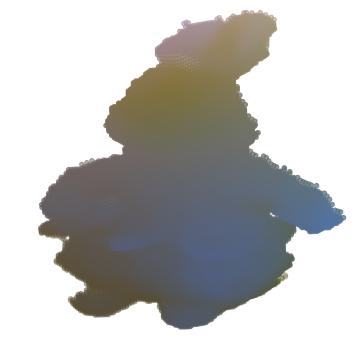}&
			\includegraphics[width=0.135\textwidth]{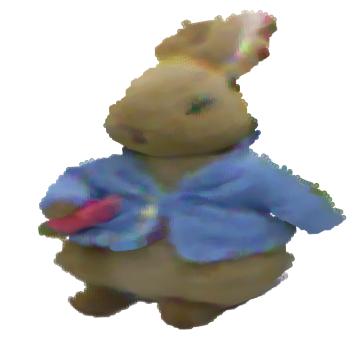}&
			\includegraphics[width=0.135\textwidth]{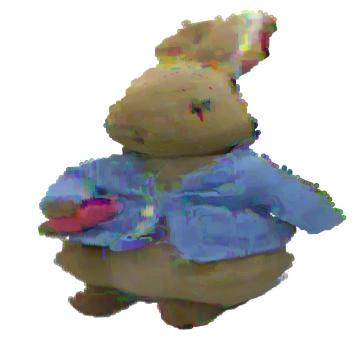}&
			\includegraphics[width=0.135\textwidth]{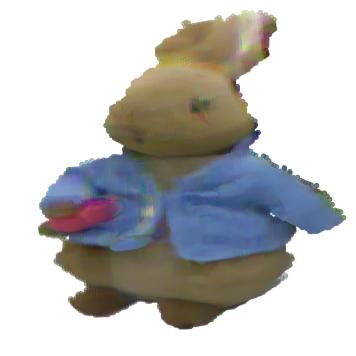}&
			\includegraphics[width=0.135\textwidth]{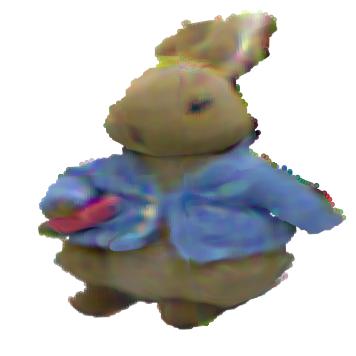}&
			\includegraphics[width=0.135\textwidth]{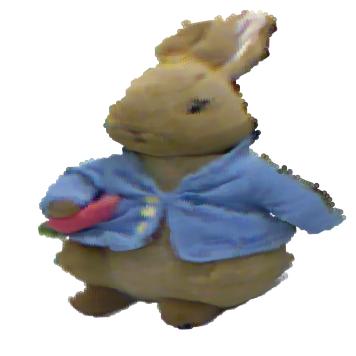}\\
			\vspace{-0.1cm}
			\includegraphics[width=0.135\textwidth]{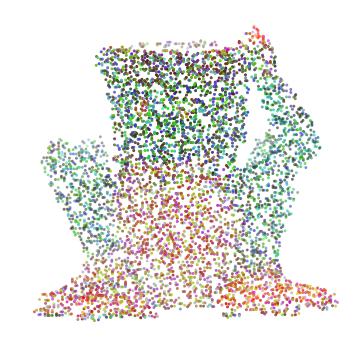}&
\includegraphics[width=0.135\textwidth]{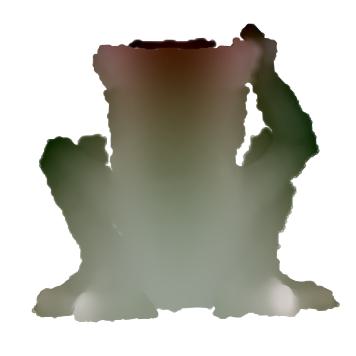}&
\includegraphics[width=0.135\textwidth]{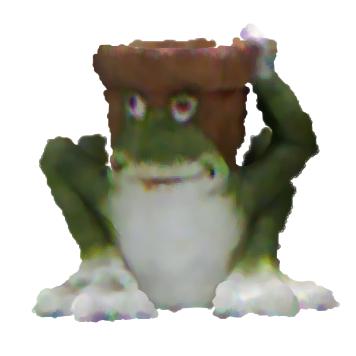}&
\includegraphics[width=0.135\textwidth]{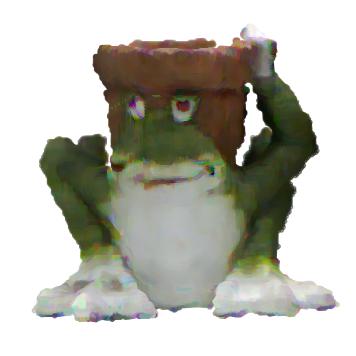}&
\includegraphics[width=0.135\textwidth]{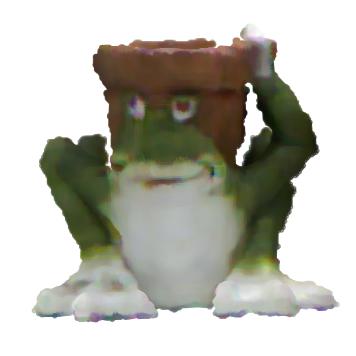}&
\includegraphics[width=0.135\textwidth]{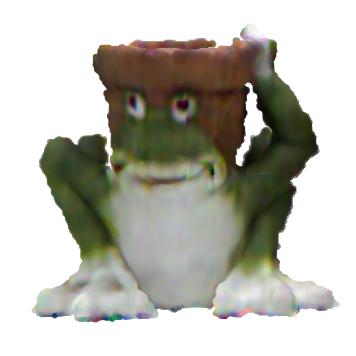}&
\includegraphics[width=0.135\textwidth]{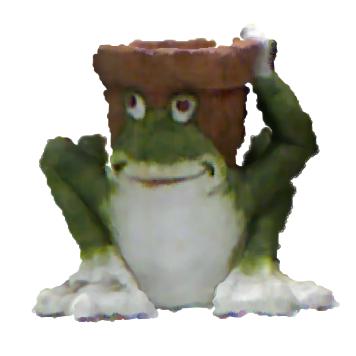}\\
\vspace{-0.2cm}
\includegraphics[width=0.135\textwidth]{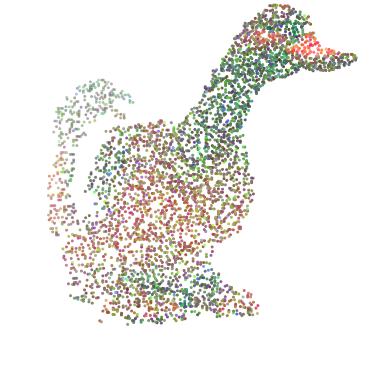}&
\includegraphics[width=0.135\textwidth]{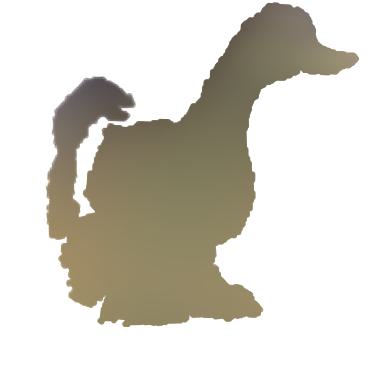}&
\includegraphics[width=0.135\textwidth]{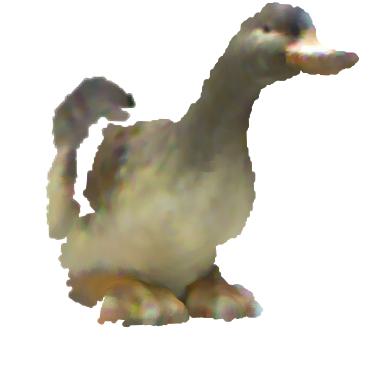}&
\includegraphics[width=0.135\textwidth]{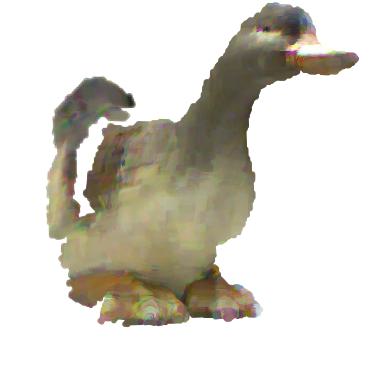}&
\includegraphics[width=0.135\textwidth]{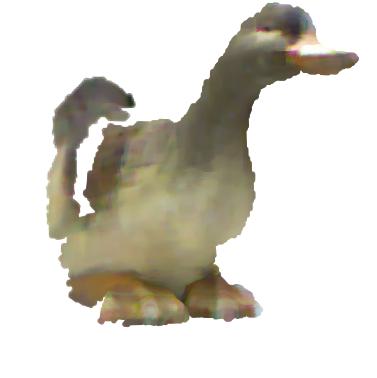}&
\includegraphics[width=0.135\textwidth]{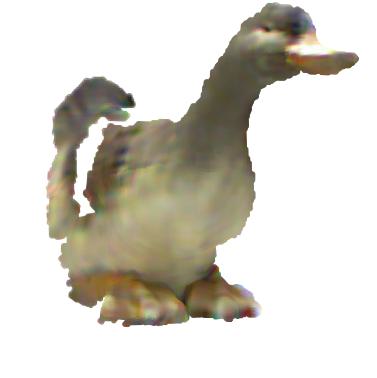}&
\includegraphics[width=0.135\textwidth]{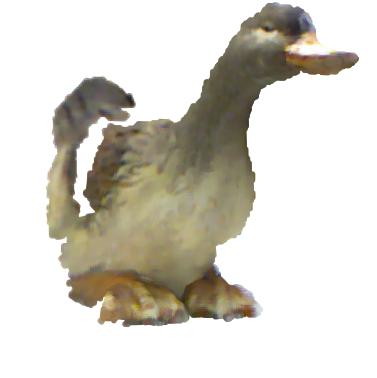}\\
			Observed &FSA-HTF&KNR&DT&RF&CRNL&Original\\
		\end{tabular}
	\end{center}
	\vspace{-0.3cm}
	\caption{From upper to lower: The results of point cloud recovery by different methods on {\it Mario}, {\it Rabbit}, {\it Frog}, and {\it Duck}.\label{fig_regression_3}}
	\vspace{-0.4cm}
\end{figure*}
The results of point cloud recovery are shown in Table \ref{tab_regression_2} and Fig. \ref{fig_regression_3}. Again we can see that our CRNL generally attains better performances than other methods from both quantitative and qualitative perspectives. While FSA-HTF is valid for the former synthetic datasets and climate datasets, it fails to well recover the complex point cloud structures, which is possibly due to its utilized shallow Fourier basis functions. As compared, our CRNL leverages the powerful representation abilities of INRs (i.e., deep neural networks) to parameterize factor functions, which could better capture complex structures of point clouds. The good performances of our method can also be attributed to its elaborate exploration on the nonlocal low-rankness and the similarity within each group and across different groups imposed by the nonlocal coupled low-rank function factorization. In general, the satisfactory performances of our method on different multivariate regression tasks have demonstrated the effectiveness of our method beyond meshgrid.
\begin{table}[t]
	\caption{The average quantitative results by different methods for point cloud recovery. The {\bf best} and \underline{second-best} values are highlighted. (NRMSE $\downarrow$ and R-Square $\uparrow$)\label{tab_regression_2}}\vspace{-0.4cm}
	\begin{center}
		\scriptsize
		\setlength{\tabcolsep}{0.6pt}
		\begin{spacing}{1.0}
			\begin{tabular}{lcccccccc}
				\toprule
				Dataset&\multicolumn{2}{c}{{\it Mario}}&\multicolumn{2}{c}{{\it Rabbit}}&\multicolumn{2}{c}{{\it Frog}}&\multicolumn{2}{c}{\it Duck}\\
				%\cmidrule{3-10}\cmidrule{12-19}
				\midrule
				Method\;&NRMSE&R-Square\;  &NRMSE&R-Square\; &NRMSE&R-Square\; &NRMSE&R-Square\\
				\midrule
				SVR&{0.505}&{0.207}&{0.318}&{0.195}&{0.365}&{0.502}&{0.301}&{0.062}\\
				
				KNR&\underline{0.134}&\underline{0.936}&\underline{0.105}&\underline{0.891}&{0.082}&{0.972}&{0.086}&\underline{0.922}\\
				
				DT&{0.185}&{0.881}&{0.148}&{0.801}&{0.103}&{0.957}&{0.111}&{0.875}\\
				
				RF&{0.139}&{0.931}&{0.108}&{0.887}&\underline{0.077}&\underline{0.976}&\underline{0.085}&\bf{0.925}\\
				
				FSA-HTF&{0.423}&{0.367}&{0.281}&{0.288}&{0.375}&{0.418}&{0.286}&{0.142}\\
				
				CRNL&\bf{0.131}&\bf{0.937}&\bf{0.097}&\bf{0.913}&\bf{0.072}&\bf{0.978}&\bf{0.078}&\underline{0.922}\\
				\bottomrule
			\end{tabular}
		\end{spacing}
	\end{center}
	\vspace{-0.5cm}
\end{table}    
\begin{table}[t]
	\caption{The results of image inpainting on color images with sampling rate 0.2 and the corresponding running time (second) by different NSS-based methods.\label{tab_share}}\vspace{-0.4cm}
	\begin{center}
		\scriptsize
		\setlength{\tabcolsep}{3.2pt}
		\begin{spacing}{1.0}
			\begin{tabular}{lcccccccc}
				\toprule
				Dataset&\multicolumn{2}{c}{{\it Peppers}}&\multicolumn{2}{c}{{\it Plane}}&\multicolumn{2}{c}{{\it Sailboat}}&\multicolumn{2}{c}{\it House}\\
				%\cmidrule{3-10}\cmidrule{12-19}
				\midrule
				Method\;&PSNR&Time\;  &PSNR&Time\; &PSNR&Time\; &PSNR&Time\\
				\midrule
				NL-CP&{24.88}&{9.68}&{30.04}&{9.54}&{23.16}&{9.36}&{22.78}&{10.09}\\
				
				NL-Tucker&{22.80}&{11.58}&{28.31}&{11.77}&{22.37}&{11.68}&{21.17}&{11.71}\\
								
				NL-FCTN&{22.13}&{14.09}&{29.95}&{14.10}&{23.26}&{17.01}&{22.52}&{15.04}\\
				
				CRNL (uncoupled)&{25.11}&{15.66}&{30.13}&{15.64}&{23.38}&{15.55}&{22.79}&{14.50}\\
				
				CRNL (coupled)&{25.40}&{5.73}&{30.60}&{5.97}&{23.67}&{5.31}&{23.32}&{5.51}\\
				\bottomrule
			\end{tabular}
		\end{spacing}
	\end{center}
	\vspace{-0.3cm}
\end{table}    
\iffalse
\begin{table}[t]
	\caption{The results of image inpainting on color images with sampling rate 0.2 and the corresponding running time (second) by different NSS-based methods.\label{tab_share}}\vspace{-0.4cm}
	\begin{center}
		\scriptsize
		\setlength{\tabcolsep}{7pt}
		\begin{spacing}{0.8}
			\begin{tabular}{lcccc}
				\toprule
				Method&{{\it Peppers}}&{{\it Plane}}&{{\it Sailboat}}&{\it House}\\
				%\cmidrule{3-10}\cmidrule{12-19}
				\midrule
				NL-CP&{9.68}&{9.54}&{9.36}&{10.09}\\
				NL-Tucker&{11.58}&{11.77}&{11.68}&{11.71}\\
				NL-FCTN&{14.09}&{14.10}&{17.01}&{15.04}\\
				Our CRNL&{5.73}&{5.97}&{5.31}&{5.51}\\
				\bottomrule
			\end{tabular}
		\end{spacing}
	\end{center}
	\vspace{-0.3cm}
\end{table}    
\fi
\subsection{Discussions}
\subsubsection{Advantages of Coupled Function Factorization}\label{sec_relation}
As compared with classical NSS-based methods, an important innovation of our method is that we propose the coupled low-rank function factorization to compactly and accurately represent continuous groups with advantageous computational efficiency. To validate these advantages, we consider the proposed method with uncoupled structure, i.e., we use different unshared factor functions to represent different groups. As compared with the original coupled structure (denoted by CRNL (coupled)), the uncoupled low-rank function factorization-based nonlocal method, denoted by CRNL (uncoupled), uses a total of $L\times (N+1)$ factor functions to represent groups, while the original CRNL (coupled) uses only $N+1$ factor functions, where $L$ denotes the number of continuous groups and $N$ denotes the number of dimensions. Hence, CRNL (coupled) is expected to hold much lower computational costs.\par 
We show the comparisons between CRNL (coupled) and CRNL (uncoupled) in Table \ref{tab_share} by taking the color image inpainting case as a showing instance. Due to the large memory costs of CRNL (uncoupled), we crop the image size to $100 \times 100\times 3$. Meanwhile, we additionally include three baselines: the nonlocal CP tensor decomposition-based method (NL-CP), the nonlocal Tucker tensor decomposition-based method (NL-Tucker), and the nonlocal fully connected tensor decomposition-based method (NL-FCTN)\cite{NLFCTN}\footnote{The NL-CP, NL-Tucker, and NL-FCTN methods are realized by replacing the coupled low-rank function factorization in our CRNL with classical Tucker/CP/FCTN decompositions \cite{SIAM_review,FCTN}.}. These three traditional NSS-based methods all employ unshared independent tensor decompositions to parameterize different nonlocal patch groups. For fairness, we implement all methods with GPU calculation on the same platform. From Table \ref{tab_share}, we can observe that CRNL (coupled) outperforms other methods including CRNL (uncoupled), and CRNL (coupled) is also more efficient than other NSS methods. Specifically, CRNL (coupled) speeds up the running time about three times against CRNL (uncoupled). The results reveal that the coupled function factorization is indeed helpful to reduce computational costs. Moreover, CRNL (coupled) has better data recovery performance than other nonlocal methods. This is because the coupled function factorization facilitates a beneficial extraction of global correlation knowledge among nonlocal groups intrinsically existed in general on and off-meshgrid data (as shown in Lemma \ref{Pro_2}). As compared, classical nonlocal methods neglect such helpful information across different groups, and hence CRNL (coupled) is expected to more accurately model the underlying structures of data to achieve better recovery performances.
\subsubsection{Sensitivity to Hyperparameters}
\begin{figure}[t]
	\scriptsize
	\setlength{\tabcolsep}{0.9pt}
	\begin{center}
		\begin{tabular}{c}
			\vspace{-0.1cm}
			\includegraphics[width=0.45\textwidth]{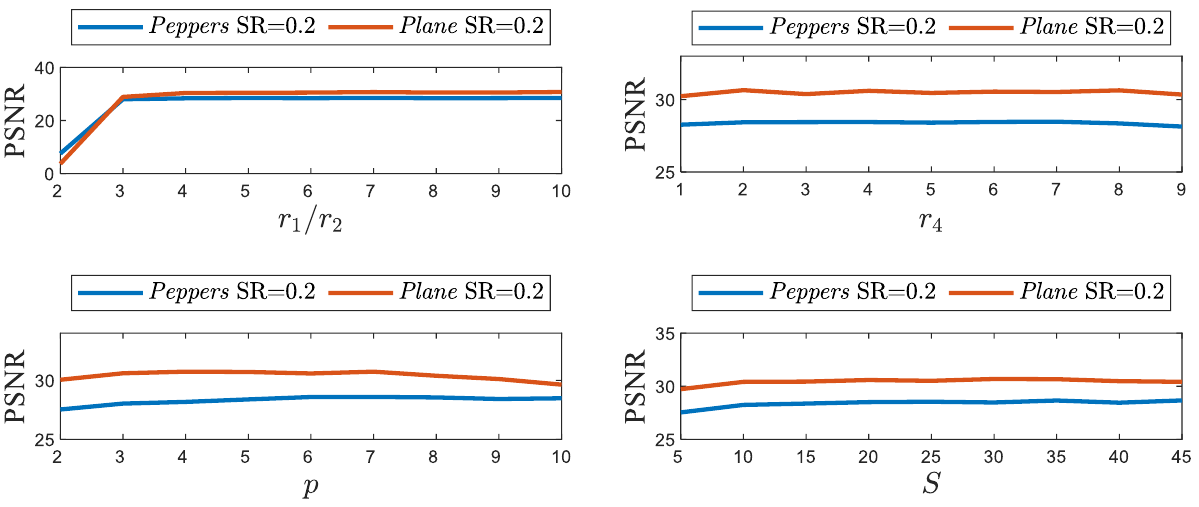}\\
		\end{tabular}
	\end{center}
	\vspace{-0.3cm}
	\caption{The PSNR values of our method w.r.t. different settings of hyperparameters (The rank $r_1$, $r_2$, and $r_4$, the cube size $p$, and the number of similar cubes $S$) for color image inpainting. Here, we fix the rank $r_3$ as the size of the spectral dimension of the color image, i.e., $r_3=3$, to ensure a valid result.\label{fig_hyper}}
	\vspace{-0.2cm}
\end{figure}

In this subsection we test the sensitivity of our method to hyperparameters. In our CRNL, the hyperparameters include the rank of the coupled low-rank function representation $(r_1,r_2,\cdots,r_{N+1})$, the nonlocal cube size $p$, and the number of similar cubes $S$. By taking the image inpainting task as an example, we test the influence of each of these hyperparameters by changing one of them and fixing the others. The results are shown in Fig. \ref{fig_hyper}. We can observe that our method is relatively robust w.r.t. these hyperparameters, which makes it easy to set suitable hyperparameters in our method and reveals its potential applicability in real scenarios.
\section{Conclusions}\label{sec_conclusion}
We have proposed the continuous representation-based nonlocal method, or CRNL, for multi-dimensional data processing on and off-meshgrid. Our CRNL first learns a continuous representation of discrete data using INR, and then stack similar continuous cubes into a group. The continuous groups are compactly represented by the coupled low-rank function factorization with shared factor functions and unshared core tensors. By virtue of the coupled function factorization, our CRNL enjoys favorable computational efficiency and finely characterizes the similarity within each group and across different groups. Extensive experiments on-meshgrid (e.g., image inpainting and denoising) and off-meshgrid (e.g., climate data prediction and point cloud recovery) have validated the effectiveness of our method as compared with state-of-the-art methods. In future work, we can consider enhancing the capabilities of CRNL by proven techniques such as iterative regularizations and weighted schemes \cite{IJCV_WNN}. Besides, more real-world applications on and off-meshgrid, such as hyperspectral fusion \cite{fusion} and spherical image processing \cite{sphere}, would be also interesting. 
\bibliographystyle{ieeetran}
\bibliography{ref}

\begin{thebibliography}{10}
\providecommand{\url}[1]{#1}
\csname url@samestyle\endcsname
\providecommand{\newblock}{\relax}
\providecommand{\bibinfo}[2]{#2}
\providecommand{\BIBentrySTDinterwordspacing}{\spaceskip=0pt\relax}
\providecommand{\BIBentryALTinterwordstretchfactor}{4}
\providecommand{\BIBentryALTinterwordspacing}{\spaceskip=\fontdimen2\font plus
\BIBentryALTinterwordstretchfactor\fontdimen3\font minus
  \fontdimen4\font\relax}
\providecommand{\BIBforeignlanguage}[2]{{%
\expandafter\ifx\csname l@#1\endcsname\relax
\typeout{** WARNING: IEEEtran.bst: No hyphenation pattern has been}%
\typeout{** loaded for the language `#1'. Using the pattern for}%
\typeout{** the default language instead.}%
\else
\language=\csname l@#1\endcsname
\fi
#2}}
\providecommand{\BIBdecl}{\relax}
\BIBdecl

\bibitem{BM3D}
K.~Dabov, A.~Foi, V.~Katkovnik, and K.~Egiazarian, ``Image denoising by sparse
  3-d transform-domain collaborative filtering,'' \emph{IEEE Transactions on
  Image Processing}, vol.~16, no.~8, pp. 2080--2095, 2007.

\bibitem{TIP_14_CS}
W.~Dong, G.~Shi, X.~Li, Y.~Ma, and F.~Huang, ``Compressive sensing via nonlocal
  low-rank regularization,'' \emph{IEEE Transactions on Image Processing},
  vol.~23, no.~8, pp. 3618--3632, 2014.

\bibitem{IJCV_WNN}
S.~Gu, Q.~Xie, D.~Meng, W.~Zuo, X.~Feng, and L.~Zhang, ``Weighted nuclear norm
  minimization and its applications to low level vision,'' \emph{International
  Journal of Computer Vision}, vol. 121, p. 183–208, 2017.

\bibitem{WNLRATV}
Y.~Chen, W.~Cao, L.~Pang, and X.~Cao, ``Hyperspectral image denoising with
  weighted nonlocal low-rank model and adaptive total variation
  regularization,'' \emph{IEEE Transactions on Geoscience and Remote Sensing},
  vol.~60, pp. 1--15, 2022.

\bibitem{NLFCTN}
W.-J. Zheng, X.-L. Zhao, Y.-B. Zheng, and Z.-F. Pang, ``Nonlocal patch-based
  fully connected tensor network decomposition for multispectral image
  inpainting,'' \emph{IEEE Geoscience and Remote Sensing Letters}, vol.~19, pp.
  1--5, 2022.

\bibitem{FTNN}
T.-X. Jiang, M.~K. Ng, X.-L. Zhao, and T.-Z. Huang, ``Framelet representation
  of tensor nuclear norm for third-order tensor completion,'' \emph{IEEE
  Transactions on Image Processing}, vol.~29, pp. 7233--7244, 2020.

\bibitem{LTDL}
X.~Gong, W.~Chen, and J.~Chen, ``A low-rank tensor dictionary learning method
  for hyperspectral image denoising,'' \emph{IEEE Transactions on Signal
  Processing}, vol.~68, pp. 1168--1180, 2020.

\bibitem{DeSCI}
Y.~Liu, X.~Yuan, J.~Suo, D.~J. Brady, and Q.~Dai, ``Rank minimization for
  snapshot compressive imaging,'' \emph{IEEE Transactions on Pattern Analysis
  and Machine Intelligence}, vol.~41, no.~12, pp. 2990--3006, 2019.

\bibitem{TV_LRMR}
W.~He, H.~Zhang, L.~Zhang, and H.~Shen, ``Total-variation-regularized low-rank
  matrix factorization for hyperspectral image restoration,'' \emph{IEEE
  Transactions on Geoscience and Remote Sensing}, vol.~54, no.~1, pp. 178--188,
  2016.

\bibitem{E3DTV}
J.~Peng, Q.~Xie, Q.~Zhao, Y.~Wang, L.~Yee, and D.~Meng, ``Enhanced 3{DTV}
  regularization and its applications on hsi denoising and compressed
  sensing,'' \emph{IEEE Transactions on Image Processing}, vol.~29, pp.
  7889--7903, 2020.

\bibitem{sp}
T.~Yokota, R.~Zdunek, A.~Cichocki, and Y.~Yamashita, ``Smooth nonnegative
  matrix and tensor factorizations for robust multi-way data analysis,''
  \emph{Signal Processing}, vol. 113, pp. 234--249, 2015.

\bibitem{spl_smooth}
O.~Debals, M.~Van~Barel, and L.~De~Lathauwer, ``Nonnegative matrix
  factorization using nonnegative polynomial approximations,'' \emph{IEEE
  Signal Processing Letters}, vol.~24, no.~7, pp. 948--952, 2017.

\bibitem{SCIS_NSS}
Q.~Xie, Q.~Zhao, Z.~Xu, and D.~Meng, ``Color and direction-invariant nonlocal
  self-similarity prior and its application to color image denoising,''
  \emph{Science China Information Sciences}, vol.~63, p. 222101, 2020.

\bibitem{PAMI_12}
Q.~Zhao, P.~Tan, Q.~Dai, L.~Shen, E.~Wu, and S.~Lin, ``A closed-form solution
  to retinex with nonlocal texture constraints,'' \emph{IEEE Transactions on
  Pattern Analysis and Machine Intelligence}, vol.~34, no.~7, pp. 1437--1444,
  2012.

\bibitem{CVPR_18_NL}
X.~Zhang, X.~Yuan, and L.~Carin, ``Nonlocal low-rank tensor factor analysis for
  image restoration,'' in \emph{IEEE/CVF Conference on Computer Vision and
  Pattern Recognition (CVPR)}, 2018, pp. 8232--8241.

\bibitem{TGRS_19_NL}
J.~Xue, Y.~Zhao, W.~Liao, and J.~C.-W. Chan, ``Nonlocal low-rank regularized
  tensor decomposition for hyperspectral image denoising,'' \emph{IEEE
  Transactions on Geoscience and Remote Sensing}, vol.~57, no.~7, pp.
  5174--5189, 2019.

\bibitem{CVPR_17}
Y.~Chang, L.~Yan, and S.~Zhong, ``Hyper-laplacian regularized unidirectional
  low-rank tensor recovery for multispectral image denoising,'' in \emph{IEEE
  Conference on Computer Vision and Pattern Recognition (CVPR)}, 2017, pp.
  5901--5909.

\bibitem{NL_CS}
L.~Feng, H.~Sun, Q.~Sun, and G.~Xia, ``Compressive sensing via nonlocal
  low-rank tensor regularization,'' \emph{Neurocomputing}, vol. 216, pp.
  45--60, 2016.

\bibitem{PAMI_SCI}
L.~Wang, Z.~Xiong, G.~Shi, F.~Wu, and W.~Zeng, ``Adaptive nonlocal sparse
  representation for dual-camera compressive hyperspectral imaging,''
  \emph{IEEE Transactions on Pattern Analysis and Machine Intelligence},
  vol.~39, no.~10, pp. 2104--2111, 2017.

\bibitem{TCYB_NLLR}
L.~Zhang, L.~Song, B.~Du, and Y.~Zhang, ``Nonlocal low-rank tensor completion
  for visual data,'' \emph{IEEE Transactions on Cybernetics}, vol.~51, no.~2,
  pp. 673--685, 2021.

\bibitem{TGRS_Zha}
Z.~Zha, B.~Wen, X.~Yuan, J.~Zhang, J.~Zhou, Y.~Lu, and C.~Zhu, ``Nonlocal
  structured sparsity regularization modeling for hyperspectral image
  denoising,'' \emph{IEEE Transactions on Geoscience and Remote Sensing},
  vol.~61, pp. 1--16, 2023.

\bibitem{SP_NL}
W.~Cao, J.~Yao, J.~Sun, and G.~Han, ``A tensor-based nonlocal total variation
  model for multi-channel image recovery,'' \emph{Signal Processing}, vol. 153,
  pp. 321--335, 2018.

\bibitem{AMM_derain}
Y.~Wang, T.-Z. Huang, X.-L. Zhao, and T.-X. Jiang, ``Video deraining via
  nonlocal low-rank regularization,'' \emph{Applied Mathematical Modelling},
  vol.~79, pp. 896--913, 2020.

\bibitem{IP_nonlocal}
J.~Lu, C.~Xu, Z.~Hu, X.~Liu, Q.~Jiang, D.~Meng, and Z.~Lin, ``A new nonlocal
  low-rank regularization method with applications to magnetic resonance image
  denoising,'' \emph{Inverse Problems}, vol.~38, no.~6, p. 065012, 2022.

\bibitem{TMM_NL}
W.~Cui, S.~Liu, F.~Jiang, and D.~Zhao, ``Image compressed sensing using
  non-local neural network,'' \emph{IEEE Transactions on Multimedia}, vol.~25,
  pp. 816--830, 2023.

\bibitem{TMM_NL2}
Y.~Sun, Y.~Yang, Q.~Liu, J.~Chen, X.-T. Yuan, and G.~Guo, ``Learning
  non-locally regularized compressed sensing network with half-quadratic
  splitting,'' \emph{IEEE Transactions on Multimedia}, vol.~22, no.~12, pp.
  3236--3248, 2020.

\bibitem{CVPR_CS}
L.~Wang, C.~Sun, M.~Zhang, Y.~Fu, and H.~Huang, ``Dnu: Deep non-local unrolling
  for computational spectral imaging,'' in \emph{IEEE/CVF Conference on
  Computer Vision and Pattern Recognition (CVPR)}, 2020, pp. 1658--1668.

\bibitem{CVPR_05}
A.~Buades, B.~Coll, and J.-M. Morel, ``A non-local algorithm for image
  denoising,'' in \emph{IEEE Computer Society Conference on Computer Vision and
  Pattern Recognition (CVPR)}, vol.~2, 2005, pp. 60--65 vol. 2.

\bibitem{TIP_Dong}
W.~Dong, G.~Shi, and X.~Li, ``Nonlocal image restoration with bilateral
  variance estimation: A low-rank approach,'' \emph{IEEE Transactions on Image
  Processing}, vol.~22, no.~2, pp. 700--711, 2013.

\bibitem{TGRS_Chen_21}
X.~Liu, X.~Chen, J.~Li, and Y.~Chen, ``Nonlocal weighted robust principal
  component analysis for seismic noise attenuation,'' \emph{IEEE Transactions
  on Geoscience and Remote Sensing}, vol.~59, no.~2, pp. 1745--1756, 2021.

\bibitem{NLLRF}
L.~Zhuang, X.~Fu, M.~K. Ng, and J.~M. Bioucas-Dias, ``Hyperspectral image
  denoising based on global and nonlocal low-rank factorizations,'' \emph{IEEE
  Transactions on Geoscience and Remote Sensing}, vol.~59, no.~12, pp.
  10\,438--10\,454, 2021.

\bibitem{TVCG_20}
H.~Chen, M.~Wei, Y.~Sun, X.~Xie, and J.~Wang, ``Multi-patch collaborative point
  cloud denoising via low-rank recovery with graph constraint,'' \emph{IEEE
  Transactions on Visualization and Computer Graphics}, vol.~26, no.~11, pp.
  3255--3270, 2020.

\bibitem{TIM_22}
D.~Zhu, H.~Chen, W.~Wang, H.~Xie, G.~Cheng, M.~Wei, J.~Wang, and F.~L. Wang,
  ``Nonlocal low-rank point cloud denoising for 3-{D} measurement surfaces,''
  \emph{IEEE Transactions on Instrumentation and Measurement}, 2022,
  doi=10.1109/TIM.2021.3139686.

\bibitem{STD}
M.~Imaizumi and K.~Hayashi, ``Tensor decomposition with smoothness,'' in
  \emph{International Conference on Machine Learning (ICML)}, vol.~70, 2017,
  pp. 1597--1606.

\bibitem{TSP_CP}
N.~Kargas and N.~D. Sidiropoulos, ``Supervised learning and canonical
  decomposition of multivariate functions,'' \emph{IEEE Transactions on Signal
  Processing}, vol.~69, pp. 1097--1107, 2021.

\bibitem{TTF}
I.~V. Oseledets, ``Constructive representation of functions in low-rank tensor
  formats,'' \emph{Constructive Approximation}, vol.~37, pp. 1--18, 2013.

\bibitem{SIAM_chebfun}
B.~Hashemi and L.~N. Trefethen, ``Chebfun in three dimensions,'' \emph{SIAM
  Journal on Scientific Computing}, vol.~39, no.~5, pp. C341--C363, 2017.

\bibitem{sine}
V.~Sitzmann, J.~Martel, A.~Bergman, D.~Lindell, and G.~Wetzstein, ``Implicit
  neural representations with periodic activation functions,'' in
  \emph{Advances in Neural Information Processing Systems (NeurIPS)}, vol.~33,
  2020, pp. 7462--7473.

\bibitem{INR_NIPS_21}
J.~Lee, J.~Tack, N.~Lee, and J.~Shin, ``Meta-learning sparse implicit neural
  representations,'' in \emph{Advances in Neural Information Processing Systems
  (NeurIPS)}, M.~Ranzato, A.~Beygelzimer, Y.~Dauphin, P.~Liang, and J.~W.
  Vaughan, Eds., vol.~34, 2021, pp. 11\,769--11\,780.

\bibitem{INR_SR_TIP}
C.~Ma, P.~Yu, J.~Lu, and J.~Zhou, ``Recovering realistic details for
  magnification-arbitrary image super-resolution,'' \emph{IEEE Transactions on
  Image Processing}, vol.~31, pp. 3669--3683, 2022.

\bibitem{cvpr_3d}
J.~Chibane, T.~Alldieck, and G.~Pons-Moll, ``Implicit functions in feature
  space for 3{D} shape reconstruction and completion,'' in \emph{IEEE/CVF
  Conference on Computer Vision and Pattern Recognition (CVPR)}, 2020, pp.
  6968--6979.

\bibitem{SAPCU}
W.~Zhao, X.~Liu, Z.~Zhong, J.~Jiang, W.~Gao, G.~Li, and X.~Ji,
  ``Self-supervised arbitrary-scale point clouds upsampling via implicit neural
  representation,'' in \emph{IEEE/CVF Conference on Computer Vision and Pattern
  Recognition (CVPR)}, 2022, pp. 1989--1997.

\bibitem{LIIF}
Y.~Chen, S.~Liu, and X.~Wang, ``Learning continuous image representation with
  local implicit image function,'' in \emph{IEEE/CVF Conference on Computer
  Vision and Pattern Recognition (CVPR)}, 2021, pp. 8624--8634.

\bibitem{INR_SR_NIPS}
J.~Yang, S.~Shen, H.~Yue, and K.~Li, ``Implicit transformer network for screen
  content image continuous super-resolution,'' in \emph{Advances in Neural
  Information Processing Systems (NeurIPS)}, vol.~34, 2021, pp.
  13\,304--13\,315.

\bibitem{TGRS_INR}
K.~Zhang, D.~Zhu, X.~Min, and G.~Zhai, ``Implicit neural representation
  learning for hyperspectral image super-resolution,'' \emph{IEEE Transactions
  on Geoscience and Remote Sensing}, vol.~61, pp. 1--12, 2023.

\bibitem{denoise_INR_1}
C.~Kim, J.~Lee, and J.~Shin, ``Zero-shot blind image denoising via implicit
  neural representations,'' \emph{arXiv:2204.02405}, 2022.

\bibitem{SIAM_review}
T.~G. Kolda and B.~W. Bader, ``Tensor decompositions and applications,''
  \emph{SIAM Review}, vol.~51, no.~3, pp. 455--500, 2009.

\bibitem{Adam}
D.~Kingma and J.~Ba, ``Adam: A method for stochastic optimization,'' in
  \emph{International Conference on Learning Representations (ICLR)}, 2015.

\bibitem{LRTFR}
Y.~Luo, X.~Zhao, Z.~Li, M.~K. Ng, and D.~Meng, ``Low-rank tensor function
  representation for multi-dimensional data recovery,'' \emph{IEEE Transactions
  on Pattern Analysis and Machine Intelligence}, 2023,
  10.1109/TPAMI.2023.3341688.

\bibitem{NLring}
Y.~Chen, W.~He, N.~Yokoya, T.-Z. Huang, and X.-L. Zhao, ``Nonlocal tensor-ring
  decomposition for hyperspectral image denoising,'' \emph{IEEE Transactions on
  Geoscience and Remote Sensing}, vol.~58, no.~2, pp. 1348--1362, 2020.

\bibitem{TRLRF}
L.~Yuan, C.~Li, D.~P. Mandic, J.~Cao, and Q.~Zhao, ``Tensor ring decomposition
  with rank minimization on latent space: An efficient approach for tensor
  completion,'' in \emph{AAAI Conference on Artificial Intelligence (AAAI)},
  2019, pp. 9151--9158.

\bibitem{DIP}
D.~Ulyanov, A.~Vedaldi, and V.~Lempitsky, ``Deep image prior,''
  \emph{International Journal of Computer Vision}, vol. 128, p. 1867–1888,
  2020.

\bibitem{HLRTF}
Y.~Luo, X.-L. Zhao, D.~Meng, and T.-X. Jiang, ``{HLRTF}: Hierarchical low-rank
  tensor factorization for inverse problems in multi-dimensional imaging,'' in
  \emph{IEEE/CVF Conference on Computer Vision and Pattern Recognition (CVPR)},
  2022, pp. 19\,303--19\,312.

\bibitem{LogDet}
M.~Yang, Q.~Luo, W.~Li, and M.~Xiao, ``3-{D} array image data completion by
  tensor decomposition and nonconvex regularization approach,'' \emph{IEEE
  Transactions on Signal Processing}, vol.~70, pp. 4291--4304, 2022.

\bibitem{TCTV}
H.~Wang, J.~Peng, W.~Qin, J.~Wang, and D.~Meng, ``Guaranteed tensor recovery
  fused low-rankness and smoothness,'' \emph{IEEE Transactions on Pattern
  Analysis and Machine Intelligence}, 2023, doi=10.1109/TPAMI.2023.3259640.

\bibitem{CAVE}
F.~Yasuma, T.~Mitsunaga, D.~Iso, and S.~K. Nayar, ``Generalized assorted pixel
  camera: Postcapture control of resolution, dynamic range, and spectrum,''
  \emph{IEEE Transactions on Image Processing}, vol.~19, no.~9, pp. 2241--2253,
  2010.

\bibitem{LRTDTV}
Y.~Wang, J.~Peng, Q.~Zhao, Y.~Leung, X.-L. Zhao, and D.~Meng, ``Hyperspectral
  image restoration via total variation regularized low-rank tensor
  decomposition,'' \emph{IEEE Journal of Selected Topics in Applied Earth
  Observations and Remote Sensing}, vol.~11, no.~4, pp. 1227--1243, 2018.

\bibitem{RCTV}
J.~Peng, H.~Wang, X.~Cao, X.~Liu, X.~Rui, and D.~Meng, ``Fast noise removal in
  hyperspectral images via representative coefficient total variation,''
  \emph{IEEE Transactions on Geoscience and Remote Sensing}, vol.~60, pp.
  1--17, 2022.

\bibitem{HSIDCNN}
Q.~Yuan, Q.~Zhang, J.~Li, H.~Shen, and L.~Zhang, ``Hyperspectral image
  denoising employing a spatial–spectral deep residual convolutional neural
  network,'' \emph{IEEE Transactions on Geoscience and Remote Sensing},
  vol.~57, no.~2, pp. 1205--1218, 2019.

\bibitem{SDeCNN}
A.~Maffei, J.~M. Haut, M.~E. Paoletti, J.~Plaza, L.~Bruzzone, and A.~Plaza, ``A
  single model {CNN} for hyperspectral image denoising,'' \emph{IEEE
  Transactions on Geoscience and Remote Sensing}, vol.~58, no.~4, pp.
  2516--2529, 2020.

\bibitem{Lee_1997}
S.~Lee, G.~Wolberg, and S.~Shin, ``Scattered data interpolation with multilevel
  b-splines,'' \emph{IEEE Transactions on Visualization and Computer Graphics},
  vol.~3, no.~3, pp. 228--244, 1997.

\bibitem{FCTN}
Y.-B. Zheng, T.-Z. Huang, X.-L. Zhao, Q.~Zhao, and T.-X. Jiang,
  ``Fully-connected tensor network decomposition and its application to
  higher-order tensor completion,'' in \emph{AAAI Conference on Artificial
  Intelligence (AAAI)}, 2021, pp. 11\,071--11\,078.

\bibitem{fusion}
Q.~Xie, M.~Zhou, Q.~Zhao, Z.~Xu, and D.~Meng, ``M{HF}-{N}et: An interpretable
  deep network for multispectral and hyperspectral image fusion,'' \emph{IEEE
  Transactions on Pattern Analysis and Machine Intelligence}, vol.~44, no.~3,
  pp. 1457--1473, 2022.

\bibitem{sphere}
J.~Li, C.~Huang, R.~Chan, H.~Feng, M.~K. Ng, and T.~Zeng, ``Spherical image
  inpainting with frame transformation and data-driven prior deep networks,''
  \emph{SIAM Journal on Imaging Sciences}, vol.~16, no.~3, pp. 1177--1194,
  2023.

\end{thebibliography}
\end{document}